\pdfoutput=1

\documentclass[11pt]{article}

\usepackage{ACL2023}

\usepackage{times}
\usepackage{latexsym}

\usepackage[T1]{fontenc}

\usepackage[utf8]{inputenc}

\usepackage{microtype}

\usepackage{inconsolata}

\usepackage{graphicx}
\usepackage{multirow}
\usepackage{booktabs}
\usepackage{arydshln}
\usepackage{colortbl}
\usepackage{enumitem}

\usepackage{amsmath,amsfonts,bm}
\usepackage[nameinlink]{cleveref}

\crefformat{section}{\S#2#1#3} 
\crefname{algorithm}{Alg.}{Algs.}
\crefformat{subsection}{\S#2#1#3}
\Crefname{equation}{Eq.}{Eqs.}
\Crefname{figure}{Fig.}{Figs.}

\newcommand{\orange}[1]{\textcolor{orange}{#1}}

\newcommand{\blue}[1]{\textcolor{blue}{#1}}

\newcommand{\teal}[1]{\textcolor{teal}{#1}}

\makeatletter   
\newcommand{\sveryshortarrow}[1][3pt]{\mathrel{%
    \vcenter{\hbox{\rule[-.5\fontdimen8\scriptfont3]
               {\scriptratio\dimexpr#1\relax}{\fontdimen8\scriptfont3}}}%
   \mkern-4mu\hbox{\let\f@size\sf@size\usefont{U}{lasy}{m}{n}\symbol{41}}}}
\makeatother

\def\eqref#1{equation~\ref{#1}}

\def\1{\bm{1}}

\def\m1{{\bm{1}}}

\DeclareMathAlphabet{\mathsfit}{\encodingdefault}{\sfdefault}{m}{sl}
\SetMathAlphabet{\mathsfit}{bold}{\encodingdefault}{\sfdefault}{bx}{n}

\def\gM{{\mathcal{M}}}

\def\sC{{\mathbb{C}}}

\DeclareMathOperator*{\argmax}{arg\,max}

\title{Unsupervised Summarization Re-ranking}

\author{Mathieu Ravaut$^1$$^,$$^2$,~
Shafiq Joty$^*$$^1$$^,$$^3$~
Nancy F. Chen$^2$ \\
$^1$ Nanyang Technological University, Singapore\\
$^2$ Institute of Infocomm Research (I$^{2}$R), A$^{*}$STAR, Singapore\\
$^3$ Salesforce AI\\
\texttt{\{mathieuj001@e.ntu, srjoty@ntu\}.edu.sg}\\
\texttt{nfychen@i2r.a-star.edu.sg}
}

\begin{document}

\maketitle
\def\thefootnote{*}\footnotetext{Work done when the author was on leave from NTU.}\def\thefootnote{\arabic{footnote}}

\begin{abstract}

With the rise of task-specific pre-training objectives, abstractive summarization models like PEGASUS offer appealing zero-shot performance on downstream summarization tasks. However, the performance of such unsupervised models still lags significantly behind their supervised counterparts. Similarly to the supervised setup, we notice a very high variance in quality among summary candidates from these models while only one candidate is kept as the summary output. In this paper, we propose to re-rank summary candidates in an \emph{unsupervised} manner, aiming to close the performance gap between unsupervised and supervised models. Our approach improves the unsupervised PEGASUS by up to 7.27\% and ChatGPT by up to 6.86\% relative mean {\sc{Rouge}} across four widely-adopted summarization benchmarks ; and achieves relative gains of 7.51\% (up to 23.73\% from XSum to WikiHow) averaged over 30 zero-shot transfer setups (finetuning on a dataset, evaluating on another).\footnote{Code for all experiments are available at \url{https://github.com/ntunlp/SummScore}.}

\end{abstract}

\section{Introduction}
\label{sec:intro}

Transformer-based encoder-decoder language models have achieved great success in abstractive summarization in the last few years, and produce fluent summaries which can be quite abstractive \citep{raffel2019exploring, lewis-etal-2020-bart, zhang2020pegasus}. These models follow the \emph{pre-train then fine-tune} paradigm: they are first pre-trained with a self-supervised objective on a large text corpus; then they are fine-tuned on the downstream dataset of interest, using the available supervision, which may be very scarce. Finding a better pre-training objective remains an active research area. Some models like T5 \citep{raffel2019exploring} and BART \citep{lewis-etal-2020-bart} adopt a more general language modeling objective (e.g., masked span generation), while others like PEGASUS \citep{zhang2020pegasus} or TED \citep{yang-etal-2020-ted} are pre-trained specifically for the task of summarizing a document. PEGASUS uses salient sentences of the document as a proxy summary label, while TED leverages the lead bias to get the pseudo-summary target.

\begin{table}[]
\resizebox{0.99\columnwidth}{!}{
\begin{tabular}{llccc}
\toprule
\multicolumn{1}{c}{\textbf{Generation method}} & \multicolumn{1}{c}{\textbf{Summary candidate}} 
& \textbf{R-1}              
& \textbf{R-2}              
& \textbf{R-L} \\
\midrule
\multirow{4}{*}{Beam search} 
& First (top beam) & 35.47 & 13.89 & 31.61 \\
& Random & 34.89 & 13.46 & 31.22 \\
& Minimum  & 26.64 & 7.68  & 23.18 \\
& Maximum (oracle) & \textbf{42.62} & \textbf{19.76} & \textbf{38.75} \\
\midrule
\multirow{4}{*}{Diverse beam search} 
& First & 34.35 & 13.02 & 30.65 \\
& Random & 31.73 & 11.22 & 28.4  \\
& Minimum  & 21.25 & 4.45  & 18.61 \\
& Maximum (oracle)  & \textbf{41.87} & \textbf{19.29} & \textbf{38.22} \\
\midrule
\multirow{4}{*}{Nucleus sampling}  
& First & 32.14 & 11.29 & 28.66 \\
& Random & 32.12 & 11.29 & 28.64 \\
& Minimum  & 24.09 & 6.49  & 21.19 \\
& Maximum (oracle) & \textbf{40.19} & \textbf{17.47} & \textbf{36.43} \\
\bottomrule
\end{tabular}
}
\caption{\small {\sc{Rouge}} results with PEGASUS (unsupervised) on CNN/DM test set, for three generation methods to produce 20 summary candidates, and four candidate selection strategies. \textbf{R-1}, \textbf{R-2}, \textbf{R-L} stands for {\sc{Rouge-1/2/L}}.}
\label{tab:1}
\vspace{-1.0em}
\end{table}

Despite the impressive success on supervised abstractive summarization tasks, unsupervised summarization remains very challenging. The LEAD-3 (extractive) baseline which simply takes the first three sentences of a document as its summary, remains far ahead of unsupervised approaches on several news summarization datasets \citep{see-etal-2017-get}, especially the popular CNN/DM dataset \citep{hermann2015teaching}. In fact, it was only improved on by \emph{supervised} abstractive models not more than five years ago \citep{narayan-etal-2018-dont}. It is expected that a model which has never seen \emph{any} summarization example would struggle, as summarization is a task that is subjective and complex even for humans \citep{kryscinski-etal-2019-neural}. Since summarization labels are expensive to collect, it is essential to develop models with good zero-shot performance. Starting from instruction-tuned GPT-3, LLMs are offering promising performance in zero-shot summarization \cite{goyal2022news}, but remain an unscalable solution as these models are rarely open-source, and extremely computationally intensive.

Recently, in the supervised setup, second-stage approaches have gathered interest in abstractive summarization research. While the base encoder-decoder model is trained with maximum-likelihood estimation (MLE) to predict each token of the ground-truth summary in an autoregressive manner, second-stage methods work with a global view at the whole sequence level. SimCLS \citep{liu-liu-2021-simcls} and SummaReranker \citep{ravaut-etal-2022-summareranker} propose to train another neural model to rank summary candidates generated by decoding methods like beam search \citep{reddy1977speech} or diverse beam search \citep{vijayakumar2016diverse}. BRIO \citep{liu-etal-2022-brio} bypasses the need for another model, and re-uses the fine-tuned model for another fine-tuning stage in which the model also learns to rank candidates in the correct order. SummaFusion \citep{ravaut-etal-2022-towards} encodes each summary candidate separately and decodes into a new, abstractive second-stage summary. Such second-stage methods have improved {\sc{Rouge-1}} state-of-the-art on CNN/DM by more than 3 points \citep{liu-etal-2022-brio}.

In this paper, we propose to re-rank summary candidates in the \emph{unsupervised} setup. Following observations made by second-stage summarization studies in the supervised setup \citep{liu-etal-2021-refsum, ravaut-etal-2022-summareranker}, we also observe large variance in performance among summary candidates in the unsupervised setup. In \Cref{tab:1}, the \emph{\textbf{oracle}} for PEGASUS, which is the summary candidate maximizing the {\sc{Rouge}} score with the reference, reaches 42.62 when using beam search with 20 beams on CNN/DM \citep{hermann2015teaching}. This is in the same range (42-45 {\sc{Rouge-1}}) as the top beam of \emph{supervised} leading models on this dataset \citep{lewis-etal-2020-bart, zhang2020pegasus}. This observation implies strong potential motivating our work: \textbf{\emph{with a perfect unsupervised summarization re-ranker, one could potentially by-pass supervised fine-tuning and just re-rank instead}}.

The main challenge lies in the fact that the re-ranker must also not access any supervision. 
Our proposed model does not train any neural model, but simply computes features indicative of summary quality to score each summary candidate, some of them which also leverage the source document. A weighted average of these features is used for candidate re-ranking, and we explore several methods to estimate the feature weights. Our method, named SummScore, is lightweight, fast and easy to use as it does not rely on a neural network. Since it is purely unsupervised, the re-ranked results can provide more refined self-supervision to the pre-trained models, complementing the pre-training with rounds of self-training.

Our contributions in this paper are threefold:

\begin{itemize}[leftmargin=*]

    \item We propose SummScore, the first system to re-rank summarization candidates in an unsupervised setup and in an unsupervised manner.
    
    \item We demonstrate the strength of SummScore by consistent performance improvement: up to +7.27\% with PEGASUS and +6.86\% with ChatGPT\footnote{https://chat.openai.com/} mean {\sc{Rouge}} gains over four unsupervised summarization datasets, +7.51\% mean {\sc{Rouge}} gains averaged over 30 zero-shot transfer setups. 
    
    \item Using the re-ranker, we derive an original and effective self-training method which continuously improves the base unsupervised summarization model, pushing PEGASUS from 35.47 to 39.76 {\sc{Rouge-1}} (+12.09\%). 
    
\end{itemize}

\section{Related Work}
\label{sec:related}

\paragraph{Unsupervised abstractive summarization} In unsupervised abstractive summarization, SummAE \citep{liu2019summae} proposes to auto-encode paragraphs with a sequence-to-sequence model and decode single-sentence summaries from the latent embeddings. SEQ3 \citep{baziotis-etal-2019-seq} also uses an auto-encoder to compress the input then reconstruct it into a differentiable manner, the encoder output serving as a summary. However, both methods stick to unsupervised \emph{sentence} summarization. More recent approaches typically rely on language models being pre-trained, then used in a zero-shot fashion. PEGASUS \citep{zhang2020pegasus} treats salient sentences as pseudo abstractive targets to build a pre-training objective. TED \citep{yang-etal-2020-ted} exploits the lead bias in news articles and takes out the first sentences of the document as pseudo summary targets for pre-training. Due to their pre-training objective built for summary generation, these pre-trained models can be directly used for unsupervised summarization. The Summary Loop \citep{laban-etal-2020-summary} uses reinforcement learning to train a model to fill-in deleted important words from the source document using the summary generated so far, then refines this summary.

\paragraph{Re-ranking in abstractive summarization} Second-stage or sequence-level methods are gaining traction recently in \emph{supervised} summarization. Among such methods, re-ranking consists in selecting a better summary candidate out of several of them produced by a base model (which has already been fine-tuned). RefSum \citep{liu-etal-2021-refsum} uses a meta-learning approach to learn how to rank summaries coming from multiple systems. SimCLS \citep{liu-liu-2021-simcls} trains a RoBERTa \citep{liu2019roberta} model with a ranking loss to learn how to rank summary candidates generated by a base BART or PEGASUS in their target metric order. SummaReranker \citep{ravaut-etal-2022-summareranker} also trains a RoBERTa re-ranker, but this time in a multi-label binary classification manner to predict whether each summary candidate maximizes each of the metrics of interest. To avoid using another neural network for re-ranking, BRIO \citep{liu2022brio} performs a second fine-tuning stage with the re-ranking loss built in the base summarization system. Each of the four models above improves the SOTA on the CNN/DM benchmark, reaching 47.78 {\sc{Rouge-1}} for BRIO. 

To the best of our knowledge, there is no work on sequence-level unsupervised abstractive summarization. Concurrently to our work, MBRD \citep{suzgun2022follow} proposes to rank generated candidates in several generation tasks using majority voting based on BERTScore \citep{zhang2019bertscore}.

\section{Method}
\label{sec:model}

\subsection{Unsupervised Summary Re-ranking}

As an unsupervised summarization re-ranking approach, our method assumes access to a zero-shot self-supervised summarization model. We refer to it as the base model $\gM_\text{base}$. Given a source document $D$, $\gM_\text{base}$ will generate $k$ \emph{summary candidates} using a \emph{generation method} to transform model predictions into a natural language summary. A widely used such generation approach is beam search, which maintains $k$ top summary candidates throughout decoding, ranking them with decreasing mean log-probability of the sequence. In the end, practitioners keep the candidate maximizing the log-probability and discard the remaining, whereas we propose to keep \emph{all} $k$ candidates and re-rank them, following \citep{ravaut-etal-2022-summareranker}.

Let $\sC = \{C_1, \ldots, C_k\}$ be the pool of candidates. Our goal in (re-)ranking the candidates is to assign to each of them a score $S$, such that $S(C_i) > S(C_j)$ if $C_i$ is a better candidate than $C_j$ (for $1 \le i,j \le k $) according to some summary quality measures. We can then select the candidate maximizing the score as the best output:

\begin{equation} 
    {C_{S}^*} = \argmax_{C_i \in \sC}~ \{ S(C_1), \ldots, S(C_k) \} \label{eq:top}
\end{equation}

Unlike re-ranking in a supervised setup, where one can compute such scores by comparing with the ground truth summary or build models to optimize them \citep{liu-liu-2021-simcls,ravaut-etal-2022-summareranker,liu-etal-2022-brio}, in our unspervised setup, we cannot assume access to the ground truth, which thus excludes scoring the candidate with regards to it (e.g., using {\sc{Rouge}}). In the following, we describe how we build our unsupervised scoring method (named \emph{SummScore}) following principles assessing the quality of a summary.


\begin{figure}[t!]
    \centering
    \includegraphics[width=\columnwidth]{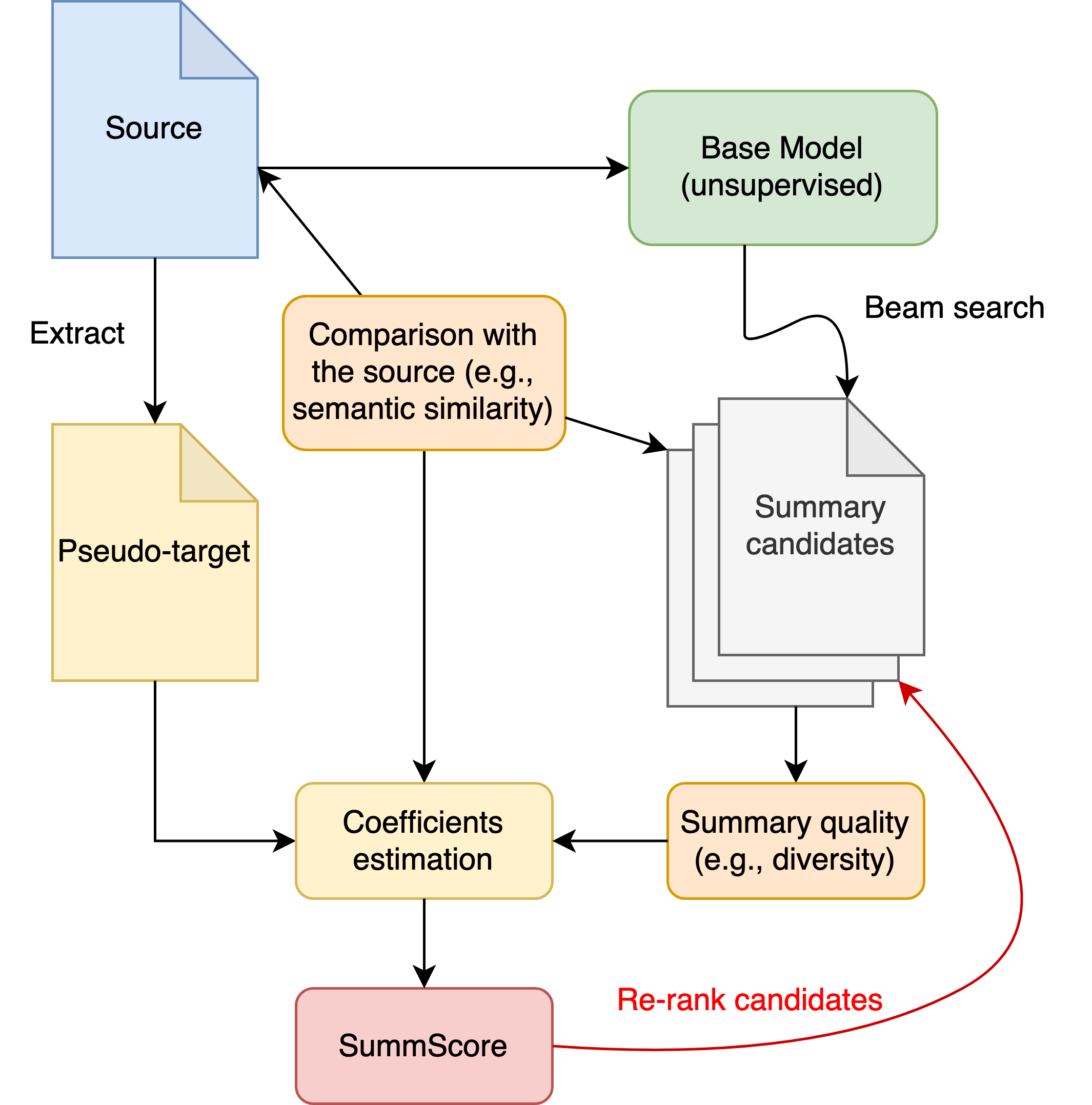}
    \caption{\small \textbf{SummScore (unsupervised) re-ranking} construction. SummScore leverages the source document for semantic similarity comparisons with summary candidates, as well as to extract a pseudo target.}
    \label{fig:1}
\end{figure}

\subsection{Multi-Objective Re-ranking Score}
\label{subsec:3_2}

We design our candidate-level SummScore as an aggregation of features, each representing desired properties for a summary. Features either come from the comparison between the summary candidate and the source, or from the candidate itself. \Cref{fig:1} synthesizes the overall SummScore re-ranking process.

\paragraph{Comparison with the source} 

One evident property of a summary is that it should stick to the source content, and contain as much of the important content as possible. The most straightforward way to measure this consists in using n-gram overlap metrics between the source document and each candidate. We use {\sc{Rouge-1}} (noted \emph{R-1}) \citep{lin-2004-rouge}, {\sc{Rouge-2}} (\emph{R-2}), and {\sc{Bleu}} \citep{papineni-etal-2002-bleu}, which form our first set of features:

\begin{equation} 
    S_{\text{overlap}} = \{\text{R-1},\text{R-2},\text{{\sc{Bleu}}} \} 
    \label{eq:overlap}
\end{equation}

The above metrics only evaluate n-gram overlap, which can be helpful penalizing summary candidates departing too much from the source, potentially hallucinating. However, they have been shown to not be well suited at evaluating semantic similarity, and might encourage too much copying. 

Thus, our next batch of SummScore features consists in model-based metrics designed to capture semantic similarity between two text items. We explore three such metrics: BERTScore \citep{zhang2019bertscore}, BARTScore \citep{yuan2021bartscore} and BLEURT \citep{sellam-etal-2020-bleurt}. BERTScore (noted \emph{BS}) computes token-level cosine similarity between the contextual embeddings of the pre-trained BERT \citep{devlin-etal-2019-bert} of each text item to compare. BARTScore (noted \emph{BaS}) uses BART \citep{lewis-etal-2020-bart} token-level log-probabilities from the pre-trained BART to score the generated text. BLEURT (noted \emph{BRT}) also leverages BERT but extends its pre-training with an additional multi-task pre-training on synthetic data. 
Our next features are: 

\begin{equation} 
    S_{\text{semantic}} = \{\text{BS}, \text{BaS}, \text{BRT} \} \label{eq:sem}
\end{equation}

When each of these metrics is referred to, it is implicit that they are used to compare a summary candidate with the source document (in contrast to the supervised case, comparing with the target).

\paragraph{Summary quality} 

A good summary  should be \emph{diverse}, meaning it should avoid repeated $n$-grams. We build a summary-level diversity score which measures the proportion of unique $n$-grams.

\begin{equation} 
    F_{\text{div}} = \frac{1}{N} \Sigma_{n=1}^{N} \frac{\text{unique $n$-grams}}{\text{total $n$-grams}}
\end{equation}

We take $N=3$ in practice.
The summary should not be too short, nor too long. We penalize summaries which deviate a lot from the average summary length on the given dataset. To build a score with increasing values being desirable, we use a smooth inverse of the absolute length difference between the summary candidate and the mean length of summaries $\mu_{\text{len}}$.

\begin{equation} 
    F_{\text{len}} = \frac{1}{\text{max}(1, |\text{length} - \mu_{\text{len}}|)}
\end{equation}

\paragraph{Final Score} 

Our final set of summary features is:

\begin{align} 
\begin{split}
    S &= S_{\text{overlap}} \cup S_{\text{semantic}}  \cup S_{\text{quality}} \\
    &= \{F_{1}, \dots, F_{|S|}\}
\end{split}
\end{align}

where $S_{\text{quality}} = \{F_{\text{div}}, F_{\text{len}} \}$. For data point $x_{i}$, SummScore simply outputs the summary candidate among the set $\sC_{i}$ maximizing a weighted combination of all features above:

\begin{align} 
    \text{SummScore}_{\theta}(\sC_{i}) =  \argmax_{C_i \in \sC_{i}}\sum\limits_{j=1}^{|S|} \theta_{j}.F_{j}(C_{i})
    \label{eqn:sumscr_output}
\end{align}

\noindent where we enforce coefficients to be $\sum\limits_{j=1}^{|S|} \theta_{j} = 1.0$

\subsection{Coefficients Estimation}
\label{subsec:3_3}

\begin{table*}
\resizebox{0.99\textwidth}{!}{
\begin{tabular}{llccccccccc}
\toprule
\multirow{2}{*}{\textbf{Dataset}} 
& \multirow{2}{*}{\textbf{Domain}} 
& \multicolumn{3}{c}{\textbf{\# Data points}} 
& \multicolumn{2}{c}{\textbf{\# Words}} 
& \multicolumn{2}{c}{\textbf{\# Tokens (PEGASUS)}} 
& \multicolumn{2}{c}{\textbf{New summary n-grams}} \\
   & & Train & Val & Test & Doc. & Summ. & Doc. & Summ. & 1-grams (\%) & 2-grams (\%) \\
\midrule
CNN/DM \cite{hermann2015teaching} & News & 287113 & 13334 & 11490 & 786.68 & 55.06 & 851.53 & 64.57 & 12.07 & 51.05\\
XSum \cite{narayan-etal-2018-dont} & News & 204045 & 11332 & 11334 & 430.18	& 23.19 & 456.96 & 26.01 & 33.98 & 83.33 \\
WikiHow \cite{koupaee2018wikihow} & Instructions & 157304 & 5600 & 5580 & 588.06 & 62.10 & 620.52 & 71.82 & 29.79 & 77.45 \\
SAMSum \cite{gliwa-etal-2019-samsum} & Dialogue & 14732 & 818 & 819 & 124.07 & 23.42 & 133.07 & 25.66 & 33.88	& 79.02 \\
\bottomrule
\end{tabular}
}
\vspace{-0.5em}
\caption{\small Statistics on the datasets used for experiments. \textbf{Doc.} is the source document, \textbf{Summ.} the summary.}
\label{tab:2}
\vspace{-1.0em}
\end{table*}

SummScore is simply a linear combination of eight features in total. Yet a last crucial question remains: how to estimate the coefficients to assign to each feature? We propose to bootstrap a pseudo-summary using sentences from the source document. Coefficients are then tuned to maximize the mean of {\sc{Rouge-1/2/L}} between the summary candidate with the highest SummScore (e.g., SummScore output candidate), and the pseudo-target. We compare three approaches to extract pseudo-targets:

\begin{itemize}[leftmargin=*]

    \item \textbf{Random-3}: As a baseline, we randomly select three sentences from the source document to form a pseudo-target.
    
    \item \textbf{LEAD-3}: This consists in the first three sentences of the document. LEAD-3 is a strong baseline for lead-biased news summarization datasets \citep{hermann2015teaching, see-etal-2017-get}, and it has even been used as a pseudo-target for summarization pre-training in TED \citep{yang-etal-2020-ted}.
    
    \item \textbf{Salient Sentences}: We follow the \emph{gap-sentences generation} idea introduced by PEGASUS pre-training objective \citep{zhang2020pegasus}, and also used by SUPERT \citep{gao-etal-2020-supert} for unsupervised summarization evaluation. A pseudo-target is constructed with salient sentences, which are defined as the source sentences maximizing the {\sc{Rouge}} with the rest of the document. The top 30\% such sentences are extracted to form a pseudo-summary. We experiment with all three standard versions {\sc{Rouge-1}}, {\sc{Rouge-2}} and {\sc{Rouge-L}} for salient sentences definition, referred to as \textbf{Salient-R1}, \textbf{Salient-R2} and \textbf{Salient-RL}, respectively. 
    
\end{itemize}

We emphasize that none of these pseudo-targets definition makes any access to human supervision. Training SummScore amounts to estimating the coefficients $\theta$ in \Cref{eqn:sumscr_output} using the pseudo-targets:

\begin{equation} 
\hat{\theta} = \argmax_{\theta} \sum\limits_{i} \mathcal{R}(\tilde{y_{i}}, \text{SummScore}_{\theta}(\sC_{i}))
\end{equation}

\noindent where $\mathcal{R}$ is the mean of {\sc{Rouge-1}}, {\sc{Rouge-2}} and {\sc{Rouge-L}}, $\sC_{i}$  is the set of candidates predicted by the base model $\gM_\text{base}$ for data point $x_{i}$, and $\tilde{y_{i}}$ is the pseudo-target. To optimize coefficients, we hill climb with randomness to maximize $\mathcal{R}$ between the SummScore selected summary candidate, and the pseudo-target. Specifically, we estimate coefficients with stochastic local search on the validation set in a hierarchical manner: we first tune coefficients for $S_{overlap}$ and $S_{semantic}$ separately, then estimate coefficients for $S_{quality} \cup \{F_{overlap}, F_{semantic}\}$, where $F_{overlap}$ (resp. $F_{semantic}$) is the set $S_{overlap}$ (resp. $S_{semantic}$) after reduction to a single feature. Such hierarchical estimation is natural given that $S_{overlap}$ (resp. $S_{semantic}$) is made of features capturing similar properties, and dramatically reduces the search space. 





\section{Experiments}
\label{sec:experiments}

\subsection{Setup}
\label{subsec:setup}

We experiment on four popular abstractive summarization datasets, from three different domains (see \Cref{tab:2} for basic statistics on each dataset):
\begin{itemize}[leftmargin=*]

    \item \textbf{CNN-DailyMail} \citep{hermann2015teaching, see-etal-2017-get} is made of 93k and 220k articles from the CNN and DailyMail newspapers, respectively. CNN/DM is the most extractive dataset among all the ones we consider and has the longest source documents.

    \item \textbf{XSum} \citep{narayan-etal-2018-dont} has 227k articles from the BBC from 2010 to 2017. This is an extreme summarization task, compressing each article into a single, very abstractive sentence. 

    \item \textbf{WikiHow} \citep{koupaee2018wikihow} contains 168k lists of short instruction sequences.

    \item \textbf{SAMSum} \citep{gliwa-etal-2019-samsum} is a dialogue summarization dataset containing 17k conversations. In this dataset, source length is significantly shorter than in the other datasets.
    
\end{itemize}

To estimate coefficients, we subsample randomly (on datasets other than SAMSum) 1,000 data points from the validation set. 
To avoid coefficients optimization to overfit, we cap each random search at 1,000 trials. Evaluation of summaries selected by SummScore is done with the standard {\sc{Rouge-1/2/L}} \citep{lin-2004-rouge} (using summary-level {\sc{Rouge-LSum}} variant for {\sc{Rouge-L}}) and BERTScore \citep{zhang2019bertscore}. We use \emph{transformers} \citep{wolf-etal-2020-transformers} and \emph{datasets} \citep{lhoest-etal-2021-datasets} for pre-trained checkpoints and datasets, respectively.



\begin{table*}[]
\resizebox{0.99\textwidth}{!}{
\begin{tabular}{llcccccccccccc}

\toprule 
\textbf{Backbone} &
{\textbf{Model}}                
& \multicolumn{3}{c}{\textbf{CNN/DM}} 
& \multicolumn{3}{c}{\textbf{XSum}} 
& \multicolumn{3}{c}{\textbf{WikiHow}}
& \multicolumn{3}{c}{\textbf{SAMSum}} \\
\textbf{$\gM_\text{base}$}        
& \textbf{Candidate Selection}     
& \textbf{R-1/R-2/R-L}
& \textbf{BS}                                 
& \textbf{Gain (\%)}  
& \textbf{R-1/R-2/R-L}
& \textbf{BS}                                 
& \textbf{Gain (\%)}   
& \textbf{R-1/R-2/R-L} 
& \textbf{BS}          
& \textbf{Gain (\%)}   
& \textbf{R-1/R-2/R-L} 
& \textbf{BS}          
& \textbf{Gain (\%)}   \\

\midrule 

\multirow{8}{*}{\textbf{PEGASUS}} 

& Top beam \cite{zhang2020pegasus} & 32.90/13.28/29.38 & \_ & \_ & 19.27/\textbf{3.00}/12.72 & \_ & \_ & 22.59/6.10/14.44 & \_ & \_ &  \_ & \_ & \_ \\
& Top beam                         & 35.47/13.89/31.61 & 86.29 & \_ & 18.77/2.86/13.85 & 85.66 & \_ & 25.49/5.91/17.99 & \textbf{84.98} & \_ & 26.64/6.32/22.75 & 86.12 & \_ \\
& Random beam                      & 34.89/13.46/31.22 & 86.11 & -1.67 & 18.58/2.81/13.90 & 85.29 & -1.31 & 25.39/6.00/18.09 & 84.82 & -0.38 & 25.27/5.80/21.78 & 85.31 & -5.26 \\

\cdashline{2-14}

& SummScore - Random-3             & 35.92$^{\dagger}$/14.26$^{\dagger}$/32.34$^{\dagger}$ & 86.28 & 1.96 & 19.37$^{\dagger}$/\textbf{2.99}$^{\dagger}$/14.52$^{\dagger}$ & 85.78$^{\dagger}$ & 3.89 & 26.29$^{\dagger}$/\textbf{6.28}$^{\dagger}$/\textbf{18.78}$^{\dagger}$ & \textbf{84.98} & 3.89 & 28.09$^{\dagger}$/\textbf{7.26}$^{\dagger}$/\textbf{24.42}$^{\dagger}$ & \textbf{86.39}$^{\dagger}$ & \textbf{7.27} \\
& SummScore - LEAD-3               & \cellcolor{blue!15} \textbf{36.92}$^{\dagger}$/\textbf{15.03}$^{\dagger}$/\textbf{33.19}$^{\dagger}$ & \cellcolor{blue!15}\textbf{86.54}$^{\dagger}$ & \cellcolor{blue!15} \textbf{5.19} & \cellcolor{blue!15}\textbf{19.62}$^{\dagger}$/\textbf{3.02}$^{\dagger}$/\textbf{14.71}$^{\dagger}$ & \cellcolor{blue!15} \textbf{85.92}$^{\dagger}$ & \cellcolor{blue!15} \textbf{5.24} & 26.17$^{\dagger}$/6.19$^{\dagger}$/18.69$^{\dagger}$ & \textbf{84.96} & 3.16 & \cellcolor{blue!15}\textbf{28.22}$^{\dagger}$/\textbf{7.16}/\textbf{24.39}$^{\dagger}$ & \cellcolor{blue!15}\textbf{86.41}$^{\dagger}$ & \cellcolor{blue!15}\textbf{7.27}\\
& SummScore - Salient-R1           & 35.54/14.05/32.04$^{\dagger}$ & 86.22 & 0.85 & 18.96/2.88/14.19$^{\dagger}$ & 85.65 & 1.52 & \textbf{26.37}$^{\dagger}$/\textbf{6.32}$^{\dagger}$/\textbf{18.81}$^{\dagger}$ & \textbf{84.92} & 4.25 & 27.89$^{\dagger}$/7.08/24.08$^{\dagger}$ & 86.25 & 5.98 \\
& SummScore - Salient-R2           & 35.65/14.12/32.14$^{\dagger}$ & 86.24 & 1.19 & 19.13$^{\dagger}$/\textbf{2.96}/14.34$^{\dagger}$ & 85.67 & 2.62 & \cellcolor{blue!15}\textbf{26.40}$^{\dagger}$/\textbf{6.30}$^{\dagger}$/\textbf{18.83}$^{\dagger}$ & \cellcolor{blue!15}\textbf{84.92} & \cellcolor{blue!15}\textbf{4.37} & 27.93$^{\dagger}$/7.04/24.14$^{\dagger}$ & 86.24 & 6.09 \\
& SummScore - Salient-RL           & 35.54/14.05/32.04$^{\dagger}$ & 86.22 & 0.85 & 19.29$^{\dagger}$/\textbf{2.99}$^{\dagger}$/14.48$^{\dagger}$ & 85.79$^{\dagger}$ & 3.63 & \textbf{26.37}$^{\dagger}$/\textbf{6.32}$^{\dagger}$/\textbf{18.81}$^{\dagger}$ & \textbf{84.92} & \textbf{4.31} & 28.01$^{\dagger}$/7.08/24.21$^{\dagger}$ & 86.21 & 6.46 \\

\midrule

\multirow{7}{*}{\textbf{ChatGPT}} 
& First                            & 40.79/16.61/36.92 & \textbf{87.93} & \_ & \textbf{30.48}/10.00/\textbf{22.16} & 88.78 & \_ & 29.61/7.28/22.14 & \textbf{86.28} & \_ & 40.82/15.57/35.15 & 90.67 & \_ \\
& Random                           & 40.79/16.61/36.92 & \textbf{87.93} & 0.00 & \cellcolor{blue!15}\textbf{30.53}/\textbf{10.20}/\textbf{22.20} & \cellcolor{blue!15}88.77 & \cellcolor{blue!15}0.48 & 29.99/7.57/\textbf{22.32} & \textbf{86.32} & \_ & 40.60/15.28/34.78 & 90.63 & -0.95 \\

\cdashline{2-14}

& SummScore - Random-3             & 41.82$^{\dagger}$/\textbf{18.11}$^{\dagger}$/37.88$^{\dagger}$ & \textbf{87.91} & 3.69 & 27.98/8.45/19.64 & 87.94 & -10.49 & 30.09/7.85/22.16 & 86.15 & 1.78 & \cellcolor{blue!15}\textbf{42.73}$^{\dagger}$/\textbf{17.45}$^{\dagger}$/\textbf{37.63}$^{\dagger}$ & \cellcolor{blue!15}\textbf{90.93}$^{\dagger}$ & \cellcolor{blue!15}\textbf{6.86} \\
& SummScore - LEAD-3               & \cellcolor{blue!15}\textbf{42.05}$^{\dagger}$/\textbf{18.20}$^{\dagger}$/\textbf{38.06}$^{\dagger}$ & \cellcolor{blue!15}\textbf{87.97} & \cellcolor{blue!15}\textbf{4.23} & 27.97/8.42/19.76 & 88.05 & -10.34 & 30.14/7.78/\textbf{22.22} & 86.21 & 1.88 & 42.57$^{\dagger}$/17.29$^{\dagger}$/\textbf{37.54}$^{\dagger}$ & \textbf{90.88}$^{\dagger}$ & 6.41 \\
& SummScore - Salient-R1           & 40.30/17.10/36.37 & 87.67  & -0.57 & 27.84/8.46/19.55 & 87.91 & -10.87 & \textbf{30.29}/\textbf{7.97}$^{\dagger}$/22.20 & 86.12 & 2.41 & 42.59$^{\dagger}$/17.26$^{\dagger}$/37.50$^{\dagger}$ & \textbf{90.86}$^{\dagger}$ & 6.36 \\
& SummScore - Salient-R2           & 40.20/17.06/36.23 & 87.65 & -0.88 & 27.79/8.47/19.57 & 87.90 & -10.87 & \cellcolor{blue!15}\textbf{30.38}/\textbf{8.00}$^{\dagger}$/\textbf{22.27} & \cellcolor{blue!15}86.13 & \cellcolor{blue!15}\textbf{2.74} & 42.43$^{\dagger}$/17.00$^{\dagger}$/37.30$^{\dagger}$ & \textbf{90.84} & 5.67 \\
& SummScore - Salient-RL           & 40.24/17.06/36.29 & 87.66 & -0.76 & 27.82/8.51/19.58 & 87.90 & -10.73 & \textbf{30.29}/\textbf{7.97}$^{\dagger}$/22.20 & 86.12 & 2.39 & 42.59$^{\dagger}$/17.26$^{\dagger}$/37.50$^{\dagger}$ & \textbf{90.86}$^{\dagger}$ & 6.36 \\

\bottomrule

\end{tabular}
}
\vspace{-0.5em}
\caption{\small Unsupervised abstractive summarization results with SummScore re-ranking on the four datasets. Models are decoded to produce 20 summary candidates. \textbf{R-1/2/L} denotes {\sc{Rouge-1/2/L}} and \textbf{BS} denotes BERTScore. \textbf{Gain} represents the mean {\sc{Rouge}} relative gain compared to \emph{our top beam or first candidate baseline}. $^{\dagger}$ marks indicate significantly better results ($p$-value of paired t-test smaller than 0.05). Best results for each (backbone, dataset) pair within 0.1 are in bold.}
\label{tab:3b}
\vspace{-1.0em}
\end{table*}

\subsection{Unsupervised Abstractive Summarization}
\label{sec:4_2}

We first apply SummScore to unsupervised abstractive summarization, using as base model ($\gM_\text{base}$) two models of different capacity: the pre-traind PEGASUS \citep{zhang2020pegasus} (loading the \emph{google/pegasus-large} checkpoint from \emph{transformers}), and the recently introduced, highly-performing ChatGPT\footnote{https://chat.openai.com/. There is a chance that this checkpoint has been trained on the dataset above.}, accessed through OpenAI API (calling the \emph{gpt-3.5-turbo} checkpoint). Due to its pre-training objective of generating gap-sentences, PEGASUS can directly be applied to the summarization task after pre-training. This is not the case of comparable sequence-to-sequence Transformer-based models T5 \citep{raffel2019exploring} and BART \citep{lewis-etal-2020-bart}, which are pre-trained with token spans generation and sequence de-noising, respectively. For ChatGPT, to lower costs, we subsample randomly 1,000 data points from the test set on datasets other than SAMSum.

We decode PEGASUS with beam search, and ChatGPT with top-p sampling with $p=0.9$ and temperature 0.8 to enhance diversity, both models with 20 candidates. We report candidate selection baselines from \Cref{tab:1}: \emph{top beam} or \emph{first}, and \emph{random} (a randomly sampled candidate). 

We show unsupervised summarization results with PEGASUS and ChatGPT with 20 summary candidates in \Cref{tab:3b}. SummScore improves the base PEGASUS by 4.37\% to 7.27\% across the four datasets. Notably, SummScore fails with ChatGPT on XSum, which we hypothesize is due to the nature of XSum and the fact that pseudo-labels from XSum source documents are too different from the ground truth labels, an issue not affecting PEGASUS because its performance range is far lower than ChatGPT. However, SummScore improves ChatGPT by 2.74\% to 6.86\% on the other datasets. We point out that SummScore gains are achieved \emph{without using any human supervision}. 

SummScore - LEAD-3 performs best for the news domain, which intuitively makes sense due to the lead bias and first sentences containing an overview of the article.
On WikiHow, SummScore - Salient-R2 works the best, yet gains are more moderate and SummScore fails to improve the BERTScore on this dataset. SummScore - Random-3 is tied with SummScore - LEAD-3 on SAMSum: we attribute it to the fact that SAMSum source documents are very short (\Cref{tab:2}), and the LEAD-3, Random-3, and entire source document all overlap a lot. \Cref{sec:appendix_a} confirms that SummScore re-ranking always finds a non-trivial (e.g., longest) candidate selection.

\begin{table*}[]
\setlength{\tabcolsep}{2pt}
\resizebox{0.99\textwidth}{!}{

\begin{tabular}{lllcccccccccccc}

\toprule

\multirow{2}{*}{\textbf{\begin{tabular}[c]{@{}l@{}}Fine-tuning\\ dataset\end{tabular}}}
& \multirow{2}{*}{\textbf{\begin{tabular}[c]{@{}l@{}}Backbone\\ $\gM_\text{base}$\end{tabular}}}
& \multirow{2}{*}{\textbf{\begin{tabular}[c]{@{}l@{}}Candidate\\ Selection\end{tabular}}}
& \multicolumn{3}{c}{\textbf{CNN/DM}} 
& \multicolumn{3}{c}{\textbf{XSum}} 
& \multicolumn{3}{c}{\textbf{WikiHow}} 
& \multicolumn{3}{c}{\textbf{SAMSum}} \\
& & & \textbf{R-1}/\textbf{R-2}/\textbf{R-L} & \textbf{BS} & \textbf{Gain (\%)} 
& \textbf{R-1}/\textbf{R-2}/\textbf{R-L} & \textbf{BS} & \textbf{Gain (\%)} 
& \textbf{R-1}/\textbf{R-2}/\textbf{R-L} & \textbf{BS} & \textbf{Gain (\%)} 
& \textbf{R-1}/\textbf{R-2}/\textbf{R-L} & \textbf{BS} & \textbf{Gain (\%)} \\

\midrule

\multirow{6}{*}{\textbf{CNN/DM}} 

& \multirow{2}{*}{PEGASUS} 
& Top beam  & \cellcolor{gray!25} & \cellcolor{gray!25} & \cellcolor{gray!25} & 21.18/3.44/16.53 & 85.95 & \_ & 24.53/5.68/18.57 & 84.87 & \_ & 31.03/9.05/28.21 & 86.39 & \_ \\
& & SummScore & \cellcolor{gray!25} & \cellcolor{gray!25} & \cellcolor{gray!25} & 21.51$^{\dagger}$/3.49/16.69 & 86.05$^{\dagger}$ & 1.31 & 25.87$^{\dagger}$/6.04$^{\dagger}$/19.37$^{\dagger}$ & 84.94$^{\dagger}$ & 5.10 & 31.98/9.59/28.78 & 86.56 & 3.03 \\

\cdashline{2-15}

& \multirow{2}{*}{BART} 
& Top beam & \cellcolor{gray!25} & \cellcolor{gray!25} & \cellcolor{gray!25} & 20.32/3.10/15.95 & 86.03 & \_ & 26.13/6.03/19.69 & 85.18 & \_ & 30.78/9.60/28.28 & 86.81 & \_ \\
& & SummScore & \cellcolor{gray!25} & \cellcolor{gray!25} & \cellcolor{gray!25} & 20.61$^{\dagger}$/3.16/16.21$^{\dagger}$ & 86.27$^{\dagger}$ & 1.60 &  26.61$^{\dagger}$/6.24$^{\dagger}$/20.01$^{\dagger}$ & 85.24 & 1.97 & 30.77/9.56/28.20 & 86.87 & -0.02 \\

\cdashline{2-15}

& \multirow{2}{*}{BRIO} 
& Top beam & \cellcolor{gray!25} & \cellcolor{gray!25} & \cellcolor{gray!25} & 23.91/5.41/19.51 & 87.07 & \_ & 29.67/8.01/22.73 & \textbf{86.04} & \_ & 35.04/13.04/32.42 & \textbf{89.11} & \_ \\
& & SummScore &\cellcolor{gray!25} & \cellcolor{gray!25} & \cellcolor{gray!25} & 23.72/5.33/19.38 & 87.06 & -0.86 & \textbf{30.08}$^{\dagger}$/\textbf{8.17}/\textbf{23.01} & \textbf{86.05} & 1.39 & \textbf{35.50}/\textbf{13.35}/\textbf{32.85} & \textbf{89.09} & 1.50 \\

\midrule

\multirow{6}{*}{\textbf{XSum}} 

& \multirow{2}{*}{PEGASUS} 
& Top beam  & 23.10/8.03/20.18 & 85.88 & \_ & \cellcolor{gray!25} & \cellcolor{gray!25} & \cellcolor{gray!25} & 15.32/3.54/11.98 & 85.38 & \_ & 23.05/4.75/19.89 & 87.03 & \_  \\
& & SummScore & 26.60$^{\dagger}$/9.47$^{\dagger}$/23.13$^{\dagger}$ & 86.47$^{\dagger}$ & \textbf{15.38} &  \cellcolor{gray!25} & \cellcolor{gray!25} & \cellcolor{gray!25} & 19.36$^{\dagger}$/4.52$^{\dagger}$/14.27$^{\dagger}$ & 85.57$^{\dagger}$ & \textbf{23.73} & 26.82$^{\dagger}$/6.39$^{\dagger}$/22.91$^{\dagger}$ & 87.39 & \textbf{17.61} \\

\cdashline{2-15}

& \multirow{2}{*}{BART} 
& Top beam  & 25.60/8.10/22.16 & 86.37 & \_ & \cellcolor{gray!25} & \cellcolor{gray!25} & \cellcolor{gray!25} & 18.31/4.30/13.71 & 85.63 & \_ & 26.92/5.98/22.20 & 88.03 & \_  \\
& & SummScore & 27.80$^{\dagger}$/9.21$^{\dagger}$/23.97$^{\dagger}$ & 86.69$^{\dagger}$ & 9.18 &  \cellcolor{gray!25} & \cellcolor{gray!25} & \cellcolor{gray!25} & 20.52$^{\dagger}$/4.92$^{\dagger}$/14.94$^{\dagger}$ & 85.81$^{\dagger}$ & 11.24 & 30.03$^{\dagger}$/7.28$^{\dagger}$/24.71$^{\dagger}$ & 88.43$^{\dagger}$ & 12.52 \\

\cdashline{2-15}

& \multirow{2}{*}{BRIO} 
& Top beam  & 25.52/8.47/22.08 & 85.97 & \_ & \cellcolor{gray!25} & \cellcolor{gray!25} & \cellcolor{gray!25} & 18.39/4.24/13.82 & 85.58 & \_ & 26.69/5.19/22.02 & 87.16 & \_  \\
& & SummScore & 28.67$^{\dagger}$/9.82$^{\dagger}$/24.58$^{\dagger}$ & 86.42$^{\dagger}$ & 12.52 & \cellcolor{gray!25} & \cellcolor{gray!25} & \cellcolor{gray!25} & 21.94$^{\dagger}$/5.31$^{\dagger}$/15.75$^{\dagger}$ & 85.66$^{\dagger}$ & 17.94 & 30.10$^{\dagger}$/7.13$^{\dagger}$/24.90$^{\dagger}$ & 87.62$^{\dagger}$ & 15.25 \\

\midrule

\multirow{4}{*}{\textbf{WikiHow}} 

& \multirow{2}{*}{PEGASUS}     
& Top beam & 27.55/9.41/24.02 & 85.20 & \_ & 28.05/8.40/21.31 & 87.86 & \_ & \cellcolor{gray!25} & \cellcolor{gray!25} & \cellcolor{gray!25} & 21.15/3.92/17.46 & 85.44 & \_ \\
& & SummScore & 30.49$^{\dagger}$/10.97$^{\dagger}$/26.74$^{\dagger}$ & 85.95$^{\dagger}$ & 11.82 & 28.10/8.33/21.30 & 87.92 & -0.05 & \cellcolor{gray!25} & \cellcolor{gray!25} & \cellcolor{gray!25} & 23.62$^{\dagger}$/4.84$^{\dagger}$/19.26$^{\dagger}$ & 85.95 & 12.20 \\

\cdashline{2-15}

& \multirow{2}{*}{BART}                                          
& Top beam & 29.39/10.52/25.26 & 85.87 & \_ & 23.79/7.19/19.05 & 87.99 & \_ & \cellcolor{gray!25} & \cellcolor{gray!25} & \cellcolor{gray!25} & 19.51/4.52/17.29 & 87.07 & \_ \\
& & SummScore & 31.30$^{\dagger}$/11.42$^{\dagger}$/26.72$^{\dagger}$ & 86.21$^{\dagger}$ & 6.54 & 25.57$^{\dagger}$/7.54$^{\dagger}$/20.11$^{\dagger}$ & \textbf{88.18}$^{\dagger}$ & \textbf{6.41} & \cellcolor{gray!25} & \cellcolor{gray!25} & \cellcolor{gray!25} & 22.48$^{\dagger}$/5.40/19.63$^{\dagger}$ & 87.15 & 14.80 \\



\midrule

\multirow{4}{*}{\textbf{SAMSum}} 

& \multirow{2}{*}{PEGASUS}    
& Top beam & 36.40/15.48/32.52 & 87.16 & \_ & 24.30/6.31/18.75 & 87.41 & \_ & 22.17/5.10/16.29 & 85.08 & \_ & \cellcolor{gray!25} & \cellcolor{gray!25} & \cellcolor{gray!25} \\
& & SummScore & \textbf{39.15}$^{\dagger}$/16.89$^{\dagger}$/35.33$^{\dagger}$ & \textbf{87.48}$^{\dagger}$ & 8.27 & 24.10/5.67/18.69 & 87.31 & -1.52 & 24.44$^{\dagger}$/5.78$^{\dagger}$/18.03$^{\dagger}$ & 85.15 & 10.74 & \cellcolor{gray!25} & \cellcolor{gray!25} & \cellcolor{gray!25} \\

\cdashline{2-15}

& \multirow{2}{*}{BART}                                           
& Top beam & 38.40/16.58/35.22 & 86.93 & \_ & 20.78/3.70/15.42 & 86.49 & \_ & 26.00/6.29/19.63 & 84.73 & \_ & \cellcolor{gray!25} & \cellcolor{gray!25} & \cellcolor{gray!25} \\
& & SummScore & \textbf{39.24}$^{\dagger}$/17.07$^{\dagger}$/\textbf{35.94}$^{\dagger}$ & 87.11$^{\dagger}$ & 2.26 & 21.22$^{\dagger}$/3.71/15.79$^{\dagger}$ & 86.59$^{\dagger}$ & 2.03 & 26.35$^{\dagger}$/6.43/19.91$^{\dagger}$ & 84.75 & 1.44 & \cellcolor{gray!25} & \cellcolor{gray!25} & \cellcolor{gray!25} \\



\midrule



\multicolumn{2}{l}{\textbf{WikiTransfer}* } 

& Top beam & 39.11/\textbf{17.25}/35.73 & \_ & \_ & \textbf{31.85}/\textbf{10.44}/\textbf{23.75} & \_ & \_ & \_ & \_ & \_ & \_ & \_ & \_ \\



\bottomrule
          
\end{tabular}
}
\caption{\small Zero-shot transfer results with SummScore re-ranking, across all twelve transfer directions over the four summarization datasets. Each model is decoded with beam search with 20 beams. \textbf{Top beam} refers to the base model performance, while \textbf{SummScore} is the candidate re-ranked by SummScore. \textbf{R-1/2/L} is {\sc{Rouge-1/2/L}}, \textbf{BS} denotes BERTScore, and \textbf{Gain (\%)} is the relative mean {\sc{Rouge}} improvement compared to the base model performance. $^{\dagger}$ marks indicate significantly better results ($p$-value of paired t-test smaller than 0.05). {Best results within 0.1 are in bold.} Greyed out cells correspond to the supervised setup, which is excluded. *WikiTransfer \cite{fabbri-etal-2021-improving} is not directly comparable due to constructing the fine-tuning dataset specifically to optimize transfer to the downstream task.}
\label{tab:4}
\vspace{-1.0em}
\end{table*}







\vspace{-0.5em}
\subsection{Zero-Shot Transfer}

Next, we investigate SummScore performance in the transfer setup, with standard-size models (discarding ChatGPT or similar models). We perform zero-shot summarization inference followed by SummScore on a target dataset where the base model $\gM_\text{base}$ was fine-tuned on \emph{another} source dataset. 
As $\gM_\text{base}$, we use three high-performing summarization models: PEGASUS \citep{zhang2020pegasus}, BART \citep{lewis-etal-2020-bart}, and the recently introduced BRIO \citep{liu-etal-2022-brio}, which achieves SOTA results on news summarization (CNN/DM \& XSum). We use publicly available fine-tuned checkpoints on CNN/DM and XSum, and PEGASUS on WikiHow. We fine-tune ourselves PEGASUS on SAMSum, and BART on WikiHow and SAMSum. Generation and fine-tuning hyper-parameters and results are in \Cref{sec:appendix_b}. 

Given the findings from \Cref{sec:4_2}, we use SummScore - LEAD-3 on CNN/DM, XSum, and SAMSum, and SummScore - Salient-R2 on WikiHow. We tune coefficients in the same process described in \Cref{subsec:setup}. To stick to a \textbf{no supervision} scenario, we do not apply SummScore on a dataset on the which the base model was fine-tuned, which would fall into the supervised learning use case. We compare SummScore zero-shot transfer performance on CNN/DM with that of SOTA WikiTransfer \citep{fabbri-etal-2021-improving}, which fine-tunes BART on external data retrieved from Wikipedia before applying the model in zero-shot summarization.

Zero-shot transfer results are displayed in \Cref{tab:4}. SummScore consistently improves transfer performance, with {\sc{Rouge}} gains of 7.51\% averaged over 30 setups: +9.43\% on CNN/DM, +1.27\% on XSum, +9.20\% on WikiHow (up to +17.64\% average when transferring from XSum) and +9.61\% on SAMSum. Notably, on CNN/DM, BART transferred from SAMSum with SummScore improves on the {\sc{Rouge-1}} and {\sc{Rouge-L}} of SOTA transfer model WikiTransfer (also using a BART backbone), despite WikiTransfer being fine-tuned on data specifically crafted to transfer better to the downstream task. We notice that SummScore helps more when the base model transfers less well, such as from single-sentence summaries XSum. 

\Cref{sec:appendix_c} evalutes re-ranking itself and shows that SummScore can also reach strong recall.

\subsection{Self-Training with Unsupervised Paraphrasing}

Using the selected summary candidate as a pseudo-target, one can naturally extend SummScore into a self-training summarization objective. Indeed, if $\gamma$ parametrizes $\gM_\text{base}$, we can further train $\gM_\text{base}$ through the objective: 


\vspace{-1.0em}

\begin{equation}
\small 
    \tilde{\gamma} = \arg \max_{\gamma} \sum_{i} \log \big( p(\text{SummScore}(\sC_{i}) | x_{i}; \gamma ) \big)
\end{equation}

\vspace{-0.5em}

This process can be repeated: if we denote new model weights by $\gamma^{k}$, we can re-apply SummScore and perform another round of self-training, yielding new model weights $\gamma^{k+1}$.

\begin{table}[h]
\resizebox{0.99\columnwidth}{!}{
\begin{tabular}{llcccc}
\toprule
\textbf{Dataset} 
& \textbf{Model}            
& \textbf{R-1} 
& \textbf{R-2} 
& \textbf{R-L} 
& \textbf{BS} \\

\midrule

\multirow{12}{*}{\textbf{CNN/DM}}    
& PEGASUS \cite{zhang2020pegasus}                 & 32.90 & 13.28 & 29.38 & \_ \\
& Summary Loop 45 \cite{laban-etal-2020-summary}  & 37.70 & 14.80 & 34.70 & \_ \\
& TED \cite{yang-etal-2020-ted}                   & 38.73 & 16.84 & 35.40 & \_ \\
& FAR-RW* \cite{zhang2022unsupervised} (SOTA)     & \textbf{40.13} & \textbf{17.00} & \textbf{36.34} & \_ \\

\cdashline{2-6}

& PEGASUS (ours)                                  & 35.47 & 13.89 & 31.61 & 86.29 \\
& PEGASUS (ours) + SummScore                      & 36.92 & 15.03 & 33.19 & 86.54 \\
& Self-training ($1^{st}$ round)                  & 36.68 & 14.52 & 32.72 & 86.49 \\
& Self-training ($1^{st}$ round) + SummScore      & 38.75 & 16.11 & 34.78 & 86.88 \\
& Self-training ($2^{nd}$ round)                  & 38.17 & 15.77 & 34.25 & 86.87 \\
& Self-training ($2^{nd}$ round) + SummScore      & 39.49 & 16.69 & 35.61 & 87.07 \\
& Self-training ($3^{rd}$ round)                  & 38.47 & 15.95 & 34.48 & 87.00 \\
& Self-training ($3^{rd}$ round) + SummScore      & 39.76 & 16.79 & 35.85 & \textbf{87.18} \\
                        
\midrule
                        
\multirow{4}{*}{\textbf{XSum}}      
& PEGASUS (ours)                             & 18.77 & 2.86 & 13.85 & 85.66 \\
& PEGASUS (ours) + SummScore                 & 19.62 & \textbf{3.02} & 14.71 & 85.92 \\
& Self-training                              & 19.33 & 2.76 & 14.18 & 86.03 \\
& Self-training + SummScore                  & \textbf{20.02} & 2.84 & \textbf{14.93} & \textbf{86.23} \\
                        
\midrule
                        
\multirow{4}{*}{\textbf{WikiHow}}   
& PEGASUS (ours)                             & 25.49 & 5.91 & 17.99 & \textbf{84.98} \\
& PEGASUS (ours) + SummScore                 & \textbf{26.40} & \textbf{6.30} & 18.83 & \textbf{84.92} \\
& Self-training                              & 26.08 & 6.08 & 18.59 & 84.89 \\
& Self-training + SummScore                  & \textbf{26.50} & \textbf{6.28} & \textbf{19.03} & \textbf{84.93} \\
                        
\midrule
                        
\multirow{4}{*}{\textbf{SAMSum}}    
& PEGASUS (ours)                             & 26.64 & 6.32 & 22.75 & 86.12 \\
& PEGASUS (ours) + SummScore                 & 28.22 & 7.16 & 24.39 & 86.41 \\
& Self-training                              & 26.96 & 6.41 & 23.40 & 86.25 \\
& Self-training + SummScore                  & \textbf{28.91} & \textbf{7.55} & \textbf{25.54} & \textbf{86.58} \\

\bottomrule

\end{tabular}
}
\caption{\small Unsupervised abstractive summarization results with SummScore re-ranking and \emph{self-training} for PEGASUS on the four datasets. We fine-tune the model with the unsupervised summary candidate which was selected by SummScore as pseudo-target, then apply again SummScore on the output. All models are decoded with beam search with 20 beams. \textbf{R-1/2/L} is {\sc{Rouge-1/2/L}}, and \textbf{BS} denotes BERTScore. Best results within 0.1 are in bold. *FAR-RW pipeline is not directly comparable due to relying on a SOTA unsupervised extractive summarization model first, then applying re-writing.}
\label{tab:5}
\end{table}

We notice that the unsupervised PEGASUS beam search summary candidates, including the one selected by SummScore, are quite extractive (see \Cref{sec:appendix_d}). This could be because the self-supervised gap-sentences are extracts from the source document. To make the pseudo-summaries more abstractive and diverse enough to mitigate the confirmation bias in self-training \citep{10.5555/3294771.3294885}, we use the paraphrasing approach proposed in FAR-RW \citep{zhang2022unsupervised}. On each dataset, we train a paraphrase model to generate the top $n$ sentences maximizing the mean {\sc{Rouge}} with the top $n$ most salient sentences, conditioning on these salient sentences. This yields an unsupervised, in-domain paraphrase model which we apply to the SummScore pseudo-labels on the training set to make them more abstractive and diverse. We refer to \Cref{sec:appendix_e} for details on the paraphrasing model training, its performance and resulting abstractiveness and diversity levels on pseudo-labels. As the unsupervised process of paraphrasing may harm the pseudo-summary quality, in practice, we apply it to the $x$\% most extractive training data points, where $x$ is among \{12.5\%, 25\%, 50\%, 100\%\}. We use 25\% for CNN/DM, 100\% for XSum, 50\% for WikiHow, and 12.5\% on SAMSum, as these provide an ideal ROUGE/abstractiveness trade-off (see \Cref{sec:appendix_d}).


For each dataset except SAMSum, we randomly subsample 50k data points from the training set and 1k from the validation set to self-train and validate the model, resulting in a self-training process much less computationally expensive than fine-tuning. We show self-training results on the test sets using PEGASUS as base model in \Cref{tab:5}. Self-training improves unsupervised summarization performance on all datasets, resulting in a self-trained model better than the base model although not as performing as SummScore. Notably, re-applying SummScore on the new model after self-training further improves performance drastically. Besides, paraphrasing self-training pseudo-labels helps maintain some degree of abstractiveness, as seen in \Cref{sec:appendix_d}. On CNN/DM, one round of self-training followed by SummScore brings PEGASUS performance above the Summary Loop, two rounds above TED, and three rounds to 39.76 {\sc{Rouge-1}}, within 1\% of SOTA model FAR-RW. 

\begin{table}[t]
\resizebox{0.99\columnwidth}{!}{
\begin{tabular}{llcccc}
\toprule
\textbf{Use case} 
& \textbf{Attribute} 
& \textbf{PEGASUS} 
& \textbf{SummScore} 
& \textbf{Tie} \\

\midrule

\multirow{2}{*}{\textbf{Unsupervised abs. summ.}}   
& Informativeness & 11.33 (1.15) & \textbf{20.67} (6.43) & 18.00 (6.93) \\  
& Factual consistency & 14.67 (4.04) & \textbf{19.33} (5.03) & 16.00	(9.00) \\  

\midrule

\multirow{2}{*}{\textbf{0-shot transfer from XSum}}          
& Informativeness & 5.67 (2.89) & \textbf{24.00} (2.00) & 20.33 (1.53) \\
& Factual consistency & 4.67 (4.51) & 18.67	(4.04) & \textbf{26.67} (3.51) \\  

\bottomrule
\end{tabular}
}
\caption{\small Human evaluation on CNN/DM with PEGASUS. Mean number of times out of 50 that each model or a tie is selected, with standard deviation in parenthesis, across two use cases and two attributes.}
\label{tab:6}
\vspace{-1.0em}
\end{table}

\vspace{-0.5em}
\subsection{Human Evaluation}

We conduct a human evaluation on 50 data points randomly sampled from CNN/DM test set. We show human participants the source news article, alongside the summary candidate from the base PEGASUS model, and the one re-ranked by SummScore. Participants are asked to pick which summary is more informative, and which is more factually consisteny, with the option of choosing a tie. We cover two use cases: unsupervised abstractive summarization, and zero-shot transfer from a model fine-tuned on XSum. In the former use case, both summaries are identical in 7/50 data points, and 4/50 data points in the latter. Human raters are three volunteer graduate students, with full professional command of English. Results are displayed in \Cref{tab:6}. Although both summaries often overlap significantly (rightmost column), resulting in a high \emph{Tie}, SummScore is strongly preferred over PEGASUS across both use cases and attributes. 

\begin{table}[]
\resizebox{0.99\columnwidth}{!}{
\begin{tabular}{lccccc}
\toprule 

\multirow{2}{*}{\textbf{Candidate selection}} 
& \multicolumn{4}{c}{\textbf{Dataset}} 
& \multirow{2}{*}{\textbf{Average}} \\
& \textbf{CNN/DM} 
& \textbf{XSum} 
& \textbf{WikiHow} 
& \textbf{SAMSum} & \\

\midrule 
PEGASUS               & 26.99 & 11.83 & 16.46 & 18.57 & 18.46 \\

\midrule

{\sc{Rouge-1}} with source   & 26.90 & 12.03 & \textbf{17.21} & \textbf{19.89} & 19.01 \\
{\sc{Rouge-2}} with source   & 26.98 & 11.93 & \textbf{17.16} & 19.62 & 18.92 \\
{\sc{Bleu}} with source      & 26.90 & 11.99 & \textbf{17.19} & \textbf{19.94} & 19.01 \\

\cdashline{1-6}

BERTScore with source & 28.19 & \textbf{12.42} & \textbf{17.11} & 19.43 & 19.29 \\
BARTScore with source & 28.11 & 12.23 & 16.60 & 19.70 & 19.16 \\
BLEURT with source    & 27.45 & 12.12 & 16.79 & 19.69 & 19.01 \\

\cdashline{1-6}

Diversity score       & 25.33 & 11.36 & 14.52 & 15.67 & 16.72 \\
Length score          & 27.07 & 11.67 & 16.66 & 18.60 & 18.50 \\

\midrule 

Plain average         & 27.75 & 12.28 & 16.96 & 19.73 & 19.18 \\
Random coefficients   & 27.75 & 12.25 & 16.84 & 19.72 & 19.14 \\

\cdashline{1-6}

\textbf{SummScore}    & \textbf{28.38} & \textbf{12.45} & \textbf{17.18} & \textbf{19.92} & \textbf{19.48} \\

\bottomrule
\end{tabular}
}
\caption{\small Ablation study for unsupervised abstractive summarization with PEGASUS. We isolate each feature of SummScore and report its re-ranking performance (picking the candidate maximizing this feature), using the mean of {\sc{Rouge-1/2/L}} as reported metric. Best results within 0.1 are in bold.}
\label{tab:7}
\vspace{-1.0em}
\end{table}

\vspace{-0.5em}
\section{Analysis}
\label{sec:analysis}

\vspace{-0.5em}
\subsection{Ablation}

To better understand SummScore performance gains, we perform an ablation study where re-ranking is done with each feature taken individually. Results for PEGASUS in unsupervised summarization are shown in \Cref{tab:7}. N-gram overlap features are very strong re-ranking baselines on WikiHow and SAMSum. In fact, {\sc{Rouge-1}} with the source is even slightly better than SummScore on WikiHow. On news datasets, semantic similarity features such as BERTScore are strong baselines. Interestingly, our hand-crafted feature \emph{diversity} has a \emph{negative} contribution when used as standalone re-ranker ; however it can help a lot when combined with the other features, acting as a regularizer by encouraging some diversity. On average, SummScore performs the best. We also report trivial feature aggregation baselines \emph{Plain average} and \emph{Random coefficients}, which SummScore outpeforms, confirming the efficiency of estimating coefficients through pseudo-labels.

In \Cref{sec:appendix_f}, we show that SummScore unsupervised re-ranking is also robust to other decoding methods diverse beam search \citep{vijayakumar2016diverse} and nucleus sampling \citep{holtzman2019curious}, and a different number of beams (5 to 20). We confirm that our default setup of beam search with 20 beams yields optimal {\sc{Rouge}} results. Echoing SummaReranker findings \citep{ravaut-etal-2022-summareranker}, gains further increase when mixing in several decoding methods.

\subsection{Qualitative Samples}

We refer to \Cref{sec:appendix_h} for full qualitative unsupervised re-ranking examples on all datasets, and to \Cref{tab:8} for an example of summary generated by the self-trained PEGASUS model on XSum. As seen, both re-ranking and self-training can improve dramatically from the unsupervised PEGASUS baseline, capturing entirely new phrases.

\subsection{Factual Consistency}

As noted in \Cref{tab:6}, SummScore summaries tend to be more factually consistent than the baseline. There is strong intuition to this result: since SummScore is built to maximize features of n-gram overlap and semantic similarity with the source, it should yield summaries closer to the source, and more factually consistent as a result. We investigate this further, and use two popular models to evaluate summarization factuality: the established \emph{factCC} \citep{kryscinski-etal-2020-evaluating} and the recently introduced state-of-the-art \emph{QAFactEval} \citep{fabbri-etal-2022-qafacteval}. factCC uses a BERT model to classify each summary sentence as consistent or inconsistent with regards to the source, and reports the average accuracy over 100\%. QAFactEval improves each step of the QA evaluation pipeline (answer selection, question generation, etc) and combines entailment with QA-based metrics into a learned metric. In \Cref{tab:9}, we observe that SummScore QAFactEval is consistently above PEGASUS, and SummScore factCC is better on news datasets too.

\subsection{Learned Coefficients}

\begin{table}[t]
\resizebox{0.99\columnwidth}{!}{
\begin{tabular}{l}

\toprule

\textbf{Source document}:\\ 
\parbox{2.0\columnwidth}{Reports speak of at least four people injured. The city is at the heart of the conflict between the Turkish government and Kurdish separatists. Interior Minister Suleyman Soylu said the blast happened at a vehicle repair unit, and appeared to be an accident. He said "it seems there is no outside interference, and the explosion came from the vehicle under repair". Mr Soylu said one person was trapped under rubble, another was seriously injured, and others had minor injuries. The blast brought a roof down, left a huge crater and a pall of smoke drifted over part of the city. The cause remains unclear. The banned Kurdistan Workers' Party (PKK) is active in the area. Turkey is five days away from a key referendum on granting President Recep Tayyip Erdogan sweeping new powers [...]} \\

\midrule

\textbf{PEGASUS summary ({\sc{Rouge-1}}: 10.53)}:\\ 
\parbox{2.0\columnwidth}{Interior Minister Suleyman Soylu said \orange{the blast happened at a vehicle repair unit}, and appeared to be an accident.} \\

\midrule

\textbf{Self-training summary ({\sc{Rouge-1}}: 32.43)}:\\ 
\parbox{2.0\columnwidth}{\orange{The blast happened at a vehicle repair unit} in the \teal{city of Diyarbakir}, \blue{near the border with Syria}.} \\

\midrule

\textbf{Ground truth summary}:\\ 
\parbox{2.0\columnwidth}{A large explosion has struck a police headquarters in the mainly Kurdish \teal{city of Diyarbakir} \blue{in south-eastern Turkey}.}\\

\bottomrule

\end{tabular}
}
\caption{\small Qualitative sample with self-training PEGASUS from the XSum dataset, after a single round of self-training.}
\label{tab:8}
\vspace{-1.0em}
\end{table}

We analyze coefficients learned by SummScore from a high level perspective in \Cref{tab:10}, gathering features from a same group together. Semantic similarity features are dominating (except for WikiHow), encouraging further research using newer semantic similarity metrics for re-ranking. 

A finer-grain analysis, covering all SummScore pseudo-labeling techniques, can be viewed in \Cref{tab:g1,tab:g2} of \Cref{sec:appendix_g}. SummScore - Salient-R1 and SummScore - Salient-RL place much more emphasis on n-gram overlap with the source. In contrast, SummScore - LEAD-3 (which we use for self-training on CNN/DM, XSum, SAMSum) uses relatively more semantic similarity features like BERTScore, suggesting that it is able to exploit key semantic content contained in initial sentences.

\begin{table}[]
\resizebox{0.99\columnwidth}{!}{
\begin{tabular}{llcc}

\toprule

\textbf{Dataset} & \textbf{Factual consistency model} & \textbf{PEGASUS} & \textbf{SummScore} \\

\midrule 

\multirow{2}{*}{\textbf{CNN/DM}}           
& factCC         & 92.45 & \textbf{93.66} \\
& QAFactEval     & 4.53  & \textbf{4.55} \\

\midrule 

\multirow{2}{*}{\textbf{XSum}}         
& factCC         & 96.78 & \textbf{97.53} \\
& QAFactEval     & 4.54  & \textbf{4.64} \\

\midrule 

\multirow{2}{*}{\textbf{WikiHow}}       
& factCC         & \textbf{96.48} & 95.85 \\
& QAFactEval     & 4.33 & \textbf{4.36} \\

\midrule 

\multirow{2}{*}{\textbf{SAMSum}}          
& factCC         & \textbf{98.35} & 96.28 \\
& QAFactEval     & 3.26 & \textbf{3.50}    \\

\bottomrule
\end{tabular}
}
\caption{Factual consistency evaluation of SummScore with PEGASUS in unsupervised abstractive summarization. We use the entire test set for \emph{factCC}, and a random sample of 500 test data points for \emph{QAFactEval}.}
\label{tab:9}
\end{table}

\begin{table}[]
\resizebox{0.99\columnwidth}{!}{
\begin{tabular}{lcccccc}
\toprule 

\multirow{2}{*}{\textbf{Dataset}}   &  \multicolumn{3}{c}{\textbf{PEGASUS}}  & \multicolumn{3}{c}{\textbf{ChatGPT}}  \\

& \textbf{N-gram} & \textbf{Semantic} & \textbf{Quality} & \textbf{N-gram} & \textbf{Semantic} & \textbf{Quality}   \\

\midrule 

CNN/DM   & 0.025 & \textbf{0.900} & 0.075 & 0.100 & \textbf{0.775} & 0.125 \\
XSum     & 0.050 & \textbf{0.950} & 0.000 & 0.250 & \textbf{0.725} & 0.025 \\
WikiHow  & \textbf{0.875} & 0.100 & 0.025 & \textbf{0.900} & 0.100 & 0.000 \\
SAMSum   & 0.000 & \textbf{1.000} & 0.000 & 0.000 & \textbf{1.000} & 0.000 \\

\bottomrule

\end{tabular}
}
\caption{Coefficients learned by SummScore in unsupervised abstractive summarization. We sum weights assigned to all features of each category defined in \Cref{subsec:3_2}.}
\label{tab:10}
\vspace{-1.0em}
\end{table}

\vspace{-0.5em}
\section{Conclusion}
\label{sec:conclusion}

We introduced SummScore, the first unsupervised abstractive summarization re-ranking system. SummScore does not rely on a neural network: instead, it builds features for each summary candidate, some of them using the source as well, and aggregates them into a final re-ranking score. Feature coefficients are estimated through tuning against a pseudo-label derived from the source document. It is a simple framework which easily supports the addition of new features. 

SummScore significantly improves the performance of the base summarization model, in terms of {\sc{Rouge}}, BERTScore, factual consistency, and human preference ; in both unsupervised and zero-shot transfer scenarios. Moreover, SummScore selected summary candidate naturally extends into a self-training objective for abstractive summarization, which improves unsupervised summarization.




\section*{Limitations}
As a second-stage method, SummScore requires access to a base abstractive summarization model generating summary candidates. Generating up to 20 summary candidates per data point can take a long time, especially on training sets, which is needed for the self-training use case. Besides, even though SummScore does not need to train a new neural network, we also need to generate all eight features for each summary candidate once all candidates are generated. N-gram overlap features are very fast, but model-based semantic similarity features (e.g, BERTScore) can be time-consuming to extract, once again, especially on entire training sets. 

While SummScore will significantly improve the quality of the base model across base models and datasets, ultimately, the performance of the final selected summary is bounded by the capacity of this base model: SummScore improves more PEGASUS than it does on ChatGPT ; but PEGASUS performance drags ChatGPT. 

Another limitation lays in the metric used to compare summary candidates with the pseudo-target. We used mean {\sc{Rouge}}, although a model-based semantic similarity metric would make sense too, but at a much greater computational cost.


\section*{Acknowledgements}

This research was supported by the SINGA schol-
arship and partially supported by the National Re-
search Foundation, Prime Minister’s Office, Singa-
pore under its Campus for Research Excellence and
Technological Enterprise (CREATE) programme. We thank anonymous reviewers for a fruitful discussion, especially with regards to evaluation of the factual consistency. We also thank Florian Le Bronnec and Jiajing Zhang for their proofreading.

\bibliography{anthology,custom}

\begin{thebibliography}{41}
\expandafter\ifx\csname natexlab\endcsname\relax\def\natexlab#1{#1}\fi

\bibitem[{Baziotis et~al.(2019)Baziotis, Androutsopoulos, Konstas, and
  Potamianos}]{baziotis-etal-2019-seq}
Christos Baziotis, Ion Androutsopoulos, Ioannis Konstas, and Alexandros
  Potamianos. 2019.
\newblock \href {https://doi.org/10.18653/v1/N19-1071} {{SEQ}{\^{}}3:
  Differentiable sequence-to-sequence-to-sequence autoencoder for unsupervised
  abstractive sentence compression}.
\newblock In \emph{Proceedings of the 2019 Conference of the North {A}merican
  Chapter of the Association for Computational Linguistics: Human Language
  Technologies, Volume 1 (Long and Short Papers)}, pages 673--681, Minneapolis,
  Minnesota. Association for Computational Linguistics.

\bibitem[{Devlin et~al.(2019)Devlin, Chang, Lee, and
  Toutanova}]{devlin-etal-2019-bert}
Jacob Devlin, Ming-Wei Chang, Kenton Lee, and Kristina Toutanova. 2019.
\newblock \href {https://doi.org/10.18653/v1/N19-1423} {{BERT}: Pre-training of
  deep bidirectional transformers for language understanding}.
\newblock In \emph{Proceedings of the 2019 Conference of the North {A}merican
  Chapter of the Association for Computational Linguistics: Human Language
  Technologies, Volume 1 (Long and Short Papers)}, pages 4171--4186,
  Minneapolis, Minnesota. Association for Computational Linguistics.

\bibitem[{Fabbri et~al.(2021)Fabbri, Han, Li, Li, Ghazvininejad, Joty, Radev,
  and Mehdad}]{fabbri-etal-2021-improving}
Alexander Fabbri, Simeng Han, Haoyuan Li, Haoran Li, Marjan Ghazvininejad,
  Shafiq Joty, Dragomir Radev, and Yashar Mehdad. 2021.
\newblock \href {https://doi.org/10.18653/v1/2021.naacl-main.57} {Improving
  zero and few-shot abstractive summarization with intermediate fine-tuning and
  data augmentation}.
\newblock In \emph{Proceedings of the 2021 Conference of the North American
  Chapter of the Association for Computational Linguistics: Human Language
  Technologies}, pages 704--717, Online. Association for Computational
  Linguistics.

\bibitem[{Fabbri et~al.(2022)Fabbri, Wu, Liu, and
  Xiong}]{fabbri-etal-2022-qafacteval}
Alexander Fabbri, Chien-Sheng Wu, Wenhao Liu, and Caiming Xiong. 2022.
\newblock \href {https://doi.org/10.18653/v1/2022.naacl-main.187}
  {{QAF}act{E}val: Improved {QA}-based factual consistency evaluation for
  summarization}.
\newblock In \emph{Proceedings of the 2022 Conference of the North American
  Chapter of the Association for Computational Linguistics: Human Language
  Technologies}, pages 2587--2601, Seattle, United States. Association for
  Computational Linguistics.

\bibitem[{Gao et~al.(2020)Gao, Zhao, and Eger}]{gao-etal-2020-supert}
Yang Gao, Wei Zhao, and Steffen Eger. 2020.
\newblock \href {https://doi.org/10.18653/v1/2020.acl-main.124} {{SUPERT}:
  Towards new frontiers in unsupervised evaluation metrics for multi-document
  summarization}.
\newblock In \emph{Proceedings of the 58th Annual Meeting of the Association
  for Computational Linguistics}, pages 1347--1354, Online. Association for
  Computational Linguistics.

\bibitem[{Gliwa et~al.(2019)Gliwa, Mochol, Biesek, and
  Wawer}]{gliwa-etal-2019-samsum}
Bogdan Gliwa, Iwona Mochol, Maciej Biesek, and Aleksander Wawer. 2019.
\newblock \href {https://doi.org/10.18653/v1/D19-5409} {{SAMS}um corpus: A
  human-annotated dialogue dataset for abstractive summarization}.
\newblock In \emph{Proceedings of the 2nd Workshop on New Frontiers in
  Summarization}, pages 70--79, Hong Kong, China. Association for Computational
  Linguistics.

\bibitem[{Goyal et~al.(2022)Goyal, Li, and Durrett}]{goyal2022news}
Tanya Goyal, Junyi~Jessy Li, and Greg Durrett. 2022.
\newblock News summarization and evaluation in the era of gpt-3.
\newblock \emph{arXiv preprint arXiv:2209.12356}.

\bibitem[{Hermann et~al.(2015)Hermann, Kocisky, Grefenstette, Espeholt, Kay,
  Suleyman, and Blunsom}]{hermann2015teaching}
Karl~Moritz Hermann, Tomas Kocisky, Edward Grefenstette, Lasse Espeholt, Will
  Kay, Mustafa Suleyman, and Phil Blunsom. 2015.
\newblock Teaching machines to read and comprehend.
\newblock \emph{Advances in neural information processing systems}, 28.

\bibitem[{Holtzman et~al.(2019)Holtzman, Buys, Du, Forbes, and
  Choi}]{holtzman2019curious}
Ari Holtzman, Jan Buys, Li~Du, Maxwell Forbes, and Yejin Choi. 2019.
\newblock The curious case of neural text degeneration.
\newblock \emph{arXiv preprint arXiv:1904.09751}.

\bibitem[{Koupaee and Wang(2018)}]{koupaee2018wikihow}
Mahnaz Koupaee and William~Yang Wang. 2018.
\newblock Wikihow: A large scale text summarization dataset.
\newblock \emph{arXiv preprint arXiv:1810.09305}.

\bibitem[{Kryscinski et~al.(2019)Kryscinski, Keskar, McCann, Xiong, and
  Socher}]{kryscinski-etal-2019-neural}
Wojciech Kryscinski, Nitish~Shirish Keskar, Bryan McCann, Caiming Xiong, and
  Richard Socher. 2019.
\newblock \href {https://doi.org/10.18653/v1/D19-1051} {Neural text
  summarization: A critical evaluation}.
\newblock In \emph{Proceedings of the 2019 Conference on Empirical Methods in
  Natural Language Processing and the 9th International Joint Conference on
  Natural Language Processing (EMNLP-IJCNLP)}, pages 540--551, Hong Kong,
  China. Association for Computational Linguistics.

\bibitem[{Kryscinski et~al.(2020)Kryscinski, McCann, Xiong, and
  Socher}]{kryscinski-etal-2020-evaluating}
Wojciech Kryscinski, Bryan McCann, Caiming Xiong, and Richard Socher. 2020.
\newblock \href {https://doi.org/10.18653/v1/2020.emnlp-main.750} {Evaluating
  the factual consistency of abstractive text summarization}.
\newblock In \emph{Proceedings of the 2020 Conference on Empirical Methods in
  Natural Language Processing (EMNLP)}, pages 9332--9346, Online. Association
  for Computational Linguistics.

\bibitem[{Kry{\'s}ci{\'n}ski et~al.(2018)Kry{\'s}ci{\'n}ski, Paulus, Xiong, and
  Socher}]{kryscinski-etal-2018-improving}
Wojciech Kry{\'s}ci{\'n}ski, Romain Paulus, Caiming Xiong, and Richard Socher.
  2018.
\newblock \href {https://doi.org/10.18653/v1/D18-1207} {Improving abstraction
  in text summarization}.
\newblock In \emph{Proceedings of the 2018 Conference on Empirical Methods in
  Natural Language Processing}, pages 1808--1817, Brussels, Belgium.
  Association for Computational Linguistics.

\bibitem[{Laban et~al.(2020)Laban, Hsi, Canny, and
  Hearst}]{laban-etal-2020-summary}
Philippe Laban, Andrew Hsi, John Canny, and Marti~A. Hearst. 2020.
\newblock \href {https://doi.org/10.18653/v1/2020.acl-main.460} {The summary
  loop: Learning to write abstractive summaries without examples}.
\newblock In \emph{Proceedings of the 58th Annual Meeting of the Association
  for Computational Linguistics}, pages 5135--5150, Online. Association for
  Computational Linguistics.

\bibitem[{Lewis et~al.(2020)Lewis, Liu, Goyal, Ghazvininejad, Mohamed, Levy,
  Stoyanov, and Zettlemoyer}]{lewis-etal-2020-bart}
Mike Lewis, Yinhan Liu, Naman Goyal, Marjan Ghazvininejad, Abdelrahman Mohamed,
  Omer Levy, Veselin Stoyanov, and Luke Zettlemoyer. 2020.
\newblock \href {https://doi.org/10.18653/v1/2020.acl-main.703} {{BART}:
  Denoising sequence-to-sequence pre-training for natural language generation,
  translation, and comprehension}.
\newblock In \emph{Proceedings of the 58th Annual Meeting of the Association
  for Computational Linguistics}, pages 7871--7880, Online. Association for
  Computational Linguistics.

\bibitem[{Lhoest et~al.(2021)Lhoest, Villanova~del Moral, Jernite, Thakur, von
  Platen, Patil, Chaumond, Drame, Plu, Tunstall, Davison, {\v{S}}a{\v{s}}ko,
  Chhablani, Malik, Brandeis, Le~Scao, Sanh, Xu, Patry, McMillan-Major, Schmid,
  Gugger, Delangue, Matussi{\`e}re, Debut, Bekman, Cistac, Goehringer, Mustar,
  Lagunas, Rush, and Wolf}]{lhoest-etal-2021-datasets}
Quentin Lhoest, Albert Villanova~del Moral, Yacine Jernite, Abhishek Thakur,
  Patrick von Platen, Suraj Patil, Julien Chaumond, Mariama Drame, Julien Plu,
  Lewis Tunstall, Joe Davison, Mario {\v{S}}a{\v{s}}ko, Gunjan Chhablani,
  Bhavitvya Malik, Simon Brandeis, Teven Le~Scao, Victor Sanh, Canwen Xu,
  Nicolas Patry, Angelina McMillan-Major, Philipp Schmid, Sylvain Gugger,
  Cl{\'e}ment Delangue, Th{\'e}o Matussi{\`e}re, Lysandre Debut, Stas Bekman,
  Pierric Cistac, Thibault Goehringer, Victor Mustar, Fran{\c{c}}ois Lagunas,
  Alexander Rush, and Thomas Wolf. 2021.
\newblock \href {https://doi.org/10.18653/v1/2021.emnlp-demo.21} {Datasets: A
  community library for natural language processing}.
\newblock In \emph{Proceedings of the 2021 Conference on Empirical Methods in
  Natural Language Processing: System Demonstrations}, pages 175--184, Online
  and Punta Cana, Dominican Republic. Association for Computational
  Linguistics.

\bibitem[{Lin(2004)}]{lin-2004-rouge}
Chin-Yew Lin. 2004.
\newblock \href {https://aclanthology.org/W04-1013} {{ROUGE}: A package for
  automatic evaluation of summaries}.
\newblock In \emph{Text Summarization Branches Out}, pages 74--81, Barcelona,
  Spain. Association for Computational Linguistics.

\bibitem[{Liu et~al.(2019{\natexlab{a}})Liu, Chung, and Ren}]{liu2019summae}
Peter~J Liu, Yu-An Chung, and Jie Ren. 2019{\natexlab{a}}.
\newblock Summae: Zero-shot abstractive text summarization using
  length-agnostic auto-encoders.
\newblock \emph{arXiv preprint arXiv:1910.00998}.

\bibitem[{Liu et~al.(2019{\natexlab{b}})Liu, Ott, Goyal, Du, Joshi, Chen, Levy,
  Lewis, Zettlemoyer, and Stoyanov}]{liu2019roberta}
Yinhan Liu, Myle Ott, Naman Goyal, Jingfei Du, Mandar Joshi, Danqi Chen, Omer
  Levy, Mike Lewis, Luke Zettlemoyer, and Veselin Stoyanov. 2019{\natexlab{b}}.
\newblock Roberta: A robustly optimized bert pretraining approach.
\newblock \emph{arXiv preprint arXiv:1907.11692}.

\bibitem[{Liu et~al.(2021)Liu, Dou, and Liu}]{liu-etal-2021-refsum}
Yixin Liu, Zi-Yi Dou, and Pengfei Liu. 2021.
\newblock \href {https://doi.org/10.18653/v1/2021.naacl-main.113} {{R}ef{S}um:
  Refactoring neural summarization}.
\newblock In \emph{Proceedings of the 2021 Conference of the North American
  Chapter of the Association for Computational Linguistics: Human Language
  Technologies}, pages 1437--1448, Online. Association for Computational
  Linguistics.

\bibitem[{Liu and Liu(2021)}]{liu-liu-2021-simcls}
Yixin Liu and Pengfei Liu. 2021.
\newblock \href {https://doi.org/10.18653/v1/2021.acl-short.135} {{S}im{CLS}: A
  simple framework for contrastive learning of abstractive summarization}.
\newblock In \emph{Proceedings of the 59th Annual Meeting of the Association
  for Computational Linguistics and the 11th International Joint Conference on
  Natural Language Processing (Volume 2: Short Papers)}, pages 1065--1072,
  Online. Association for Computational Linguistics.

\bibitem[{Liu et~al.(2022{\natexlab{a}})Liu, Liu, Radev, and
  Neubig}]{liu-etal-2022-brio}
Yixin Liu, Pengfei Liu, Dragomir Radev, and Graham Neubig. 2022{\natexlab{a}}.
\newblock \href {https://doi.org/10.18653/v1/2022.acl-long.207} {{BRIO}:
  Bringing order to abstractive summarization}.
\newblock In \emph{Proceedings of the 60th Annual Meeting of the Association
  for Computational Linguistics (Volume 1: Long Papers)}, pages 2890--2903,
  Dublin, Ireland. Association for Computational Linguistics.

\bibitem[{Liu et~al.(2022{\natexlab{b}})Liu, Liu, Radev, and
  Neubig}]{liu2022brio}
Yixin Liu, Pengfei Liu, Dragomir Radev, and Graham Neubig. 2022{\natexlab{b}}.
\newblock Brio: Bringing order to abstractive summarization.
\newblock \emph{arXiv preprint arXiv:2203.16804}.

\bibitem[{Narayan et~al.(2018)Narayan, Cohen, and
  Lapata}]{narayan-etal-2018-dont}
Shashi Narayan, Shay~B. Cohen, and Mirella Lapata. 2018.
\newblock \href {https://doi.org/10.18653/v1/D18-1206} {Don{'}t give me the
  details, just the summary! topic-aware convolutional neural networks for
  extreme summarization}.
\newblock In \emph{Proceedings of the 2018 Conference on Empirical Methods in
  Natural Language Processing}, pages 1797--1807, Brussels, Belgium.
  Association for Computational Linguistics.

\bibitem[{Papineni et~al.(2002)Papineni, Roukos, Ward, and
  Zhu}]{papineni-etal-2002-bleu}
Kishore Papineni, Salim Roukos, Todd Ward, and Wei-Jing Zhu. 2002.
\newblock \href {https://doi.org/10.3115/1073083.1073135} {{B}leu: a method for
  automatic evaluation of machine translation}.
\newblock In \emph{Proceedings of the 40th Annual Meeting of the Association
  for Computational Linguistics}, pages 311--318, Philadelphia, Pennsylvania,
  USA. Association for Computational Linguistics.

\bibitem[{Raffel et~al.(2019)Raffel, Shazeer, Roberts, Lee, Narang, Matena,
  Zhou, Li, and Liu}]{raffel2019exploring}
Colin Raffel, Noam Shazeer, Adam Roberts, Katherine Lee, Sharan Narang, Michael
  Matena, Yanqi Zhou, Wei Li, and Peter~J Liu. 2019.
\newblock Exploring the limits of transfer learning with a unified text-to-text
  transformer.
\newblock \emph{arXiv preprint arXiv:1910.10683}.

\bibitem[{Ravaut et~al.(2022{\natexlab{a}})Ravaut, Joty, and
  Chen}]{ravaut-etal-2022-summareranker}
Mathieu Ravaut, Shafiq Joty, and Nancy Chen. 2022{\natexlab{a}}.
\newblock \href {https://doi.org/10.18653/v1/2022.acl-long.309}
  {{S}umma{R}eranker: A multi-task mixture-of-experts re-ranking framework for
  abstractive summarization}.
\newblock In \emph{Proceedings of the 60th Annual Meeting of the Association
  for Computational Linguistics (Volume 1: Long Papers)}, pages 4504--4524,
  Dublin, Ireland. Association for Computational Linguistics.

\bibitem[{Ravaut et~al.(2022{\natexlab{b}})Ravaut, Joty, and
  Chen}]{ravaut-etal-2022-towards}
Mathieu Ravaut, Shafiq Joty, and Nancy Chen. 2022{\natexlab{b}}.
\newblock \href {https://aclanthology.org/2022.emnlp-main.581} {Towards summary
  candidates fusion}.
\newblock In \emph{Proceedings of the 2022 Conference on Empirical Methods in
  Natural Language Processing}, pages 8488--8504, Abu Dhabi, United Arab
  Emirates. Association for Computational Linguistics.

\bibitem[{Reddy(1977)}]{reddy1977speech}
Raj Reddy. 1977.
\newblock Speech understanding systems: A summary of results of the five-year
  research effort at carnegie mellon university.
\newblock \emph{Pittsburgh, Pa}.

\bibitem[{See et~al.(2017)See, Liu, and Manning}]{see-etal-2017-get}
Abigail See, Peter~J. Liu, and Christopher~D. Manning. 2017.
\newblock \href {https://doi.org/10.18653/v1/P17-1099} {Get to the point:
  Summarization with pointer-generator networks}.
\newblock In \emph{Proceedings of the 55th Annual Meeting of the Association
  for Computational Linguistics (Volume 1: Long Papers)}, pages 1073--1083,
  Vancouver, Canada. Association for Computational Linguistics.

\bibitem[{Sellam et~al.(2020)Sellam, Das, and Parikh}]{sellam-etal-2020-bleurt}
Thibault Sellam, Dipanjan Das, and Ankur Parikh. 2020.
\newblock \href {https://doi.org/10.18653/v1/2020.acl-main.704} {{BLEURT}:
  Learning robust metrics for text generation}.
\newblock In \emph{Proceedings of the 58th Annual Meeting of the Association
  for Computational Linguistics}, pages 7881--7892, Online. Association for
  Computational Linguistics.

\bibitem[{Shazeer and Stern(2018)}]{shazeer2018adafactor}
Noam Shazeer and Mitchell Stern. 2018.
\newblock Adafactor: Adaptive learning rates with sublinear memory cost.
\newblock In \emph{International Conference on Machine Learning}, pages
  4596--4604. PMLR.

\bibitem[{Suzgun et~al.(2022)Suzgun, Melas-Kyriazi, and
  Jurafsky}]{suzgun2022follow}
Mirac Suzgun, Luke Melas-Kyriazi, and Dan Jurafsky. 2022.
\newblock Follow the wisdom of the crowd: Effective text generation via minimum
  bayes risk decoding.
\newblock \emph{arXiv preprint arXiv:2211.07634}.

\bibitem[{Tarvainen and Valpola(2017)}]{10.5555/3294771.3294885}
Antti Tarvainen and Harri Valpola. 2017.
\newblock Mean teachers are better role models: Weight-averaged consistency
  targets improve semi-supervised deep learning results.
\newblock In \emph{Proceedings of the 31st International Conference on Neural
  Information Processing Systems}, NIPS'17, page 1195–1204, Red Hook, NY,
  USA. Curran Associates Inc.

\bibitem[{Vijayakumar et~al.(2016)Vijayakumar, Cogswell, Selvaraju, Sun, Lee,
  Crandall, and Batra}]{vijayakumar2016diverse}
Ashwin~K Vijayakumar, Michael Cogswell, Ramprasath~R Selvaraju, Qing Sun,
  Stefan Lee, David Crandall, and Dhruv Batra. 2016.
\newblock Diverse beam search: Decoding diverse solutions from neural sequence
  models.
\newblock \emph{arXiv preprint arXiv:1610.02424}.

\bibitem[{Wolf et~al.(2020)Wolf, Debut, Sanh, Chaumond, Delangue, Moi, Cistac,
  Rault, Louf, Funtowicz, Davison, Shleifer, von Platen, Ma, Jernite, Plu, Xu,
  Le~Scao, Gugger, Drame, Lhoest, and Rush}]{wolf-etal-2020-transformers}
Thomas Wolf, Lysandre Debut, Victor Sanh, Julien Chaumond, Clement Delangue,
  Anthony Moi, Pierric Cistac, Tim Rault, Remi Louf, Morgan Funtowicz, Joe
  Davison, Sam Shleifer, Patrick von Platen, Clara Ma, Yacine Jernite, Julien
  Plu, Canwen Xu, Teven Le~Scao, Sylvain Gugger, Mariama Drame, Quentin Lhoest,
  and Alexander Rush. 2020.
\newblock \href {https://doi.org/10.18653/v1/2020.emnlp-demos.6} {Transformers:
  State-of-the-art natural language processing}.
\newblock In \emph{Proceedings of the 2020 Conference on Empirical Methods in
  Natural Language Processing: System Demonstrations}, pages 38--45, Online.
  Association for Computational Linguistics.

\bibitem[{Yang et~al.(2020)Yang, Zhu, Gmyr, Zeng, Huang, and
  Darve}]{yang-etal-2020-ted}
Ziyi Yang, Chenguang Zhu, Robert Gmyr, Michael Zeng, Xuedong Huang, and Eric
  Darve. 2020.
\newblock \href {https://doi.org/10.18653/v1/2020.findings-emnlp.168} {{TED}: A
  pretrained unsupervised summarization model with theme modeling and
  denoising}.
\newblock In \emph{Findings of the Association for Computational Linguistics:
  EMNLP 2020}, pages 1865--1874, Online. Association for Computational
  Linguistics.

\bibitem[{Yuan et~al.(2021)Yuan, Neubig, and Liu}]{yuan2021bartscore}
Weizhe Yuan, Graham Neubig, and Pengfei Liu. 2021.
\newblock Bartscore: Evaluating generated text as text generation.
\newblock \emph{arXiv preprint arXiv:2106.11520}.

\bibitem[{Zhang et~al.(2020)Zhang, Zhao, Saleh, and Liu}]{zhang2020pegasus}
Jingqing Zhang, Yao Zhao, Mohammad Saleh, and Peter Liu. 2020.
\newblock Pegasus: Pre-training with extracted gap-sentences for abstractive
  summarization.
\newblock In \emph{International Conference on Machine Learning}, pages
  11328--11339. PMLR.

\bibitem[{Zhang et~al.(2019)Zhang, Kishore, Wu, Weinberger, and
  Artzi}]{zhang2019bertscore}
Tianyi Zhang, Varsha Kishore, Felix Wu, Kilian~Q Weinberger, and Yoav Artzi.
  2019.
\newblock Bertscore: Evaluating text generation with bert.
\newblock \emph{arXiv preprint arXiv:1904.09675}.

\bibitem[{Zhang et~al.(2022)Zhang, Liang, Zuo, and Li}]{zhang2022unsupervised}
Zhihao Zhang, Xinnian Liang, Yuan Zuo, and Zhoujun Li. 2022.
\newblock Unsupervised abstractive summarization via sentence rewriting.
\newblock \emph{Computer Speech \& Language}, page 101467.

\end{thebibliography}
\bibliographystyle{acl_natbib}

\appendix

\section{Overlap with Simple Baselines}
\label{sec:appendix_a}

\begin{table}[h]
\resizebox{0.99\columnwidth}{!}{
\begin{tabular}{lcccc}

\toprule 

\textbf{\begin{tabular}[c]{@{}l@{}}Simple Candidate \\ Selection\end{tabular}} 
& \textbf{CNN/DM} & \textbf{XSum} & \textbf{WikiHow} & \textbf{SAMSum} \\

\midrule 

Max R-1 w. source         & 16.33 & 21.38 & 60.36 & 38.71 \\
Max R-2 w. source         & 21.45 & 24.32 & \textbf{66.94} & 44.93 \\
Max BLEU w. source        & 16.68 & 18.92 & 58.44 & 38.95 \\
Max BS w. source          & 43.46 & \textbf{69.98} & 35.50 & \textbf{58.61} \\
Max BaS w. source         & \textbf{47.15} & 46.06 & 13.85 & 52.50 \\
Max BRT w. source         & 14.74 & 13.43 & 15.75 & 23.32 \\
Max diversity feature     & 5.40 & 5.95 & 1.45 & 4.40 \\
Max length feature        & 11.80 & 7.46 & 14.19 & 13.68 \\

\cdashline{1-5}

Top beam                  & 15.05 & 12.85 & 9.18 & 30.28 \\
\emph{Oracle} candidate   & 15.24 & 12.27 & 10.73 & 18.44 \\  
\emph{Worst} candidate    & 5.33 & 7.48 & 7.65 & 7.20 \\   
\emph{Longest} candidate  & 20.58 & 22.74 & 64.43 & 51.28 \\

\bottomrule 

\end{tabular}
}
\caption{\small Overlap with simple re-reranking methods (\%) in unsupervised abstractive summarization with PEGASUS. We report the fraction (in percentage) of test set data points on the which SummScore falls back to a trivial summary candidate selection: maximizing one of the input features, picking the top beam, one oracle or worst candidate, or the longest one. All setups are with beam search with 20 candidates, thus a random baseline corresponds to 5\% overlap.}
\label{tab:a1}
\end{table}

\begin{table}[h]
\resizebox{0.99\columnwidth}{!}{
\begin{tabular}{lcccc}

\toprule 

\textbf{\begin{tabular}[c]{@{}l@{}}Simple Candidate \\ Selection\end{tabular}} 
& \textbf{CNN/DM} & \textbf{XSum} & \textbf{WikiHow} & \textbf{SAMSum} \\

\midrule 

Max R-1 w. source         & 16.00 & 32.10 & 58.70 & 14.53 \\
Max R-2 w. source         & 33.50 & 50.30 & \textbf{79.80} & 17.34 \\
Max BLEU w. source        & 17.80 & 31.20 & 57.10 & 12.58 \\
Max BS w. source          & \textbf{54.50} & \textbf{75.50} & 44.80 & 24.05 \\
Max BaS w. source         & 52.20 & 26.50 & 24.60 & \textbf{54.09} \\
Max BRT w. source         & 10.20 & 14.60 & 14.20 & 29.79 \\
Max diversity feature     & 9.60 & 1.90 & 1.00 & 3.30 \\
Max length feature        & 3.50 & 0.80 & 2.10 & 11.48 \\

\cdashline{1-5}

\emph{Oracle} candidate   & 9.00 & 1.80 & 9.00 & 12.21 \\
\emph{Worst} candidate    & 4.80 & 12.50 & 6.10 & 3.17 \\
\emph{Longest} candidate  & 10.90 & 22.70 & 39.60 & 6.47 \\

\bottomrule 

\end{tabular}
}
\caption{\small Overlap with simple re-reranking methods (\%) in unsupervised abstractive summarization with ChatGPT. We report the fraction (in percentage) of test set data points on the which SummScore falls back to a trivial summary candidate selection: maximizing one of the input features, picking one oracle or worst candidate, or the longest one. All setups are with beam search with 20 candidates, thus a random baseline corresponds to 5\% overlap.}
\label{tab:a2}
\end{table}

We perform a sanity check counting the percentage of time that SummScore falls back to a \emph{trivial} method of re-ranking summary candidates. For each feature described in \Cref{subsec:3_2}, we report the overlap between SummScore and a re-ranking approach consisting in picking the summary candidate maximing this feature. We also report baselines consisting in picking the top beam, an oracle or a \emph{worst} candidate, and the longest candidate. As seen in \Cref{tab:a1,tab:a2}, across both backbones PEGASUS and ChatGPT, SummScore never collapses to a trivial candidate selection, and we see similar patterns on the same dataset (e.g., highest overlap with a single feature selection is with BERTScore with source feature on CNN/DM).

\section{Generation \& Fine-Tuning Details}
\label{sec:appendix_b}

In \Cref{tab:b_1}, we show generation hyper-parameters used for each dataset to generate beam search summary candidates used in \Cref{tab:3b}. For the transfer setup shown in \Cref{tab:4}, we use as generation hyper-parameters on each target dataset the parameters used on that dataset for \Cref{tab:3b}. For instance, PEGASUS-XSum, PEGASUS-WikiHow and PEGASUS-SAMSum, when transferred to CNN/DM, are decoded with hyper-parameters of PEGASUS-CNN/DM shown in \Cref{tab:b_1}.

\begin{table}[h]
\resizebox{0.99\columnwidth}{!}{
\begin{tabular}{llcccc}

\toprule 

\textbf{Dataset}         
& \textbf{Model} 
& \textbf{\begin{tabular}[c]{@{}c@{}}Max source\\ length\end{tabular}} & \textbf{\begin{tabular}[c]{@{}c@{}}Max target\\ length\end{tabular}} & \textbf{\begin{tabular}[c]{@{}c@{}}Length\\ penalty\end{tabular}} & \textbf{\begin{tabular}[c]{@{}c@{}}Trigram\\ blocking\end{tabular}} \\

\midrule 

\multirow{3}{*}{CNN/DM}  
& PEGASUS & \multirow{3}{*}{1024} & \multirow{3}{*}{128} & 0.8  & Yes \\
& BART & & & 1.0 & Yes \\
& BRIO &  & & 1.0 & Yes \\

\midrule 

\multirow{3}{*}{XSum}    
& PEGASUS & \multirow{3}{*}{512} & \multirow{3}{*}{64} & 0.8 & Yes \\
& BART & & & 1.0 & Yes \\
& BRIO & & & 0.8 & Yes \\

\midrule 

\multirow{2}{*}{WikiHow} 
& PEGASUS & \multirow{2}{*}{512} & \multirow{2}{*}{128} & 0.6 & No \\
& BART & & & 1.0 & Yes  \\

\midrule 

\multirow{2}{*}{SAMSum}  
& PEGASUS & \multirow{2}{*}{512} & \multirow{2}{*}{64} & 0.8 & No \\
& BART & & & 1.0 & Yes        \\

\toprule 

\end{tabular}
}
\caption{\small Generation hyper-parameters for each dataset and model used to produce summary candidates.}
\label{tab:b_1}
\end{table}

For experiments shown in \Cref{tab:4}, we fine-tune ourselves BART on WikiHow dataset, and PEGASUS and BART on SAMSum dataset. Fine-tuning hyper-parameters are shown in \Cref{tab:b_2}. We perform early stopping with regards to the mean ROUGE on the validation set. Our BART reaches 44.21/19.31/34.67 {\sc{Rouge-1/2/L}} on WikiHow test set, our PEGASUS 52.33/27.97/44.02 {\sc{Rouge-1/2/L}} on SAMSum test set, and our BART 52.78/28.28/44.08 {\sc{Rouge-1/2/L}}.

\begin{table}[h]
\resizebox{0.99\columnwidth}{!}{
\begin{tabular}{llccccccc}

\toprule 

\textbf{Dataset}        
& \textbf{Model} 
& \textbf{Epochs} 
& \textbf{Optimizer} 
& \textbf{Scheduler} 
& \textbf{LR} 
& \textbf{BS} 
& \textbf{LS} 
& \textbf{\begin{tabular}[c]{@{}c@{}}Eval\\ steps\end{tabular}} \\

\midrule 

WikiHow 
& BART & 15 & Adam & none & 1e-5 & 80 & 0.1 & 250 \\
\multirow{2}{*}{SAMSum}
& PEGASUS & 30 & Adam & none & 1e-4 & 256 & 0.1 & 50 \\
& BART & 30 & Adam & linear & 1e-5 & 80 & 0.1 & 50 \\

\bottomrule

\end{tabular}
}
\caption{\small Fine-tuning hyper-parameters used to fine-tune BART on WikiHow and PEGASUS and BART on SAMSum.}
\label{tab:b_2}
\end{table}

\section{Recall Analysis}
\label{sec:appendix_c}

\begin{figure*}[]
    \centering
    \includegraphics[width=\textwidth]{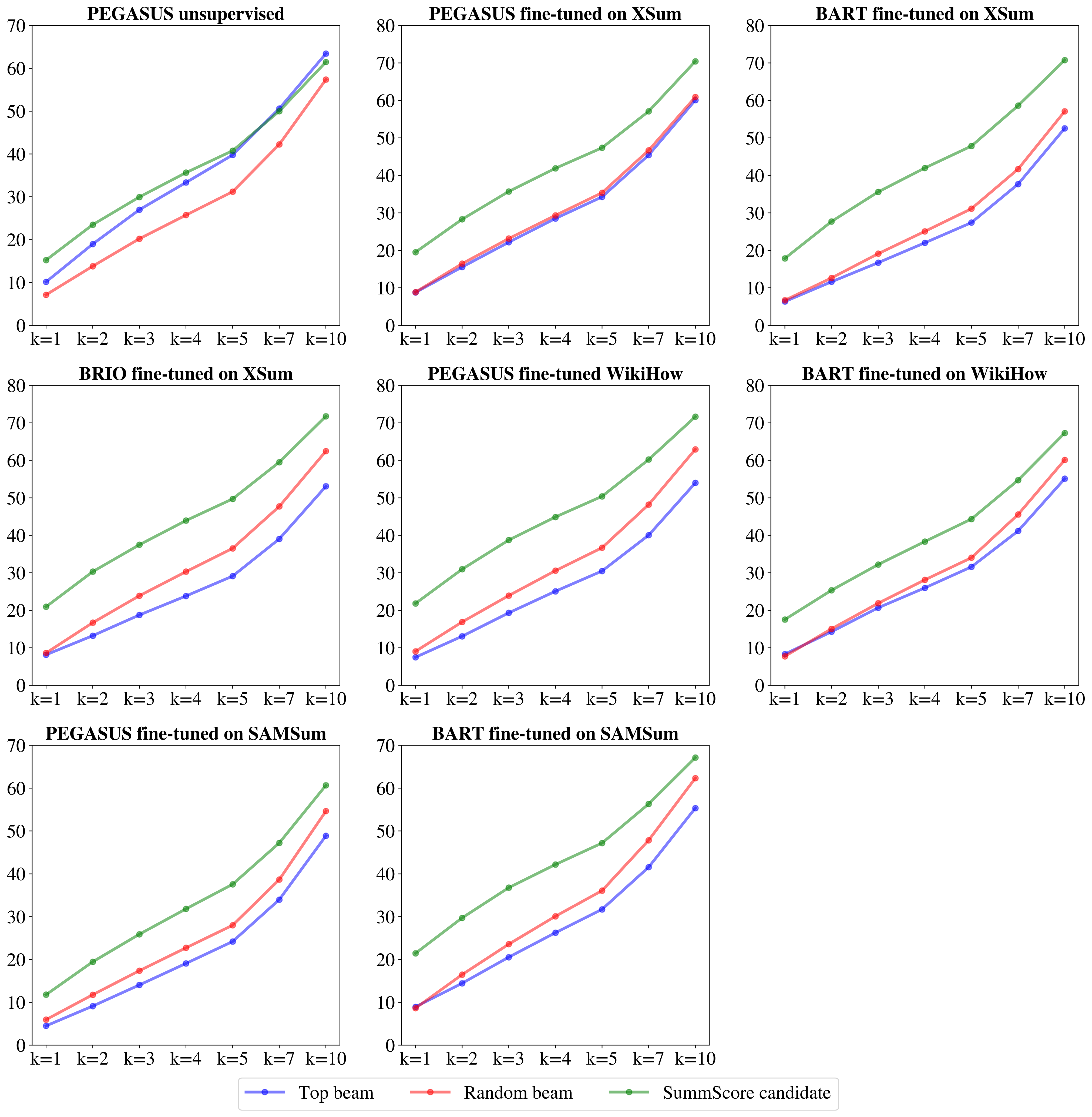}
    \caption{\small Recall curves on CNN/DM with PEGASUS backbone. The top left plot corresponds to unsupervised summarization re-ranking from \Cref{tab:3b}, and the next seven plots to all zero-shot transfer summarization setups from \Cref{tab:4}. Each re-ranking setup has 20 summary candidates, and we show recall over \emph{any oracle candidate} for several thresholds $k \in \{1,2,3,4,5,7,10\}$.}
    \label{fig:c_1}
\end{figure*}

\begin{figure*}[]
    \centering
    \includegraphics[width=\textwidth]{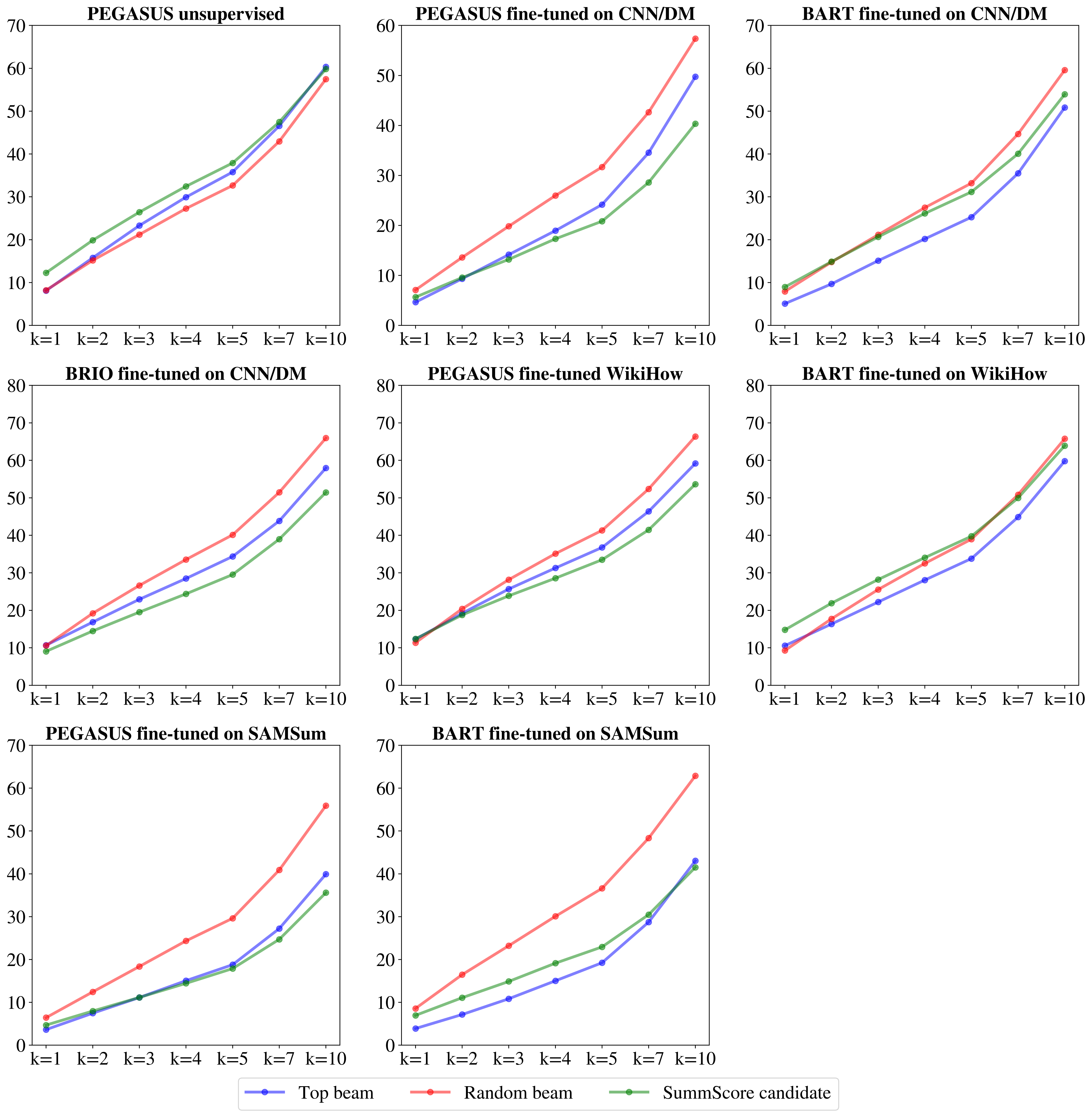}
    \caption{\small Recall curves on XSum with PEGASUS backbone. The top left plot corresponds to unsupervised summarization re-ranking from \Cref{tab:3b}, and the next seven plots to all zero-shot transfer summarization setups from \Cref{tab:4}. Each re-ranking setup has 20 summary candidates, and we show recall over \emph{any oracle candidate} for several thresholds $k \in \{1,2,3,4,5,7,10\}$.}
    \label{fig:c_2}
\end{figure*}

\begin{figure*}[]
    \centering
    \includegraphics[width=\textwidth]{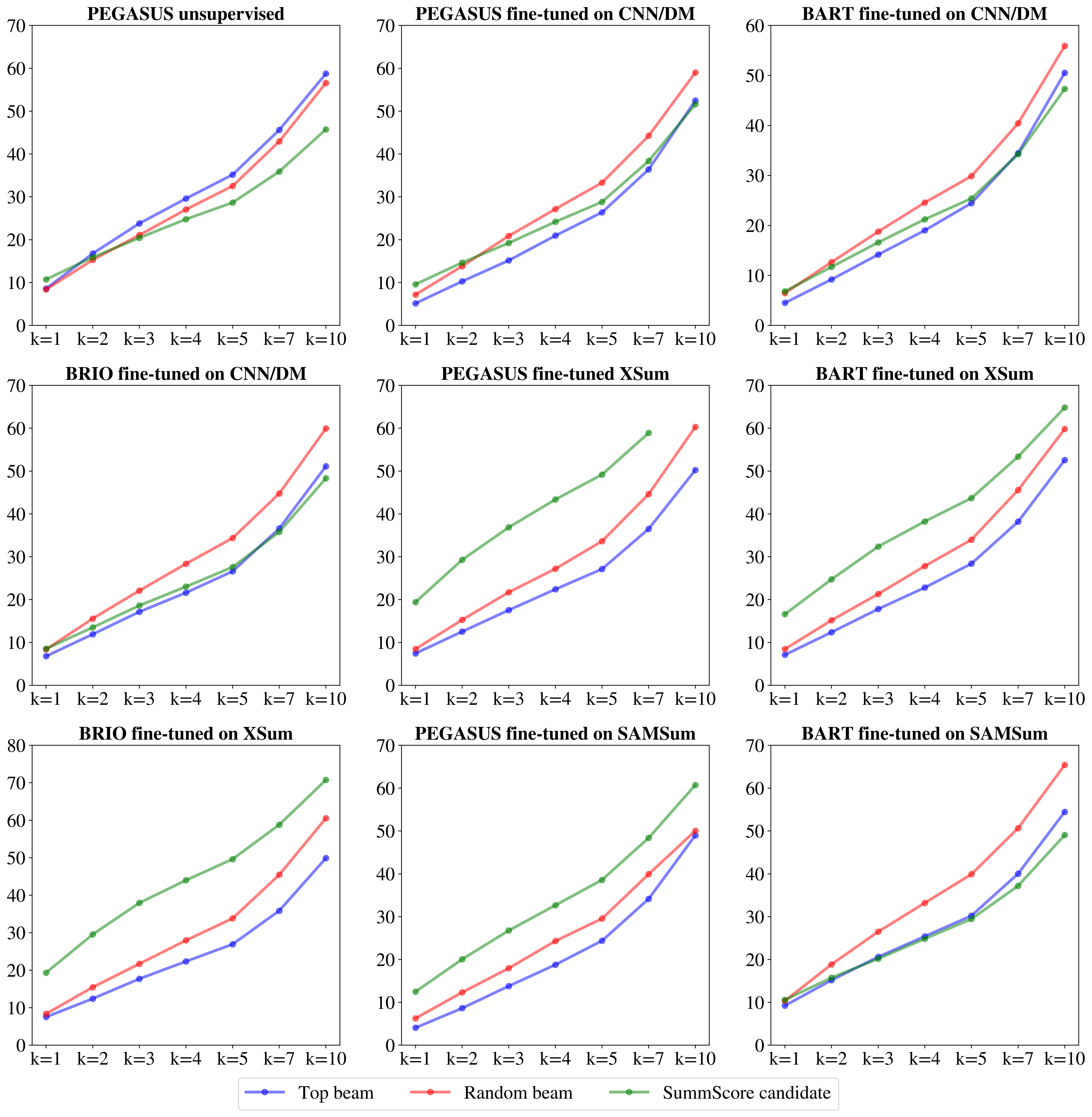}
    \caption{\small Recall curves on WikiHow with PEGASUS backbone. The top left plot corresponds to unsupervised summarization re-ranking from \Cref{tab:3b}, and the next eight plots to all zero-shot transfer summarization setups from \Cref{tab:4}. Each re-ranking setup has 20 summary candidates, and we show recall over \emph{any oracle candidate} for several thresholds $k \in \{1,2,3,4,5,7,10\}$.}
    \label{fig:c_3}
\end{figure*}

\begin{figure*}[]
    \centering
    \includegraphics[width=\textwidth]{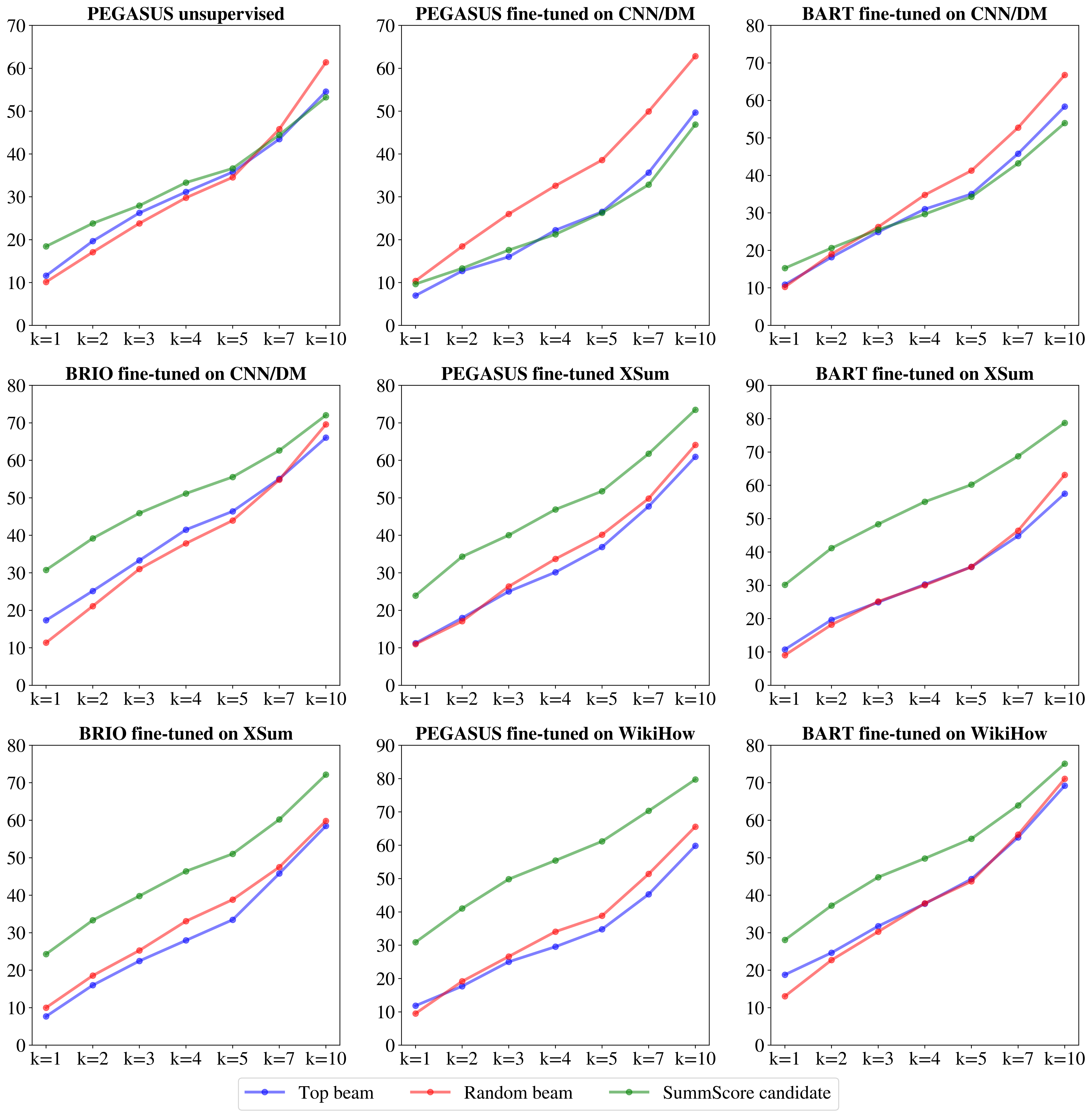}
    \caption{\small Recall curves on SAMSum with PEGASUS backbone. The top left plot corresponds to unsupervised summarization re-ranking from \Cref{tab:3b}, and the next eight plots to all zero-shot transfer summarization setups from \Cref{tab:4}. Each re-ranking setup has 20 summary candidates, and we show recall over \emph{any oracle candidate} for several thresholds $k \in \{1,2,3,4,5,7,10\}$.}
    \label{fig:c_4}
\end{figure*}

Besides the quality of the selected summary, we also analyze re-ranking performance itself. In \Cref{fig:c_1}, \Cref{fig:c_2}, \Cref{fig:c_3} and \Cref{fig:c_4}, we show recall curves on each dataset and for all unsupervised and zero-shot summarization setups. Recall@k is defined as the probability of outputting \emph{one} of the oracle summary candidates (candidates maximizing the mean {\sc{Rouge}} with the target) among the first k candidates. We compare SummScore with the baseline beam search output, and a random candidate selection baseline. 

In most cases, SummScore (green curves) provides higher recall, with the notable exception of XSum, where both beam search and SummScore and XSum can fail to improve the random baseline.


\section{Abstractiveness Analysis}
\label{sec:appendix_d}

In \Cref{tab:d}, we show {\sc{Rouge}} results from \Cref{tab:5} alongside abstractiveness results, as measured per the fraction of novel n-grams in output summaries, for re-ranking and self-training experiments. Maximizing both {\sc{Rouge}} and abstractiveness is notoriously difficult, as easy solutions for abstractiveness optimization can deviate a lot from the source, resulting in a harmed {\sc{Rouge}} score. 

\begin{table*}[h]
\resizebox{0.99\textwidth}{!}{
\begin{tabular}{llccccccc}

\toprule

\multirow{3}{*}{\textbf{Dataset}} 
& \multirow{3}{*}{\textbf{Model}}                  
& \multicolumn{4}{c}{\textbf{ROUGE}}                                                        
& \multicolumn{3}{c}{\textbf{Abstractiveness (new n-grams)}}                           \\
& & \textbf{Mean R} & \textbf{R-1} & \textbf{R-2} & \textbf{R-L} 
& {\textbf{\begin{tabular}[c]{@{}c@{}}New\\ 1-grams\end{tabular}}}
& {\textbf{\begin{tabular}[c]{@{}c@{}}New\\ 2-grams\end{tabular}}}
& {\textbf{\begin{tabular}[c]{@{}c@{}}New\\ 3-grams\end{tabular}}} \\

\midrule 

\multirow{12}{*}{\textbf{CNN/DM}}  
& PEGASUS                                                                    & 26.99 & 35.47 & 13.89 & 31.61 & 0.19 & 0.89 & 2.44 \\
& PEGASUS + SummScore LEAD-3                                                 & 28.38 & 36.92 & 15.03 & 33.19 & 0.19 & 0.94 & 2.73 \\

\cdashline{2-9}

& PEGASUS + SummScore LEAD-3 - paraphrasing 100\%                            & 22.46 & 29.72 & 11.07 & 26.58 & \textbf{14.01} & \textbf{35.18} & \textbf{44.23} \\
& PEGASUS + SummScore LEAD-3 - paraphrasing 50\%                             & 25.37 & 33.24 & 13.02 & 29.83 & 7.29 & 18.34 & 23.77 \\
& PEGASUS + SummScore LEAD-3 - paraphrasing 25\% (\textbf{pseudo-labels})    & \underline{26.85} & \underline{35.06} & \underline{13.99} & \underline{31.49} & \underline{3.73} & \underline{9.71} & \underline{13.36} \\
& PEGASUS + SummScore LEAD-3 - paraphrasing 12.5\%                           & 27.61 & 35.99 & 14.50 & 32.35 & 1.95 & 5.29 & 7.98 \\

\cdashline{2-9}

& PEGASUS self-trained (1st round)                                           & 27.98 & 36.68 & 14.52 & 32.72 & 0.25 & 0.66 & 1.84 \\
& PEGASUS self-trained (1st round) + SummScore LEAD-3                        & 29.88 & 38.75 & 16.11 & 34.78 & 0.10 & 0.43 & 1.60 \\
& PEGASUS self-trained (2nd round)                                           & 29.40 & 38.17 & 15.77 & 34.25 & 0.66 & 1.49 & 2.61 \\
& PEGASUS self-trained (2nd round) + SummScore LEAD-3                        & 30.59 & 39.49 & \textbf{16.69} & 35.61 & 0.21 & 0.93 & 2.15 \\
& PEGASUS self-trained (3rd round)                                           & 29.63 & 38.47 & 15.95 & 34.48 & 0.68 & 1.72 & 2.74 \\
& PEGASUS self-trained (3rd round) + SummScore LEAD-3                        & \textbf{30.80} & \textbf{39.76} & \textbf{16.79} & \textbf{35.85} & 0.11 & 0.99 & 2.25 \\

\midrule

\multirow{8}{*}{\textbf{XSum}}    
& PEGASUS                                                                    & 11.83 & 18.77 & 2.86 & 13.85 & 0.20 & 0.44 & 1.16 \\
& PEGASUS + SummScore LEAD-3                                                 & 12.45 & 19.62 & 3.02 & 14.71 & 0.19 & 0.60 & 2.04 \\

\cdashline{2-9}

& PEGASUS + SummScore LEAD-3 - paraphrasing 100\% (\textbf{pseudo-labels})   & \underline{\textbf{12.98}} & \underline{\textbf{20.19}} & \underline{\textbf{3.60}} & \underline{\textbf{15.16}} & \underline{\textbf{12.94}} & \underline{\textbf{30.30}} & \underline{\textbf{37.63}} \\
& PEGASUS + SummScore LEAD-3 - paraphrasing 50\%                             & 12.75 & 19.94 & 3.32 & 14.97 & 6.55 & 15.46 & 19.87 \\
& PEGASUS + SummScore LEAD-3 - paraphrasing 25\%                             & 12.61 & 19.79 & 3.18 & 14.86 & 3.41 & 8.06 & 10.96 \\
& PEGASUS + SummScore LEAD-3 - paraphrasing 12.5\%                           & 12.52 & 19.71 & 3.10 & 14.77 & 1.83 & 4.36 & 6.53 \\

\cdashline{2-9}

& PEGASUS self-trained                                                       & 12.09 & 19.33 & 2.76 & 14.18 & 1.49 & 3.20 & 4.43 \\
& PEGASUS self-trained + SummScore LEAD-3                                    & 12.60 & 20.02 & 2.84 & 14.93 & 0.66 & 1.99 & 3.55 \\

\midrule

\multirow{8}{*}{\textbf{WikiHow}} 
& PEGASUS                                                                    & 16.46 & 25.49 & 5.91 & 17.99 & 0.48 & 1.12 & 2.36 \\
& PEGASUS + SummScore R-2                                                    & \textbf{17.17} & 26.40 & \textbf{6.30} & 18.83 & 0.80 & 2.47 & 5.05 \\

\cdashline{2-9}

& PEGASUS + SummScore R-2 - paraphrasing 100\%                               & 16.75 & 25.59 & 6.19 & 18.47 & \textbf{4.65} & \textbf{17.13} & \textbf{26.14} \\
& PEGASUS + SummScore R-2 - paraphrasing 50\% (\textbf{pseudo-labels})       & \underline{16.97} & \underline{26.01} & \underline{\textbf{6.26}} & \underline{18.62} & \underline{2.79} & \underline{9.82} & \underline{15.55} \\
& PEGASUS + SummScore R-2 - paraphrasing 25\%                                & 17.08 & 26.24 & \textbf{6.27} & 18.73 & 1.81 & 6.14 & 10.28 \\
& PEGASUS + SummScore R-2 - paraphrasing 12.5\%                              & 17.13 & 26.32 & \textbf{6.28} & 18.79 & 1.31 & 4.34 & 7.71 \\

\cdashline{2-9}

& PEGASUS self-trained                                                       & 16.92 & 26.08 & 6.08 & 18.59 & 0.84 & 1.80 & 3.56 \\
& PEGASUS self-trained + SummScore R-2                                       & \textbf{17.27} & \textbf{26.50} & \textbf{6.28} & \textbf{19.03} & 0.61 & 1.71 & 4.02 \\

\midrule

\multirow{8}{*}{\textbf{SAMSum}}  
& PEGASUS                                                                    & 18.57 & 26.64 & 6.32 & 22.75 & 0.30 & 1.35 & 2.81 \\
& PEGASUS + SummScore LEAD-3                                                 & 19.92 & 28.22 & 7.16 & 24.39 & 0.54 & 1.73 & 3.85 \\

\cdashline{2-9}

& PEGASUS + SummScore LEAD-3 - paraphrasing 100\%                            & 15.95 & 22.84 & 4.14 & 20.88 & \textbf{15.08} & \textbf{37.45} & \textbf{50.66} \\
& PEGASUS + SummScore LEAD-3 - paraphrasing 50\%                             & 17.77 & 25.34 & 5.55 & 22.43 & 7.45 & 18.83 & 26.22 \\
& PEGASUS + SummScore LEAD-3 - paraphrasing 25\%                             & 18.88 & 26.83 & 6.40 & 23.41 & 3.93 & 9.75 & 14.23 \\
& PEGASUS + SummScore LEAD-3 - paraphrasing 12.5\% (\textbf{pseudo-labels})  & \underline{19.33} & \underline{27.41} & \underline{6.73} & \underline{23.84} & \underline{2.28} & \underline{5.85} & \underline{9.29} \\

\cdashline{2-9}

& PEGASUS self-trained                                                       & 18.92 & 26.96 & 6.41 & 23.40 & 0.36 & 1.51 & 3.35 \\
& PEGASUS self-trained + SummScore LEAD-3                                    & \textbf{20.67} & \textbf{28.91} & \textbf{7.55} & \textbf{25.54} & 0.60 & 2.18 & 4.93 \\

\toprule

\end{tabular}
}
\caption{\small {\sc{Rouge}} and abstractiveness for several models: the unsupervised PEGASUS (first sub-block), re-ranking with SummScore (first sub-block), paraphrasing the resulting pseudo-labels (second sub-block), self-training with the pseudo-labels (third sub-block), then re-ranking self-training outputs with SummScore again (third sub-block). All results are on the test set, results of self-training pseudo-labels are underlined, and highest numbers within 0.1 are in bold.}
\label{tab:d}
\end{table*}

The unsupervised PEGASUS (first row of each block) is very extractive and only produces a small fraction of novel n-grams. SummScore selected summaries, despite maximizing a score which maximizes the mean {\sc{Rouge}} with pseudo-labels extracted from the source document, both improve the {\sc{Rouge}} and the abstractiveness level. However, SummScore re-ranking applied to self-trained models tends to reduce their abstractiveness level, although it stays above the level of the baseline PEGASUS. Paraphrased summaries drastically increase abstractiveness, at the expense of {\sc{Rouge}} - except on XSum where paraphrasing also improves {\sc{Rouge}}, motivating our choice to use 100\% paraphrased summaries as pseudo-labels. We confirm that our pseudo-labels for self-training, made of a blend of SummScore selected summaries and selected summaries being paraphrased, maintains high {\sc{Rouge}} while being much more abstractive than the baseline PEGASUS.

\section{Paraphrasing Model}
\label{sec:appendix_e}

For each dataset, we fine-tune BART-large \cite{lewis-etal-2020-bart} (from the pre-training checkpoint \emph{facebook/bart-large} in HuggingFace transformers \cite{wolf-etal-2020-transformers}) for paraphrasing. The model is trained to paraphrase blocks of $n=3$ sentences on CNN/DM, $n=1$ sentence on XSum, and $n=2$ sentences on WikiHow and SAMSum, in line with average summary lengths on these datasets. We train the model with Adafactor \cite{shazeer2018adafactor} for 5 epochs, with effective batch size 32, learning rate 2e-5, and no weight decay nor label smoothing. We evaluate every 500 optimization steps on CNN/DM, XSum, and WikiHow, and every 100 steps on SAMSum. At inference, we use beam search with beam width 5 and length penalty of 1.0, and block repeated trigrams like in \cite{kryscinski-etal-2018-improving}.

\begin{table}[h!]
\resizebox{1.00\columnwidth}{!}{
\begin{tabular}{lcccc}

\toprule

\textbf{Dataset}         & CNN/DM & XSum  & WikiHow & SAMSum \\

\midrule

\textbf{Paraphrasing model}    & 32.88 & 15.58 & 20.34 & 17.44 \\

\bottomrule
\end{tabular}
}
\caption{\small {\sc{Rouge}} results of the paraphrasing model, on the validation set of each dataset. We report the mean of {\sc{Rouge-1/2/L}}.}
\label{tab:e_1}
\end{table}

We track the mean of {\sc{Rouge-1}}, {\sc{Rouge-2}} and {\sc{Rouge-L}} between the generated paraphrase and target paraphrase on the validation set during training, and perform early stopping. Best mean {\sc{Rouge}} results are shown in \Cref{tab:e_1}.

\begin{table}[h]
\resizebox{1.00\columnwidth}{!}{
\begin{tabular}{lcccc}

\toprule 

\textbf{Dataset} 
& \textbf{Mean R} 
& \textbf{\begin{tabular}[c]{@{}c@{}}New \\ 1-grams\end{tabular}}
& \textbf{\begin{tabular}[c]{@{}c@{}}New\\ 2-grams\end{tabular}} 
& \textbf{\begin{tabular}[c]{@{}c@{}}New \\ 3-grams\end{tabular}} \\

\midrule 

CNN/DM           & 55.80 & 17.28 & 34.58 & 39.61 \\
XSum             & 62.13 & 20.93 & 34.60 & 38.59 \\
WikiHow          & 81.26 & 7.96 & 20.14 & 25.60 \\
SAMSum           & 50.64 & 22.52 & 41.29 & 52.02 \\

\toprule 

\end{tabular}
}
\caption{\small Impact of paraphrasing on the pseudo-targets. We report mean {\sc{Rouge}} and percentage of novel n-grams between the paraphrased pseudo-targets and the original pseudo-targets, on the \emph{training} set of each dataset since this is the subset that paraphrasing is applied to.}
\label{tab:e_2}
\end{table}

Next, we study the impact of the paraphrasing model on the SummScore pseudo-targets. In \Cref{tab:e_2}, we compute the mean {\sc{Rouge}} between pseudo-targets and their paraphrase, and analyze the novel n-grams. We point out that the paraphrasing is only applied to the \emph{training} pseudo-labels as the goal of paraphrasing is to encourage the model to learn diversity during self-training, hence \Cref{tab:e_2} reporting results on \emph{training sets}. On each dataset, the mean {\sc{Rouge}} is in the 50-80 range, indicating that the paraphrased pseudo-labels do not deviate too much from the original pseudo-labels and yet is able to re-write some content. Besides, there is a high proportion of new n-grams: more than 10\% new 1-grams (with the exception of WikiHow on the which the paraphrasing model seems to struggle more to rephrase the input), and more than 20\% 2-grams.

\section{Other Summary Candidates Setups}
\label{sec:appendix_f}

\begin{table}[h!]
\resizebox{\columnwidth}{!}{
\begin{tabular}{llcccc}
\toprule
\multirow{2}{*}{\textbf{\begin{tabular}[c]{@{}l@{}}Decoding\\ method\end{tabular}}}
& \multirow{2}{*}{\textbf{\begin{tabular}[c]{@{}l@{}}Candidate\\ Selection\end{tabular}}}
& \multicolumn{4}{c}{\textbf{\# Candidates}} \\
& & \textbf{5} 
& \textbf{10} 
& \textbf{15} 
& \textbf{20} \\

\midrule

\multirow{2}{*}{Beam search}                                    
& PEGASUS                    & 26.74 & 27.00 & 27.00 & 26.99 \\
& SummScore                  & 27.46 & 28.01 & 28.33 & 28.38 \\

\cdashline{1-6}

\multirow{2}{*}{Diverse beam search}                            
& PEGASUS                    & 26.08 & 26.08 & 26.07 & 26.01 \\
& SummScore                  & 26.98 & 27.48 & 27.76 & 27.87 \\

\cdashline{1-6}

\multirow{2}{*}{Nucleus sampling}                                                                          
& PEGASUS                    & 23.92 & 23.95 & 24.04 & 24.03 \\
& SummScore                  & 26.13 & 26.57 & 26.85 & 27.11 \\ 

\midrule

\multirow{2}{*}{All three methods}                                                                          
& \multirow{2}{*}{SummScore} & \textbf{15} & \textbf{30} & \textbf{45} & \textbf{60} \\ 
&                            & 27.87 & 28.35 & 28.34 & \textbf{28.59} \\

\bottomrule

\end{tabular}
}
\caption{\small Candidate generation setups. We compare several summary candidates generation setups with PEGASUS on CNN/DM, varying both the decoding method and the number of candidates. We report the mean of {\sc{Rouge-1/2/L}}. Best results within 0.1 are in bold.}
\label{tab:f}
\end{table}

\begin{table*}[t]
\resizebox{\textwidth}{!}{
\begin{tabular}{llcccccccc}

\toprule 

\textbf{Dataset}        
& \textbf{Model}                                 
& \textbf{ROUGE-1} 
& \textbf{ROUGE-2} 
& \textbf{BLEU} 
& \textbf{BERTScore} 
& \textbf{BARTScore} 
& \textbf{BleuRT} 
& \textbf{Diversity} 
& \textbf{Length} \\

\midrule 

\multirow{8}{*}{CNN/DM} 
& SummScore - Random-3                                      & 0.0000 & \textbf{0.5700} & 0.0300 & 0.2681 & 0.0000 & 0.0069 & 0.1250 & 0.0000 \\
& SummScore - LEAD-3 (\textbf{selected SummScore version})  & 0.0000 & 0.0250 & 0.0000 & \textbf{0.4275} & 0.3375 & 0.1350 & 0.0500 & 0.0250 \\
& SummScore - Salient-R1                                    & 0.0850 & \textbf{0.7650} & 0.0000 & 0.1000 & 0.0031 & 0.0219 & 0.0000 & 0.0250 \\
& SummScore - Salient-R2                                    & 0.1444 & 0.1856 & \textbf{0.4950} & 0.1050 & 0.0000 & 0.0450 & 0.0000 & 0.0250 \\
& SummScore - Salient-RL                                    & 0.1062 & \textbf{0.7438} & 0.0000 & 0.1000 & 0.0031 & 0.0219 & 0.0000 & 0.0250 \\
& Self-training (1st round) + SummScore - LEAD-3            & 0.0000 & 0.0000 & 0.0000 & \textbf{0.4500} & 0.4275 & 0.0225 & 0.1000 & 0.0000 \\
& Self-training (2nd round) + SummScore - LEAD-3            & 0.0000 & 0.0000 & 0.0000 & \textbf{0.6338} & 0.2925 & 0.0488 & 0.0250 & 0.0000 \\
& Self-training (3rd round) + SummScore - LEAD-3            & 0.0000 & 0.0500 & 0.0000 & \textbf{0.8075} & 0.1425 & 0.0000 & 0.0000 & 0.0000 \\

\midrule 

\multirow{6}{*}{XSum} 
& SummScore - Random-3                                      & 0.0287 & \textbf{0.5462} & 0.0000 & 0.1200 & 0.0900 & 0.1900 & 0.0250 & 0.0000 \\
& SummScore - LEAD-3 (\textbf{selected SummScore version})  & 0.0500 & 0.0000 & 0.0000 & \textbf{0.7837} & 0.1425 & 0.0238 & 0.0000 & 0.0000 \\
& SummScore - Salient-R1                                    & 0.1275 & \textbf{0.7225} & 0.0000 & 0.0338 & 0.0000 & 0.0413 & 0.0000 & 0.0750 \\
& SummScore - Salient-R2                                    & \textbf{0.8000} & 0.0000 & 0.0000 & 0.0000 & 0.0000 & 0.2000 & 0.0000 & 0.0000 \\
& SummScore - Salient-RL                                    & 0.1200 & 0.1600 & \textbf{0.5200} & 0.1550 & 0.0000 & 0.0450 & 0.0000 & 0.0000 \\
& Self-training (1st round) + SummScore - LEAD-3            & 0.0000 & 0.0000 & 0.0000 & \textbf{0.5550} & 0.3700 & 0.0000 & 0.0250 & 0.0500 \\

\midrule 

\multirow{6}{*}{WikiHow} 
& SummScore - Random-3                                         & 0.0100 & 0.0400 & 0.0000 & \textbf{0.9025} & 0.0238 & 0.0238 & 0.0000 & 0.0000 \\
& SummScore - LEAD-3                                           & 0.0000 & 0.0000 & 0.0000 & \textbf{0.7312} & 0.2437 & 0.0000 & 0.0250 & 0.0000 \\
& SummScore - Salient-R1                                       & 0.1094 & \textbf{0.7656} & 0.0000 & 0.0825 & 0.0000 & 0.0175 & 0.0250 & 0.0000 \\
& SummScore - Salient-R2 (\textbf{selected SummScore version}) & \textbf{0.8750} & 0.0000 & 0.0000 & 0.0825 & 0.0000 & 0.0175 & 0.0250 & 0.0000 \\
& SummScore - Salient-RL                                       & 0.2625 & \textbf{0.6125} & 0.0000 & 0.0825 & 0.0000 & 0.0175 & 0.0250 & 0.0000 \\
& Self-training (1st round) + SummScore - Salient-R2           & \textbf{0.5031} & 0.1750 & 0.1969 & 0.0625 & 0.0050 & 0.0325 & 0.0250 & 0.0000 \\

\midrule 

\multirow{6}{*}{SAMSum} 
& SummScore - Random-3                                         & 0.0300 & 0.2625 & 0.0075 & \textbf{0.4900} & 0.2100 & 0.0000 & 0.0000 & 0.0000 \\
& SummScore - LEAD-3 (\textbf{selected SummScore version})     & 0.0000 & 0.0000 & 0.0000 & \textbf{0.7750} & 0.2250 & 0.0000 & 0.0000 & 0.0000 \\
& SummScore - Salient-R1                                       & 0.1650 & \textbf{0.6600} & 0.0000 & 0.0000 & 0.0000 & 0.0000 & 0.1250 & 0.0500\\
& SummScore - Salient-R2                                       & 0.0731 & \textbf{0.8044} & 0.0975 & 0.0000 & 0.0000 & 0.0000 & 0.0000 & 0.0250\\
& SummScore - Salient-RL                                       & 0.1950 & \textbf{0.7800} & 0.0000 & 0.0000 & 0.0000 & 0.0000 & 0.0000 & 0.0250 \\
& Self-training (1st round) + SummScore - LEAD-3               & 0.0000 & 0.0000 & 0.0000 & \textbf{0.8500} & 0.1500 & 0.0000 & 0.0000 & 0.0000\\

\bottomrule

\end{tabular}
}
\caption{\small Coefficients learned by SummScore with PEGASUS for each feature in each dataset and with each pseudo-labels construction technique. Highest feature values for each model are in bold.}
\label{tab:g1}
\end{table*}

\begin{table*}[t]
\resizebox{\textwidth}{!}{
\begin{tabular}{llcccccccc}

\toprule 

\textbf{Dataset}        
& \textbf{Model}                                 
& \textbf{ROUGE-1} 
& \textbf{ROUGE-2} 
& \textbf{BLEU} 
& \textbf{BERTScore} 
& \textbf{BARTScore} 
& \textbf{BleuRT} 
& \textbf{Diversity} 
& \textbf{Length} \\

\midrule 

\multirow{5}{*}{CNN/DM} 
& SummScore - Random-3                                      & 0.0600 & 0.2400 & 0.0000 & \textbf{0.3881} & 0.1437 & 0.0431 & 0.1250 & 0.0000 \\
& SummScore - LEAD-3 (\textbf{selected SummScore version})  & 0.0000 & 0.0975 & 0.0025 & \textbf{0.5038} & 0.2712 & 0.0000 & 0.1250 & 0.0000 \\
& SummScore - Salient-R1                                    & 0.2925 & \textbf{0.6075} & 0.0000 & 0.0925 & 0.0025 & 0.0050 & 0.0000 & 0.0000 \\
& SummScore - Salient-R2                                    & \textbf{0.3825} & 0.3375 & 0.1800 & 0.0850 & 0.0075 & 0.0075 & 0.0000 & 0.0000 \\
& SummScore - Salient-RL                                    & 0.2925 & \textbf{0.6075} & 0.0000 & 0.0825 & 0.0000 & 0.0175 & 0.0000 & 0.0000 \\
\midrule 

\multirow{5}{*}{XSum} 
& SummScore - Random-3                                      & 0.0581 & \textbf{0.4844} & 0.2325 & 0.1350 & 0.0150 & 0.0500 & 0.0250 & 0.0000 \\
& SummScore - LEAD-3 (\textbf{selected SummScore version})  & 0.0250 & 0.2250 & 0.0000 & \textbf{0.6525} & 0.0544 & 0.0181 & 0.0250 & 0.0000 \\
& SummScore - Salient-R1                                    & 0.1575 & \textbf{0.6525} & 0.0900 & 0.0775 & 0.0025 & 0.0200 & 0.0000 & 0.0000 \\
& SummScore - Salient-R2                                    & 0.2700 & \textbf{0.4950} & 0.1350 & 0.0800 & 0.0050 & 0.0150 & 0.0000 & 0.0000 \\
& SummScore - Salient-RL                                    & 0.3600 & \textbf{0.5400} & 0.0000 & 0.0750 & 0.0050 & 0.0200 & 0.0000 & 0.0000 \\

\midrule 

\multirow{5}{*}{WikiHow} 
& SummScore - Random-3                                         & 0.0600 & \textbf{0.4800} & 0.0600 & 0.3000 & 0.0281 & 0.0469 & 0.0250 & 0.0000 \\
& SummScore - LEAD-3                                           & 0.0000 & 0.1187 & 0.0063 & \textbf{0.7200} & 0.0800 & 0.0000 & 0.0750 & 0.0000 \\
& SummScore - Salient-R1                                       & \textbf{0.4950} & 0.3150 & 0.0900 & 0.0850 & 0.0050 & 0.0100 & 0.0000 & 0.0000 \\ 
& SummScore - Salient-R2 (\textbf{selected SummScore version}) & 0.3825 & \textbf{0.4950} & 0.0225 & 0.0875 & 0.0050 & 0.0075 & 0.0000 & 0.0000 \\
& SummScore - Salient-RL                                       & \textbf{0.4950} & 0.3150 & 0.0900 & 0.0850 & 0.0050 & 0.0100 & 0.0000 & 0.0000 \\

\midrule 

\multirow{5}{*}{SAMSum} 
& SummScore - Random-3 (\textbf{selected SummScore version})   & 0.0000 & 0.0000 & 0.0000 & 0.0925 & 0.3006 & \textbf{0.5319} & 0.0000 & 0.0750 \\
& SummScore - LEAD-3                                           & 0.0000 & 0.0000 & 0.0000 & 0.0750 & 0.3250 & \textbf{0.6000} & 0.0000 & 0.0000 \\
& SummScore - Salient-R1                                       & 0.0000 & 0.0000 & 0.0000 & 0.1500 & 0.2250 & \textbf{0.6250} & 0.0000 & 0.0000 \\
& SummScore - Salient-R2                                       & 0.0000 & 0.0000 & 0.0000 & 0.0250 & 0.2250 & \textbf{0.7500} & 0.0000 & 0.0000 \\
& SummScore - Salient-RL                                       & 0.0000 & 0.0000 & 0.0000 & 0.1500 & 0.2250 & \textbf{0.6250} & 0.0000 & 0.0000 \\

\bottomrule

\end{tabular}
}
\caption{\small Coefficients learned by SummScore with ChatGPT for each feature in each dataset and with each pseudo-labels construction technique. Highest feature values for each model are in bold.}
\label{tab:g2}
\end{table*}

In \Cref{tab:f}, we apply SummScore outside of the standard beam search with 20 beams setup. Results show that SummScore performance continuously improves with more summary candidates, whereas the top beam stays around the same level. Besides, SummScore relative gains are stronger with lower quality decoding methods diverse beam search and nucleus sampling. Lastly, combining 20 summary candidates from each of the three decoding methods yields a pool of 60 summary candidates, out of the which SummScore re-ranking can improve by an extra +0.21 mean {\sc{Rouge}} the performance compared to re-ranking 20 beam search candidates (28.59 mean {\sc{Rouge}} vs 28.38). Overall, we recommend our default setup of beam search with 20 beams to apply SummScore re-ranking. A greater number of beams becomes difficult to fit into a standard GPU with 16 GB memory.

\section{Learned Coefficients}
\label{sec:appendix_g}

In \Cref{tab:g1} (PEGASUS backbone) and \Cref{tab:g2} (ChatGPT backbone), we show coefficients found by SummScore (for each of the five methods to select pseudo-labels which we studided), and on each dataset, including when applying SummScore again on top the self-trained models. For the sake of conciseness, we do not include SummScore coefficients obtained in zero-shot setups. \emph{BERTScore with source} appears as the feature which consistently receives the highest weight for SummScore - Random-3 and SummScore - LEAD-3 ; while \emph{ROUGE-2 with source} dominates for SummScore - Salient-R1/R2/RL. \emph{Diversity} and \emph{Length} features are significantly less used.

\section{Re-ranking Examples}
\label{sec:appendix_h}

In the following, beam search output (for PEGASUS) or the first candidate from top-p sampling (for ChatGPT) is in \orange{orange}, SummScore selected summary candidate in \blue{blue}, and oracle candidate(s) in \teal{teal}. 
On each dataset, we show one re-ranking example on the unsupervised PEGASUS and/or ChatGPT (\Cref{tab:3b}), one zero-shot re-ranking example selected from \Cref{tab:4}, and one re-ranking example applied on top of the self-trained PEGASUS (\Cref{tab:5}).

\begin{table*}
\centering
\resizebox{0.99\textwidth}{!}{
\begin{tabular}{lll}
\toprule
\multicolumn{3}{c}{\textbf{CNN/DM: re-ranking from the unsupervised PEGASUS}} \\

\midrule

\multicolumn{2}{l}{Source} & \begin{tabular}[c]{@{}l@{}}
Royal Dutch Shell Plc said it . has filed a complaint in federal court in Alaska seeking an . order to remove Greenpeace activists who climbed aboard an oil . rig \\
in the Pacific Ocean bound for the Arctic on Monday in a . protest against Arctic drilling. The environmental group said in a statement its team would . occupy the \\
underside of the main deck of the Polar Pioneer, which is under contract to Shell, and plans to unfurl a banner . with the names of millions of people opposed to Arctic \\
drilling. The group said the activists would not interfere with the . vessel's navigation. Scroll down for video . On the rig: Greenpeace activists scale the Polar \\
Pioneer drill rig in the Pacific Ocean . Map: The activists boarded the rig just 750 miles northwest of Hawaii as it makes its journey to the Arctic . At dawn on Monday,\\
the six, from the USA, Germany, New Zealand, Australia, Sweden and Austria, sped towards the Polar Pioneer in inflatable boats launched from the Greenpeace ship Esperanza . \\
Climbers: All Greenpeace activists aboard the rig are experienced climbers and say they don't plan to interfere with the ship's course . 'We're here to highlight that \\
in less than 100 days Shell is . going to the Arctic to drill for oil,' 32-year-old Johno Smith, . one of the six to board the Blue Marlin, the ship carrying the . rig, \\
said in the statement. 'Shell's actions are exploiting the melting ice to increase . a man-made disaster. Climate change is real,' he added. Shell said in an emailed \\
statement that it has met with .groups against oil drilling off Alaska's shores and 'respect . their views' but condemned the boarding. 'We can confirm that protesters \\
from Greenpeace have . illegally boarded the Polar Pioneer, under contract to Shell, jeopardizing not only the safety of the crew on board, but the . protesters themselves,'\\
Shell said. The move comes just days after the U.S. Interior Department . upheld a 2008 lease sale in the Chukchi Sea off Alaska, moving. Shell a step closer to returning\\
to oil and gas exploration in . the Arctic since it suffered mishaps in the region in 2012. The people vs shell: The activists hope they will draw media attention to \\
oil drilling in the Arctic . Reveal a list: Greenpeace activists scale the Polar Pioneer drill rig in the Pacific Ocean to unfurl a banner with the  names of millions \\
of people  opposed to Arctic drilling . Long haul: The activists used ropes and climbing equipment to scale the 38,000-tonne platform . Many environmentalists oppose \\
offshore energy exploration in . the Arctic, saying that once production begins any oil spill . would be extremely difficult to clean up. Oil industry interests say the \\
Arctic will be important to . the United States' energy security in coming decades when output . from shale formations is expected to wane. Images published by Greenpeace\\
showed the activists using . climbing gear to move from an inflatable boat onto the Blue . Marlin heavy-lift vessel towing the Pioneer, one of two drill . rigs heading \\
to the region, as it cruised some 750 miles (1,207 . km) northwest of Hawaii. The six activists planned to camp on the 38,000-tonne Polar Pioneer platform, which they \\
boarded using inflatable boats from he Greenpeace vessel 'Esperanza.' Tweeting from the rig: Aliyah Field tweeted she'd love some coffee but that the sunrise over the \\
Pacific is gorgeous even from the side of the oil rig . Many names: Aliyah maybe referring to the list of names the activists will hang showing all the people who are \\
opposed to oil drilling in the arctic . The six -- from the United States, Germany, New Zealand, Australia, Sweden and Austria -- have supplies for several days and can\\
communicate with the outside world, Greenpeace said. 'We made it! We're on Shell's platform. And we're not alone. Everyone can help turn this into a platform for people \\
power!' tweeted one of the six, Aliyah Field. Johno Smith from New Zealand added: 'We're here to highlight that in less than 100 days Shell is going to the Arctic to \\
drill for oil. 'This pristine environment needs protecting for future generations and all life that will call it home. But instead Shell's actions are exploiting the \\
melting ice to increase a man-made disaster.' A Shell spokeswoman, Kelly op de Weegh, blasted the action. 'We can confirm that protestors from Greenpeace have illegally\\
boarded the 'Polar Pioneer,'under contract to Shell, jeopardizing not only the safety of the crew on board, but the protestors themselves,' she said on Monday. She added:\\
'Shell has met with organizations and individuals who oppose energy exploration offshore Alaska. We respect their views and value the dialogue. 'We will not, however, \\
condone the illegal tactics employed by Greenpeace. Nor will we allow these stunts to distract from preparations underway to execute a safe and responsible exploration \\
program,' she said in a statement.
\end{tabular} \\

\cdashline{1-3}

Beam \#1 & Summary & \begin{tabular}[c]{@{}l@{}}
\orange{'We're here to highlight that in less than 100 days Shell is.}
\end{tabular} \\
& Scores & Mean ROUGE: \textbf{6.55} (rank 11) || SummScore rank: \textbf{20}  \\

\cdashline{1-3}

Beam \#2 & Summary & \begin{tabular}[c]{@{}l@{}}
'We're here to highlight that in less than 100 days Shell is.' Greenpeace activists boarded the Polar Pioneer drill rig in the Pacific Ocean to unfurl a banner with \\
the names of millions of people opposed to oil drilling in the Arctic. Greenpeace activists climbed aboard an oil drilling rig off the coast of Alaska on Monday in a \\
protest against Shell's plans to drill for oil in Arctic waters, the environmental group said in a statement. The group said the activists would not interfere with the \\
vessel's navigation, but the company said it had filed a complaint in federal court in Alaska seeking an order to remove the activists.
\end{tabular} \\
& Scores & Mean ROUGE: 43.17 (rank 3) || SummScore rank: 2  \\

\cdashline{1-3}

Beam \#3 & Summary & \begin{tabular}[c]{@{}l@{}}
\blue{'We're here to highlight that in less than 100 days Shell is.' Greenpeace activists boarded the Polar Pioneer drill rig in the Pacific Ocean to unfurl a banner with the}\\
\blue{names of millions of people opposed to oil drilling in the Arctic. Greenpeace activists climbed aboard an oil drilling rig off the coast of Alaska on Monday in a protest}\\
\blue{against Shell's plans to drill for oil in Arctic waters, the environmental group said in a statement. The group said the activists would not interfere with the ship's} \\
\blue{navigation, but the company said it had filed a complaint in federal court in Alaska seeking an order to remove the activists.}
\end{tabular} \\
& Scores & Mean ROUGE: \textbf{42.85} (rank 4) || SummScore rank: \textbf{1} (\textbf{SummScore output}) \\

\cdashline{1-3}

Beam \#4 & Summary & \begin{tabular}[c]{@{}l@{}}
'We're here to highlight that in less than 100 days Shell is.' Greenpeace activists boarded the Polar Pioneer drill rig in the Pacific Ocean to unfurl a banner with the names \\
of millions of people opposed to oil drilling in the Arctic. Greenpeace activists climbed aboard an oil drilling rig off the coast of Alaska on Monday in a protest against Shell's \\
plans to drill for oil in Arctic waters, the environmental group said in a statement. The group said the activists would not interfere with the vessel's navigation, but the company \\
said it had filed a complaint in federal court in Alaska seeking an order to remove the activists from
\end{tabular} \\
& Scores & Mean ROUGE: 43.59 (rank 2) || SummScore rank: 12  \\

\cdashline{1-3}

... & & \\

\cdashline{1-3}

Beam \#10  & Summary          & \begin{tabular}[c]{@{}l@{}}
\teal{'We're here to highlight that in less than 100 days Shell is.' Greenpeace activists boarded the Polar Pioneer drill rig in the Pacific Ocean to unfurl a banner with the names} \\
\teal{of millions of people opposed to oil drilling in the Arctic. Greenpeace activists climbed aboard an oil drilling rig off the coast of Alaska on Monday in a protest against} \\
\teal{Shell's plans to drill for oil in Arctic waters, the environmental group said in a statement. The group said the activists would not interfere with the vessel's navigation,}\\
\teal{but the company said it had filed a complaint in federal court in Alaska seeking an order to remove them from the}
\end{tabular} \\
& Scores & Mean ROUGE: \textbf{43.91} (rank 1) || SummScore rank: \textbf{6}  \\

\cdashline{1-3}

... & & \\

\midrule

\multicolumn{2}{l}{Reference} & \begin{tabular}[c]{@{}l@{}}
Shell has filed a complaint in federal court in Alaska seeking an order to remove Greenpeace activists who climbed aboard an oil rig in the Pacific . \\
The environmental group said in a statement its team would occupy the underside of the main deck of the Polar Pioneer . \\
The six activists are camping on the 38,000-tonne Polar Pioneer platform, which they boarded using inflatable boats from the Greenpeace vessel 'Esperanza' \\
'We made it! We're on Shell's platform. And we're not alone. Everyone can help turn this into a platform for people power!' tweeted Aliyah Field . \\

\end{tabular}   \\
\bottomrule
\end{tabular}
}
\caption{\small SummScore re-ranking applied to the unsupervised PEGASUS with beam search on CNN/DM.}
\label{tab:h_1_a}
\end{table*}

\begin{table*}
\centering
\resizebox{0.99\textwidth}{!}{
\begin{tabular}{lll}
\toprule
\multicolumn{3}{c}{\textbf{CNN/DM: re-ranking from ChatGPT}} \\

\midrule

\multicolumn{2}{l}{Source} & \begin{tabular}[c]{@{}l@{}}
Although Hillary Clinton boasts a robust 3.6 million Twitter followers, not even a vast right-wing conspiracy would be able to interact with 2 million of them. According\\
to two popular online measuring tools, no more than 44 per cent of her Twitter fan base consists of real people who are active in using the social media platform. And at\\
least 15 per cent – more than 544,000 – are completely fake. StatusPeople.com, the oldest publicly available Twitter-auditing tool, reports that 44 per cent of the former\\
secretary of state's followers are 'good'; 15 per cent are 'fake'; and 41 per cent are 'inactive,' meaning that they never tweet or reply to any tweets. SCROLL DOWN FOR \\
VIDEO . FAKERS: According to one popular online audit tool, only 44 per cent of Hillary Clinton's Twitter followers are real people who participate on the social media \\
platform . 'I'M RUNNING FOR PRESIDENT': Clinton has cast herself as a champion of 'everyday Americans' Another Twitter sleuthing website sampled more than 320,000 of \\
Clinton's followers and found that a much larger number of them were 'fake' Just 4 per cent of President Barack Obama's Twitter followers, by comparison, are considered\\
fake. The White House worked overtime to purge most of them after a September 2013 report found that more than half of his followers didn't really exist. Michelle Obama's\\
Twitter audience is 25 per cent fake, according to StatusPeople, along with 21 per cent of Vice President Joe Biden's. Another tool, TwitterAudit.com, sampled 320,000 \\
of Mrs. Clinton's followers and found that 18 per cent were fake. The new measurements will add to the Clinton presidential campaign's embarrassment following news on \\
Tuesday that a large number of her Facebook fans may represent 'likes' that were purchased rather than earned. REALLY? Hillary Clinton's Twitter follower-count appears \\
to be significantly inflated . FACEBOOK FAKERY: Clinton boasts two-thirds of a million Facebook 'likes,' but more than 46,000 of them list 'Baghdad' as their hometown .\\
Vocativ reported that at least 7 per cent of them listed Baghdad, Iraq as their hometown, a larger number than any U.S. city. That would represent more than 46,000 people.\\
Additional evidence of digital astroturfing is that while most of her U.S. Facebook fans are older than 55, most of the Baghdad contingent is in the 18-34 age range. \\
While Clinton was America's top diplomat, her State Department was buying Facebook 'likes,'according to an Associated Press report from last year. 'In 2013, the State\\
Department, which has more than 400,000 likes and was recently most popular in Cairo, said it would stop buying Facebook fans after its inspector general criticized the\\
agency for spending \$630,000 to boost the numbers,' the wire service reported. Clinton left the agency while the inspector general was investigating. ASTROTURF? A woman\\
cast as a kindly, gardening grandma in Clinton's campaign launch video is actually a former abortion lobbyist who once helmed a campaign for Texas state Senator \\
Wendy Davis . WHERE'S SCOOBY? Clinton's caravan was spotted in Maumee, Ohio on Monday when she lunched at Chipotle, but no one recognized her . Facebook says it has \\
entire teams of employees tasked with rooting out schemes to artificially boost statistics. 'When we catch fraudulent activity, we work to counter and prevent it – including\\
blocking accounts and removing fake likes,' the company reports on its website. Clinton is in Monticello, Iowa on Tuesday for the first stop on her nascent presidential\\
'listening' tour. She was driven from New York in a van that her campaign aides have dubbed 'Scooby,' after the 'Mystery Machine' van in the Scooby-Doo cartoon series. \\
Daily Mail Online visited Kirkwood Community College on Monday, the site of the event, and heard from students that the former first lady is a 'control freak' who may \\
be pursuing legal status for illegal immigrants because the Democratic Party needs a new pool of loyal voters.
\end{tabular} \\

\cdashline{1-3}

Candidate \#1 & Summary & \begin{tabular}[c]{@{}l@{}}
\orange{Online measuring tools show that only 44\% of Hillary Clinton's Twitter followers are real people who actively use the platform, with 15\% completely fake.} \\
\orange{Comparatively, just 4\% of President Obama's Twitter followers are fake. Clinton's Facebook audience has also come under scrutiny, with reports of purchased likes and} \\
\orange{a large number of followers from Baghdad, Iraq.}
\end{tabular} \\
& Scores & Mean ROUGE: \textbf{23.88} (rank 9) || SummScore rank: \textbf{18}  \\

\cdashline{1-3}

Candidate \#2 & Summary & \begin{tabular}[c]{@{}l@{}}
At least 15\% of Hillary Clinton's Twitter followers are fake, according to two online measuring tools, with more than 544,000 completely fake. In comparison, just 4\% of \\
Barack Obama's Twitter followers are fake. Clinton's recent embarrassment of having a large number of Facebook likes purchased rather than earned further adds to the \\
digital astroturfing speculation.
\end{tabular} \\
& Scores & Mean ROUGE: 19.60 (rank 26) || SummScore rank: 15  \\

\cdashline{1-3}

Candidate \#3 & Summary & \begin{tabular}[c]{@{}l@{}}
An investigation has revealed that at least 15\% of Hillary Clinton's Twitter followers are completely fake. The same investigation also found that only 44\% of her Twitter\\
followers are real people who actively use the platform. This adds to the Clinton presidential campaign's embarrassment following news on Tuesday that a large number of her\\
Facebook fans may represent 'likes' that were purchased rather than earned.
\end{tabular} \\
& Scores & Mean ROUGE: 23.88 (rank 8) || SummScore rank: 5 \\

\cdashline{1-3}

Candidate \#4 & Summary & \begin{tabular}[c]{@{}l@{}}
According to two online measuring tools, only 44\% of Hillary Clinton's 3.6 million Twitter followers are real people who actively use the platform, with at least 15\% being \\
fake. In comparison, only 4\% of President Barack Obama's Twitter followers are considered fake. Clinton's Facebook page has also been called into question, with evidence of \\
'likes' being purchased rather than earned.
\end{tabular} \\
& Scores & Mean ROUGE: 27.09 (rank 3) || SummScore rank: 10  \\

\cdashline{1-3}

Candidate \#5 & Summary & \begin{tabular}[c]{@{}l@{}}
According to online measuring tools, over half of Hillary Clinton's Twitter followers are either fake or inactive. One tool reported that only 44\% of her followers are real \\
people who are active on the platform. Additionally, a large number of Clinton's Facebook fans may have been purchased rather than earned, with over 46,000 of them listing Baghdad,\\
Iraq as their hometown.
\end{tabular} \\
& Scores & Mean ROUGE: 24.82 (rank 7) || SummScore rank: 7  \\

\cdashline{1-3}

... & & \\

\cdashline{1-3}

Candidate \#17  & Summary & \begin{tabular}[c]{@{}l@{}}
\blue{According to online measuring tools, only 44\% of Hillary Clinton's 3.6 million Twitter followers consist of real people who are active on the social media platform. At least}\\
\blue{15\% of her followers, or more than 544,000, are completely fake. This comes after news that a large number of her Facebook fans may represent 'likes' that were purchased rather}\\
\blue{than earned.}
\end{tabular} \\
& Scores & Mean ROUGE: \textbf{31.34} (rank 1) || SummScore rank: \textbf{1} (\textbf{SummScore output}) \\

\cdashline{1-3}

... & & \\

\midrule

\multicolumn{2}{l}{Reference} & \begin{tabular}[c]{@{}l@{}}
Two different online audit tools say no more than 44 per cent of Hillary's 3.6 million Twitter fans are real people who participate in the platform . \\
The newly minted presidential candidate is fending off accusations that her Facebook page is full of fake 'likes' \\
Her Facebook fan base includes more people from Baghdad, Iraq than any US city . \\
When she was secretary of state, her agency paid \$630,000 to bulk up its Facebook likes, but pledged to stop after she left .

\end{tabular}   \\
\bottomrule
\end{tabular}
}
\caption{\small SummScore re-ranking applied to ChatGPT with top-p sampling on CNN/DM.}
\label{tab:h_1_b}
\end{table*}

\begin{table*}
\centering
\resizebox{0.99\textwidth}{!}{
\begin{tabular}{lll}
\toprule
\multicolumn{3}{c}{\textbf{CNN/DM: re-ranking from the PEGASUS trained on WikiHow}} \\

\midrule

\multicolumn{2}{l}{Source} & \begin{tabular}[c]{@{}l@{}}
Assault: Dr Sahar Hussain attacked two Tube workers because she didn't want to miss the last train home . A GP attacked two Tube workers while screaming 'I'm a doctor' because \\
she did not want to miss the last train home on a Friday night. Dr Sahar Hussain, 53, panicked when she was unable to get through the gates at Leicester Square station, and started\\
ranting at staff. She denied assaulting the two workers, saying she was worried about being stranded on her own in central London because she is a Muslim woman. But Hussain has\\
now been found guilty and ordered to pay a total of £2,250 in fines, compensation and court costs - and she could face disciplinary action from the General Medical Council. In \\
video footage captured on her own mobile phone, Hussain could be heard to shout: 'I'm a doctor actually, I work for the NHS. I'm a doctor. Get me through the gate, I'm going to\\
miss my train.' City of London Magistrates' Court heard Hussain arrived at the station around 11.30pm on June 20 last year, trying to get home to Woodford Green after socialising\\
with friends in the West End. When she was refused entry by the automatic gates, she demanded that ticket seller Malcolm Shaw let her through before lashing out at his colleague\\
Indira Ramsaroop, who was trying to help. Hussain, originally from Iraq, screamed and shouted at Mrs Ramsaroop as she thrust a camera phone into her face before grabbing her by \\
the arm. The 24-year-old Transport for London worker was then chased by the doctor as she tried to flee to the control room, bumping her head on the way. In the video on Hussain's\\
phone she was heard shouting: 'This woman is on something, she's not sober is she? You're in work and you're not sober. Get me through the gate.' During the scuffle Hussain, a \\
mother of one who helps train GPs at two universities, also grabbed Mr Shaw by the arms, leaving him with scratches. Mrs Ramsaroop was close to tears in court as she told how she\\
had to take almost two weeks off work following the incident, adding: 'I had a lot of sleepless nights. It had an impact on myself with customers when I came back to work. 'I have\\
felt very let down to have been threatened and been running away in my place of work. It actually affected me for a very long time and I got quite ill just at the worrying and \\
fear.' Row: The assault took place on a Friday night at Leicester Square station in central London . Hussain admitted losing her temper, telling the court: 'I'm very sorry about\\
the way I expressed myself with my agitation and frustration.' District Judge Quentin Purdy found her guilty of two counts of assault by beating, saying: 'The evidence is overwhelming.\\
You completely lost your self-control. 'Unusually for the sort of incident this court regularly deals with, there is no hint of intoxication in any way. 'But the explanation is\\
there. You were anxious about missing the last train, concerned about being late for work or unable to work the next day, cultural embarrassment by being left in London as a \\
married Muslim woman. 'It was completely out of character, but as we have all seen, you acted in an unpleasant and hysterical fashion towards numerous members of staff.' He said\\
she had subjected Mrs Ramsaroop to 'completely unjustified violence', as well as 'humiliation' for falsely suggesting she was intoxicated at work. The judge added: 'You have lost \\
a great deal by your lack of self-control. Your good name is now lost with these two convictions for assault.' Hussain was fined £500 for each assault and ordered her to pay £500\\
compensation to Mrs Ramsaroop, £200 to Mr Shaw, £500 prosecution costs and a £50 victim surcharge.

\end{tabular} \\

\cdashline{1-3}

Beam \#1 & Summary & \begin{tabular}[c]{@{}l@{}}
\orange{Try to get through the gate because you didn't want to miss the last train home. Attack the workers because you didn't want to miss the last train home.}
\end{tabular} \\
& Scores & Mean ROUGE: \textbf{13.80} (rank 6) || SummScore rank: \textbf{17}  \\

\cdashline{1-3}

Beam \#2 & Summary & \begin{tabular}[c]{@{}l@{}}
Tell the court she was worried about being stranded on her own in central London because she is a Muslim woman. Describe to the court how she screamed: 'I'm a doctor, get me \\
through the gate.'
\end{tabular} \\
& Scores & Mean ROUGE: 40.97 (rank 2) || SummScore rank: 2  \\

\cdashline{1-3}

Beam \#3 & Summary & \begin{tabular}[c]{@{}l@{}}
Tell the court she was worried about being stranded on her own in central London because she is a Muslim woman. Describe to the court how she screamed: 'Get me through the \\
gate, get me through the gate.'
\end{tabular} \\
& Scores & Mean ROUGE: 39.29 (rank 4) || SummScore rank: 10  \\

\cdashline{1-3}

Beam \#4 & Summary & \begin{tabular}[c]{@{}l@{}}
Try to get through the gate because you don't want to miss the last train home. Attack the workers because you didn't want to miss the last train home.
\end{tabular} \\
& Scores & Mean ROUGE: 13.80 (rank 6) || SummScore rank: 18  \\

\cdashline{1-3}

Beam \#5 & Summary & \begin{tabular}[c]{@{}l@{}}
Try to get through a gate because you don't want to miss the last train home. Attack the workers because you don't want to miss the last train home.
\end{tabular} \\
& Scores & Mean ROUGE: 13.09 (rank 7) || SummScore rank: 20  \\

\cdashline{1-3}

Beam \#6 & Summary & \begin{tabular}[c]{@{}l@{}}
Try to get through the gate because you don't want to miss the last train home. Attack the workers because you don't want to miss the last train home.
\end{tabular} \\
& Scores & Mean ROUGE: 13.80 (rank 6) || SummScore rank: 19  \\

\cdashline{1-3}

Beam \#7 & Summary & \begin{tabular}[c]{@{}l@{}}
Tell the court she was worried about being stranded on her own in central London because she is a Muslim woman. Describe to the court how she yelled: 'Get me through\\
the gate, get me through the gate.'
\end{tabular} \\
& Scores & Mean ROUGE: 39.29 (rank 3) || SummScore rank: 8  \\

\cdashline{1-3}

... & & \\

\cdashline{1-3}

Beam \#10  & Summary          & \begin{tabular}[c]{@{}l@{}}
\blue{Tell the court she was worried about being stranded on her own in central London because she is a Muslim woman. Describe to the court how she screamed: 'I'm a doctor,} \\
\blue{get me through the gate!'}
\end{tabular} \\
& Scores & Mean ROUGE: \textbf{40.97} (rank 2) || SummScore rank: \textbf{1} (\textbf{SummScore output}) \\

\cdashline{1-3}

... & & \\

\cdashline{1-3}

Beam \#14  & Summary & \begin{tabular}[c]{@{}l@{}}
\teal{Tell the court she was worried about being stranded on her own in central London because she is a Muslim woman. Describe to the court how she screamed: 'I'm a doctor,} \\
\teal{get through the gate.'}
\end{tabular} \\
& Scores & Mean ROUGE: \textbf{42.04} (rank 1) || SummScore rank: \textbf{9}  \\

\cdashline{1-3}

... & & \\

\midrule

\multicolumn{2}{l}{Reference} & \begin{tabular}[c]{@{}l@{}}
Shell has filed a complaint in federal court in Alaska seeking an order to remove Greenpeace activists who climbed aboard an oil rig in the Pacific . \\
The environmental group said in a statement its team would occupy the underside of the main deck of the Polar Pioneer . \\
The six activists are camping on the 38,000-tonne Polar Pioneer platform, which they boarded using inflatable boats from the Greenpeace vessel 'Esperanza' \\
'We made it! We're on Shell's platform. And we're not alone. Everyone can help turn this into a platform for people power!' tweeted Aliyah Field . \\

\end{tabular}   \\
\bottomrule
\end{tabular}
}
\caption{\small SummScore re-ranking applied to the PEGASUS fine-tuned on WikiHow with beam search on CNN/DM.}
\label{tab:h_1_c}
\end{table*}

\begin{table*}
\centering
\resizebox{0.99\textwidth}{!}{
\begin{tabular}{lll}
\toprule
\multicolumn{3}{c}{\textbf{CNN/DM: re-ranking from the self-trained PEGASUS}} \\

\midrule

\multicolumn{2}{l}{Source} & \begin{tabular}[c]{@{}l@{}}
Grandparents have pleaded for the safe return to Australia of two young children whose mother took them from Melbourne to the Islamic State capital in Syria.\\
Former Melbourne woman Dullel Kassab fled to Raqqa in Syria with her children last year, and she regularly boasts on Twitter that her four-year-old daughter\\
and two-year-old son sleep with toy guns next to their beds and her daughter likes watching IS videos of 'Muslims killing bad ppl.' The children's paternal \\
grandparents say they are worried Kassab, 28, is 'brainwashing' the children, after their father was killed near the Syria-Turkey border last year, The Herald\\
Sun reported. Former Melbourne woman Dullel Kassab fled to Raqqa in Syria from Melbourne with her children last year . Kassab posts pictures to Twitter of \\
airstrikes hitting blocks away from their Raqqa apartment . 'We miss the children a lot. Their safety and religion has been compromised and we are deeply \\
worried but unable to do anything about it,' a family spokesman told the Herald Sun. 'We pray they come back but it does not look good.' Kassab's Twitter paints\\
a picture of their life in the city the terrorist group IS have made their headquarters, where the children cannot go to school and airstrikes hit blocks away \\
from their apartment. The 28-year-old has a new husband, as the Islamic State does not permit unmarried foreign women to stay in Raqqa. In social media posts \\
she boasts about her children's distaste for Kuffar (non-believers). A photo of another airstrike a day later.  The children's paternal grandparents say they\\
are worried Kassab, 28, is 'brainwashing' the children, after their father was killed near the Syria-Turkey border last year . On her Twitter account she boasts\\
about her children's distaste for Kuffar (non-believers) 'My 4y/o encouraging her little bro to eat his eggs – "C'mon eat ur eggs so u can be big \& strong \& \\
fight the Kuffar!" Allah yehmikum! [sic]' she wrote in December. '\#Awkward Just asked my 4yo wat she wants 2 watch.. "Muslims killing bad ppl" (i.e. \#IS vids \\
obv not beheading ones) LOL [sic],' she wrote in October. Kassab has also complained the 12 to 17-year-olds are now regarded as children when 'in the past they\\
were warriors'. And during the Sydney Lindt café siege in December last year she sent a series of tweets joking that it was exciting. 'This is the most excitement\\
Sydney has seen since the 2000 Olympics!' she posted. Kassab also posts pictures of the Islamic State capital - including this of a 'double rainbow' And during\\
the Sydney Lindt café siege last year Kassab sent a series of tweets joking that it was 'exciting' 'I guess attack the coffee shop wasn't a bad idea, It's a \\
long night… One needs caffeine and chocolate!! [sic]' Kassab also posts pictures of the Islamic State capital, and of Nutella and Twix and Snickers chocolate \\
bars with the caption: 'Im really appreciating \#globalization right about now! \#SimplePleasures Another reason to love \#IS [sic].' The 28-year-old's father \\
Jalal Kassab said he was worried about his grandchildren living in a war zone, but said the threat of imprisonment made it difficult for his daughter to return\\
to Australia. 'I know she wants to come back and we are trying everything we can to bring her back,' Mr Kassab told the Herald Sun. Another former Melbourne \\
woman Zehra Duman last month shared a series of propaganda pictures she says shows her 'five star jihad' lifestyle . In photographs posted to a Twitter several\\
women are pictured standing under an Islamic State flag, reclining against a clean white BMW M5 and wielding machine guns . In one tweet, Duman said: 'US + \\
Australia, how does it feel that all 5 of us were born n raised in your lands, \& now here thirsty for ur blood?'

\end{tabular} \\

\cdashline{1-3}

Beam \#1 & Summary & \begin{tabular}[c]{@{}l@{}}
\orange{The children's paternal grandparents say they are worried Kassab, 28, is 'brainwashing' the children, after their father was killed near the Syria-Turkey border}\\
\orange{last year. 'We pray they come back but it does not look good.' Kassab's Twitter paints a picture of their life in the city the terrorist group IS have made their}\\
\orange{headquarters, where the children cannot go to school and airstrikes hit blocks away from their apartment.}
\end{tabular} \\
& Scores & Mean ROUGE: \textbf{14.89} (rank 4) || SummScore rank: \textbf{6}  \\

\cdashline{1-3}

Beam \#2 & Summary & \begin{tabular}[c]{@{}l@{}}
'We pray they come back but it does not look good.' Kassab's Twitter paints a picture of their life in the city the terrorist group IS have made their headquarters, \\
where the children cannot go to school and airstrikes hit blocks away from their apartment. The children's paternal grandparents say they are worried Kassab, 28, is \\
'brainwashing' the children, after their father was killed near the Syria-Turkey border last year.
\end{tabular} \\
& Scores & Mean ROUGE: 14.89 (rank 4) || SummScore rank: 11  \\

\cdashline{1-3}

Beam \#3 & Summary & \begin{tabular}[c]{@{}l@{}}
The children's paternal grandparents say they are worried Kassab, 28, is 'brainwashing' the children, after their father was killed near the Syria-Turkey border last \\
year, The Herald Sun reported. 'We pray they come back but it does not look good.' Kassab's Twitter paints a picture of their life in the city the terrorist group IS \\
have made their headquarters, where the children cannot go to school and airstrikes hit blocks away from their apartment.
\end{tabular} \\
& Scores & Mean ROUGE: 14.41 (rank 6) || SummScore rank: 5  \\

\cdashline{1-3}

Beam \#4 & Summary & \begin{tabular}[c]{@{}l@{}}
'We pray they come back but it does not look good.' Kassab's Twitter paints a picture of their life in the city the terrorist group IS have made their headquarters, \\
where the children cannot go to school and airstrikes hit blocks away from their apartment. 'My 4y/o encouraging her little bro to eat his eggs $–$ 'C'mon eat ur eggs \\
so u can be big \& strong \& fight the Kuffar!' Allah yehmikum!
\end{tabular} \\
& Scores & Mean ROUGE: 9.92 (rank 10) || SummScore rank: 13  \\

\cdashline{1-3}

... & & \\

\cdashline{1-3}

Beam \#9  & Summary          & \begin{tabular}[c]{@{}l@{}}
\blue{Former Melbourne woman Dullel Kassab fled to Raqqa in Syria with her children last year, and she regularly boasts on Twitter that her four-year-old daughter and} \\
\blue{two-year-old son sleep with toy guns next to their beds and her daughter likes watching IS videos of 'Muslims killing bad ppl.' The children's paternal grandparents} \\
\blue{say they are worried Kassab, 28, is 'brainwashing' the children, after their father was killed near the Syria-Turkey border last year.}
\end{tabular} \\
& Scores & Mean ROUGE: \textbf{57.48} (rank 1) || SummScore rank: \textbf{1} (\textbf{SummScore output}) \\

\cdashline{1-3}

... & & \\

\midrule

\multicolumn{2}{l}{Reference} & \begin{tabular}[c]{@{}l@{}}
Grandparents have pleaded for the safe return of two children in Syria . \\
Former Melbourne woman Dullel Kassab fled to Raqqa in Syria with her four-year-old daughter and two-year-old son last year .\\
She said her daughter likes watching IS videos of 'Muslims killing bad ppl'

\end{tabular}   \\
\bottomrule
\end{tabular}
}
\caption{\small Self-trained PEGASUS with beam search on CNN/DM.}
\label{tab:h_1_d}
\end{table*}

\begin{table*}
\centering
\resizebox{0.99\textwidth}{!}{
\begin{tabular}{lll}
\toprule
\multicolumn{3}{c}{\textbf{XSum: re-ranking from the unsupervised PEGASUS}} \\

\midrule

\multicolumn{2}{l}{Source} & \begin{tabular}[c]{@{}l@{}}
Acting Taoiseach Enda Kenny of Fine Gael and Micheál Martin of  Fianna Fáil hope to avoid a second election. Mr Martin has said his party will facilitate a minority\\
government, but will not support a programme for government. Water charges are one of the main sticking points to reaching agreement. A commission to consider the \\
future of national water utility Irish Water is one of the proposals being considered. Fianna Fáil want to see the immediate removal of water charges, but Fine Gael\\
see a role for them. Following the election, almost two months ago, Fine Gael had 50 seats, Fianna Fáil 44, Sinn Féin 23 and the Labour Party got seven. But no party\\
was able to form a majority government and TDs have so far failed to elect a taoiseach.

\end{tabular} \\

\cdashline{1-3}

Beam \#1 & Summary & \begin{tabular}[c]{@{}l@{}}
\orange{Fianna Fil want to see the immediate removal of water charges, but Fine Gael see a role for them.}
\end{tabular} \\
& Scores & Mean ROUGE: \textbf{8.77} (rank 5) || SummScore rank: \textbf{14}  \\

\cdashline{1-3}

Beam \#2 & Summary & \begin{tabular}[c]{@{}l@{}}
Following the election, almost two months ago, Fine Gael had 50 seats, Fianna Fil 44, Sinn Féin 23 and the Labour Party got seven.
\end{tabular} \\
& Scores & Mean ROUGE: 6.06 (rank 9) || SummScore rank: 6  \\

\cdashline{1-3}

Beam \#3 & Summary & \begin{tabular}[c]{@{}l@{}}
Acting Taoiseach Enda Kenny of Fine Gael and Michel Martin of Fianna Fil hope to avoid a second election.
\end{tabular} \\
& Scores & Mean ROUGE: 7.02 (rank 7) || SummScore rank: 15  \\

\cdashline{1-3}

Beam \#4 & Summary & \begin{tabular}[c]{@{}l@{}}
After the election, almost two months ago, Fine Gael had 50 seats, Fianna Fil 44, Sinn Féin 23 and the Labour Party got seven.
\end{tabular} \\
& Scores & Mean ROUGE: 6.06 (rank 9) || SummScore rank: 7  \\

\cdashline{1-3}

Beam \#5 & Summary & \begin{tabular}[c]{@{}l@{}}
The election, almost two months ago, Fine Gael had 50 seats, Fianna Fil 44, Sinn Féin 23 and the Labour Party got seven.
\end{tabular} \\
& Scores & Mean ROUGE: 6.20 (rank 8) || SummScore rank: 12  \\

\cdashline{1-3}

Beam \#6 & Summary & \begin{tabular}[c]{@{}l@{}}
A commission to consider the future of national water utility Irish Water is one of the proposals being considered. Fianna Fil want to see the immediate removal\\
of water charges, but Fine Gael see a role for them.
\end{tabular} \\
& Scores & Mean ROUGE: 10.53 (rank 4) || SummScore rank: 4  \\

\cdashline{1-3}

Beam \#7 & Summary & \begin{tabular}[c]{@{}l@{}}
Mr Martin has said his party will facilitate a minority government, but will not support a programme for government. Fianna Fil want to see the immediate removal\\
of water charges, but Fine Gael see a role for them.
\end{tabular} \\
& Scores & Mean ROUGE: 17.63 (rank 3) || SummScore rank: 2  \\

\cdashline{1-3}

Beam \#8 & Summary & \begin{tabular}[c]{@{}l@{}}
following the election, almost two months ago, Fine Gael had 50 seats, Fianna Fil 44, Sinn Féin 23 and the Labour Party got seven.
\end{tabular} \\
& Scores & Mean ROUGE: 6.06 (rank 9) || SummScore rank: 8  \\

\cdashline{1-3}

Beam \#9 & Summary & \begin{tabular}[c]{@{}l@{}}
Follow the election, almost two months ago, Fine Gael had 50 seats, Fianna Fil 44, Sinn Féin 23 and the Labour Party got seven.
\end{tabular} \\
& Scores & Mean ROUGE: 6.06 (rank 9) || SummScore rank: 13  \\

\cdashline{1-3}

Beam \#10 & Summary & \begin{tabular}[c]{@{}l@{}}
During the election, almost two months ago, Fine Gael had 50 seats, Fianna Fil 44, Sinn Féin 23 and the Labour Party got seven.
\end{tabular} \\
& Scores & Mean ROUGE: 6.06 (rank 9) || SummScore rank: 9  \\

\cdashline{1-3}

Beam \#11 & Summary & \begin{tabular}[c]{@{}l@{}}
acting Taoiseach Enda Kenny of Fine Gael and Michel Martin of Fianna Fil hope to avoid a second election.
\end{tabular} \\
& Scores & Mean ROUGE: 7.02 (rank 7) || SummScore rank: 20  \\

\cdashline{1-3}

Beam \#12 & Summary & \begin{tabular}[c]{@{}l@{}}
Fianna Fil wants to see the immediate removal of water charges, but Fine Gael see a role for them.
\end{tabular} \\
& Scores & Mean ROUGE: 8.77 (rank 5) || SummScore rank: 16  \\

\cdashline{1-3}

Beam \#13 & Summary & \begin{tabular}[c]{@{}l@{}}
\blue{Mr Martin has said his party will facilitate a minority government, but will not support a programme for government. Fianna Fil want to see the immediate} \\
\blue{removal of water charges, but Fine Gael see a role for them. However, no party was able to form a majority government and TDs have so far failed}
\end{tabular} \\
& Scores & Mean ROUGE: \textbf{19.28} (rank 2) || SummScore rank: \textbf{1} (\textbf{SummScore output}) \\

\cdashline{1-3}

Beam \#14 & Summary & \begin{tabular}[c]{@{}l@{}}
While Fianna Fil want to see the immediate removal of water charges, but Fine Gael see a role for them.
\end{tabular} \\
& Scores & Mean ROUGE: 8.55 (rank 6) || SummScore rank: 19  \\

\cdashline{1-3}

Beam \#15 & Summary & \begin{tabular}[c]{@{}l@{}}
Fianna Fil wanted to see the immediate removal of water charges, but Fine Gael see a role for them.
\end{tabular} \\
& Scores & Mean ROUGE: 8.77 (rank 5) || SummScore rank: 17  \\

\cdashline{1-3}

Beam \#16 & Summary & \begin{tabular}[c]{@{}l@{}}
\teal{Mr Martin has said his party will facilitate a minority government, but will not support a programme for government.}
\end{tabular} \\
& Scores & Mean ROUGE: \textbf{21.25} (rank 1) || SummScore rank: \textbf{10}  \\

\cdashline{1-3}

Beam \#17 & Summary & \begin{tabular}[c]{@{}l@{}}
Fianna Fil hope to see the immediate removal of water charges, but Fine Gael see a role for them.
\end{tabular} \\
& Scores & Mean ROUGE: 8.77 (rank 5) || SummScore rank: 18  \\

\cdashline{1-3}

Beam \#18 & Summary & \begin{tabular}[c]{@{}l@{}}
Mr Martin has said his party will facilitate a minority government, but will not support a programme for government. Fianna Fil want to see the immediate \\
removal of water charges, but Fine Gael see a role for them. However, no party was able to form a majority government and TDs so far failed to
\end{tabular} \\
& Scores & Mean ROUGE: 19.28 (rank 2) || SummScore rank: 3  \\

\cdashline{1-3}

Beam \#19 & Summary & \begin{tabular}[c]{@{}l@{}}
Following the election, almost two months ago, Fine Gael had 50 seats, Fianna Fil 44, Sinn Féin 23 and the Labour Party got seven..
\end{tabular} \\
& Scores & Mean ROUGE: 6.06 (rank xx) || SummScore rank: 11  \\

\cdashline{1-3}

Beam \#20 & Summary & \begin{tabular}[c]{@{}l@{}}
Mr Martin has said his party will facilitate a minority government, but will not support a programme for government. Fianna Fil want to see the immediate \\
removal of water charges, but Fine Gael see a role for them. However, no party was able to form a majority government and TDs will so far failed
\end{tabular} \\
& Scores & Mean ROUGE: 19.28 (rank 2) || SummScore rank: 5  \\

\midrule

\multicolumn{2}{l}{Reference} & \begin{tabular}[c]{@{}l@{}}
Negotiators from Ireland's two biggest parties are to meet for a final effort to form a minority government.

\end{tabular}   \\
\bottomrule
\end{tabular}
}
\caption{\small SummScore re-ranking applied to the unsupervised PEGASUS with beam search on XSum.}
\label{tab:h_2_a}
\end{table*}

\begin{table*}
\centering
\resizebox{0.99\textwidth}{!}{
\begin{tabular}{lll}
\toprule
\multicolumn{3}{c}{\textbf{XSum: re-ranking from the BART trained on WikiHow}} \\

\midrule

\multicolumn{2}{l}{Source} & \begin{tabular}[c]{@{}l@{}}
Cardiff narrowly missed out on the Championship play-offs last term, finishing eighth in the table. But under new head coach Paul Trollope, Pilkington thinks \\
the Welsh side are capable of better. "We fell just short [last season] and we were all really disappointed with that. We felt we had the squad to get in the \\
play-offs definitely," he said. "It's going to be a tough season again, just like it was last year, but hopefully we can be right up there come the end of the \\
season." Pilkington, 28, scored Cardiff's first goal in their 4-0 friendly win at Shrewsbury on Saturday. The Republic of Ireland international played in attack,\\
having moved from the wing to a central forward role last season. Pilkington made the transition under former Cardiff boss Russell Slade, who was replaced by \\
Trollope in May. Trollope had been working under Slade before then, and Pilkington believes that smooth transition has helped the Bluebirds in their preparations \\
for the new campaign. "The gaffer was working with us last season so we know how he works," he added. "We know what he's about and we know we have to be on our \\
toes all the time. "He's come in and changed a few things. It's been good up to now so hopefully we can carry on working really well."

\end{tabular} \\

\cdashline{1-3}

Beam \#1 & Summary & \begin{tabular}[c]{@{}l@{}}
\orange{Aim to improve on last season's finish., Expect a tough season.}
\end{tabular} \\
& Scores & Mean ROUGE: \textbf{5.13} (rank 11) || SummScore rank: \textbf{12}  \\

\cdashline{1-3}

Beam \#2 & Summary & \begin{tabular}[c]{@{}l@{}}
Aim to improve on last season's finish.,Pilkington believes the Bluebirds are capable of reaching the play-offs.
\end{tabular} \\
& Scores & Mean ROUGE: 24.63 (rank 2) || SummScore rank: 5  \\

\cdashline{1-3}

Beam \#3 & Summary & \begin{tabular}[c]{@{}l@{}}
Aim to improve on last season's finish., Believe the Bluebirds have the squad to get back up the table.
\end{tabular} \\
& Scores & Mean ROUGE: 17.89 (rank 7) || SummScore rank: 6  \\

\cdashline{1-3}

Beam \#4 & Summary & \begin{tabular}[c]{@{}l@{}}
Reach the play-offs again.
\end{tabular} \\
& Scores & Mean ROUGE: 7.02 (rank 10) || SummScore rank: 18  \\

\cdashline{1-3}

Beam \#5 & Summary & \begin{tabular}[c]{@{}l@{}}
\blue{Aim to improve on last season's finish.,Pilkington believes the Bluebirds have the squad to get back up the table.}
\end{tabular} \\
& Scores & Mean ROUGE: \textbf{23.20} (rank 4) || SummScore rank: \textbf{1} (\textbf{SummScore output}) \\

\cdashline{1-3}

Beam \#6 & Summary & \begin{tabular}[c]{@{}l@{}}
Aim to improve on last season's finish.,Pilkington believes the Bluebirds have the squad to reach the play-offs.
\end{tabular} \\
& Scores & Mean ROUGE: 23.90 (rank 3) || SummScore rank: 2  \\

\cdashline{1-3}

Beam \#7 & Summary & \begin{tabular}[c]{@{}l@{}}
Expect to improve on last season's finish.,Pilkington believes the Bluebirds have the squad to get back up the table.
\end{tabular} \\
& Scores & Mean ROUGE: 23.20 (rank 4) || SummScore rank: 3  \\

\cdashline{1-3}

Beam \#8 & Summary & \begin{tabular}[c]{@{}l@{}}
\teal{Aim to improve on last season's finish.,Pilkington believes the Bluebirds have the squad to challenge for promotion.}
\end{tabular} \\
& Scores & Mean ROUGE: \textbf{41.06} (rank 1) || SummScore rank: \textbf{7}  \\

\cdashline{1-3}

Beam \#9 & Summary & \begin{tabular}[c]{@{}l@{}}
Aim to improve on last season's finish.,Pilkington believes the Bluebirds are capable of reaching the play-offs again.
\end{tabular} \\
& Scores & Mean ROUGE: 23.90 (rank 3) || SummScore rank: 4  \\

\cdashline{1-3}

Beam \#10 & Summary & \begin{tabular}[c]{@{}l@{}}
Aim to improve on last season's finish., Believe in the squad.
\end{tabular} \\
& Scores & Mean ROUGE: 12.82 (rank 9) || SummScore rank: 9  \\

\cdashline{1-3}

Beam \#11 & Summary & \begin{tabular}[c]{@{}l@{}}
Aim to improve on last season's finish., Expect a tough season again.
\end{tabular} \\
& Scores & Mean ROUGE: 4.94 (rank 12) || SummScore rank: 8  \\

\cdashline{1-3}

Beam \#12 & Summary & \begin{tabular}[c]{@{}l@{}}
Aim to improve on last season's finish., Believe in the squad.
\end{tabular} \\
& Scores & Mean ROUGE: 12.82 (rank 9) || SummScore rank: 11  \\

\cdashline{1-3}

Beam \#13 & Summary & \begin{tabular}[c]{@{}l@{}}
Aim to improve on last season's finish., Expect to challenge for promotion again.
\end{tabular} \\
& Scores & Mean ROUGE: 21.79 (rank 6) || SummScore rank: 16  \\

\cdashline{1-3}

Beam \#14 & Summary & \begin{tabular}[c]{@{}l@{}}
Prepare for a tough season.
\end{tabular} \\
& Scores & Mean ROUGE: 14.04 (rank 8) || SummScore rank: 13  \\

\cdashline{1-3}

Beam \#15 & Summary & \begin{tabular}[c]{@{}l@{}}
Aim to improve on last season's finish., Believe in the squad.
\end{tabular} \\
& Scores & Mean ROUGE: 12.82 (rank 9) || SummScore rank: 10  \\

\cdashline{1-3}

Beam \#16 & Summary & \begin{tabular}[c]{@{}l@{}}
Prepare for a tough season.
\end{tabular} \\
& Scores & Mean ROUGE: 14.04 (rank 8) || SummScore rank: 14  \\

\cdashline{1-3}

Beam \#17 & Summary & \begin{tabular}[c]{@{}l@{}}
Aim to improve on last season's finish., Expect to challenge for promotion.
\end{tabular} \\
& Scores & Mean ROUGE: 22.62 (rank 5) || SummScore rank: 17  \\

\cdashline{1-3}

Beam \#18 & Summary & \begin{tabular}[c]{@{}l@{}}
Prepare for a tough season.
\end{tabular} \\
& Scores & Mean ROUGE: 14.04 (rank 8) || SummScore rank: 15  \\

\cdashline{1-3}

Beam \#19 & Summary & \begin{tabular}[c]{@{}l@{}}
Prepare for another tough season.
\end{tabular} \\
& Scores & Mean ROUGE: 14.04 (rank 8) || SummScore rank: 20  \\

\cdashline{1-3}

Beam \#20 & Summary & \begin{tabular}[c]{@{}l@{}}
Prepare for another tough season.
\end{tabular} \\
& Scores & Mean ROUGE: 14.04 (rank 8) || SummScore rank: 19  \\

\midrule

\multicolumn{2}{l}{Reference} & \begin{tabular}[c]{@{}l@{}}
Cardiff City forward Anthony Pilkington believes the Bluebirds can challenge for promotion this season.

\end{tabular}   \\
\bottomrule
\end{tabular}
}
\caption{\small SummScore re-ranking applied to the BART fine-tuned on WikiHow with beam search on XSum.}
\label{tab:h_2_b}
\end{table*}

\begin{table*}
\centering
\resizebox{0.99\textwidth}{!}{
\begin{tabular}{lll}
\toprule
\multicolumn{3}{c}{\textbf{XSum: re-ranking from the self-trained PEGASUS}} \\

\midrule

\multicolumn{2}{l}{Source} & \begin{tabular}[c]{@{}l@{}}
In a statement seen by the BBC, his wife said he had been out drinking with another man in the city of Yangon on Friday night before his death. The Foreign \\
Office said it was investigating the death of a British national and was providing support to the victim's family. News agency AFP quoted police as saying the \\
dead man was a 47-year-old who had suffered wounds to his head and chest. According to reports, the man - who had been working at an international school - \\
was found dead at an apartment in Yangon, previously known as Rangoon. The Foreign Office says it is working with local authorities in Myanmar, formerly Burma,\\
to determine what happened.
\end{tabular} \\

\cdashline{1-3}

Beam \#1 & Summary & \begin{tabular}[c]{@{}l@{}}
\orange{The Foreign Office said it was investigating the death of a British national and was providing support to the victim's family.}
\end{tabular} \\
& Scores & Mean ROUGE: \textbf{13.07} (rank 10) || SummScore rank: \textbf{3}  \\

\cdashline{1-3}

Beam \#2 & Summary & \begin{tabular}[c]{@{}l@{}}
In a statement seen by the BBC, his wife said he had been out drinking with another man in the city of Yangon on Friday night before his death.
\end{tabular} \\
& Scores & Mean ROUGE: 10.85 (rank 12) || SummScore rank: 4  \\

\cdashline{1-3}

Beam \#3 & Summary & \begin{tabular}[c]{@{}l@{}}
According to reports, the man - who had been working at an international school - was found dead at an apartment in Yangon, previously known as Rangoon.
\end{tabular} \\
& Scores & Mean ROUGE: 20.61 (rank 8) || SummScore rank: 7  \\

\cdashline{1-3}

Beam \#4 & Summary & \begin{tabular}[c]{@{}l@{}}
\teal{The man, who has not been named, was found dead at an apartment in Yangon, Myanmar, on Saturday.}
\end{tabular} \\
& Scores & Mean ROUGE: \textbf{31.39} (rank 1) || SummScore rank: \textbf{14}  \\

\cdashline{1-3}

Beam \#5 & Summary & \begin{tabular}[c]{@{}l@{}}
The man, who has not been named, was found dead at an apartment in Yangon, formerly known as Rangoon, on Saturday.
\end{tabular} \\
& Scores & Mean ROUGE: 24.88 (rank 6) || SummScore rank: 12  \\

\cdashline{1-3}

Beam \#6 & Summary & \begin{tabular}[c]{@{}l@{}}
According to reports, the man - who had been working at an international school - was found dead at an apartment in Yangon, formerly known as Rangoon.
\end{tabular} \\
& Scores & Mean ROUGE: 20.61 (rank 8) || SummScore rank: 5  \\

\cdashline{1-3}

Beam \#7 & Summary & \begin{tabular}[c]{@{}l@{}}
\blue{The man, who has not been named, was found dead at an apartment in Yangon, previously known as Rangoon.}
\end{tabular} \\
& Scores & Mean ROUGE: \textbf{26.39} (rank 4) || SummScore rank: \textbf{1} (\textbf{SummScore output}) \\

\cdashline{1-3}

Beam \#8 & Summary & \begin{tabular}[c]{@{}l@{}}
The man, who has not been named, was found dead at an apartment in Yangon, formerly known as Rangoon.
\end{tabular} \\
& Scores & Mean ROUGE: 26.39 (rank 4) || SummScore rank: 2  \\

\cdashline{1-3}

Beam \#9 & Summary & \begin{tabular}[c]{@{}l@{}}
The man, who has not been named, was found dead at an apartment in Yangon, previously known as Rangoon, on Saturday.
\end{tabular} \\
& Scores & Mean ROUGE: 24.88 (rank 6) || SummScore rank: 11  \\

\cdashline{1-3}

Beam \#10 & Summary & \begin{tabular}[c]{@{}l@{}}
The Foreign Office said it was working with local authorities in Myanmar, formerly Burma, to determine what happened.
\end{tabular} \\
& Scores & Mean ROUGE: 12.64 (rank 11) || SummScore rank: 10  \\

\cdashline{1-3}

Beam \#11 & Summary & \begin{tabular}[c]{@{}l@{}}
The Foreign Office says it is working with local authorities in Myanmar, formerly Burma, to determine what happened.
\end{tabular} \\
& Scores & Mean ROUGE: 12.64 (rank 11) || SummScore rank: 11  \\

\cdashline{1-3}

Beam \#12 & Summary & \begin{tabular}[c]{@{}l@{}}
The man, who has not been named, was found dead at an apartment in Yangon, formerly Burma, on Saturday.
\end{tabular} \\
& Scores & Mean ROUGE: 26.39 (rank 4) || SummScore rank: 18  \\

\cdashline{1-3}

Beam \#13 & Summary & \begin{tabular}[c]{@{}l@{}}
Media playback is unsupported on your device 11 August 2015 Last updated at 08:00 BST The Foreign Office said it was investigating the death of a British national in the city of Yangon.
\end{tabular} \\
& Scores & Mean ROUGE: 9.78 (rank 13) || SummScore rank: 19  \\

\cdashline{1-3}

Beam \#14 & Summary & \begin{tabular}[c]{@{}l@{}}
Media playback is unsupported on your device 11 August 2015 Last updated at 08:00 BST The man, who has not been named, was found dead at an apartment in Yangon.
\end{tabular} \\
& Scores & Mean ROUGE: 19.33 (rank 9) || SummScore rank: 20  \\

\cdashline{1-3}

Beam \#15 & Summary & \begin{tabular}[c]{@{}l@{}}
The man, who has not been named, was found dead at an apartment in Yangon, the capital of Myanmar, on Saturday.
\end{tabular} \\
& Scores & Mean ROUGE: 28.69 (rank 2) || SummScore rank: 16  \\

\cdashline{1-3}

Beam \#16 & Summary & \begin{tabular}[c]{@{}l@{}}
According to reports, the man - who had been working at an international school - was found dead at an apartment in Yangon, previously known as Burma.
\end{tabular} \\
& Scores & Mean ROUGE: 20.61 (rank 8) || SummScore rank: 15  \\

\cdashline{1-3}

Beam \#17 & Summary & \begin{tabular}[c]{@{}l@{}}
The man, who has not been named, was found dead at an apartment in the city of Yangon on Saturday.
\end{tabular} \\
& Scores & Mean ROUGE: 25.61 (rank 5) || SummScore rank: 13  \\

\cdashline{1-3}

Beam \#18 & Summary & \begin{tabular}[c]{@{}l@{}}
The Foreign Office said the man, who has not been named, was found dead at an apartment in Yangon, previously known as Rangoon.
\end{tabular} \\
& Scores & Mean ROUGE: 23.53 (rank 7) || SummScore rank: 9  \\

\cdashline{1-3}

Beam \#19 & Summary & \begin{tabular}[c]{@{}l@{}}
The Foreign Office said the man, who has not been named, was found dead at an apartment in Yangon, formerly known as Rangoon.
\end{tabular} \\
& Scores & Mean ROUGE: 23.53 (rank 7) || SummScore rank: 8  \\

\cdashline{1-3}

Beam \#20 & Summary & \begin{tabular}[c]{@{}l@{}}
The man, who has not been named, was found dead at an apartment in Yangon on Saturday.
\end{tabular} \\
& Scores & Mean ROUGE: 28.11 (rank 3) || SummScore rank: 17  \\

\midrule

\multicolumn{2}{l}{Reference} & \begin{tabular}[c]{@{}l@{}}
A British man believed to be a teacher has been found dead in Myanmar.

\end{tabular}   \\
\bottomrule
\end{tabular}
}
\caption{\small Self-trained PEGASUS with beam search on XSum.}
\label{tab:h_2_c}
\end{table*}

\begin{table*}
\centering
\resizebox{0.99\textwidth}{!}{
\begin{tabular}{lll}
\toprule
\multicolumn{3}{c}{\textbf{WikiHow: re-ranking from the unsupervised PEGASUS}} \\

\midrule

\multicolumn{2}{l}{Source} & \begin{tabular}[c]{@{}l@{}}
On your scanner tool, set the PPI (Pixel per inch) to 350 so that it will create a high-quality image.; , Click on the program’s shortcut icon on your desktop or programs list \\
to launch it. Wait until the program starts. Any version will do, but the latest one is better because they have additional useful functions. , On the Menu bar located at the\\
top of the screen, go to File and click “Open.” Locate the image, select it, then click “Open.” , To change it, go to “Image” on the Menu bar then click “Mode” and choose “RGB.”\\
To turn your scanned image fully grayscale, just in case you didn’t change it on your scanner setting and the image has traces of colors, change the Adjustment, under “Image”\\
on the Menu bar, into “Black and White.” , On the Layer panel on the palette dock located at the bottom left of the screen, create a new layer by clicking a small paper-like \\
icon beside the folder icon. , Change the layer name to “Skin” as this will be the first part of the image that you will add a color to. , Press “B” on your keyboard or get\\
the Brush tool on the toolbar located at the left side of the screen. The icon is a brush. , Pick a nice skin tone color on the Swatches, which are also on the Palette dock. \\
You may also set the foreground color to get more colors on the Toolbar, below the Zoom Tool, which is the two square that indicates as foreground and background color. A skin \\
tone color is the lightest Orange or Yellow Orange. , Drag and move your mouse to the skin tone of your image. To avoid misplacing the colors, use the Zoom tool or simply \\
press CTRL + Z. Use the Eraser tool, which is also located on the toolbar, to delete some misplaced part; but change the Brush Preset, which is on the Palette Dock, and choose \\
“Soft Round” so it will create pixelated color when you erase them. , Once you’re done with the skin, repeat Steps 1 to 6 with other parts of the body. For example, if you \\
want to color the clothes, add a layer and rename the layer “Clothes.” After renaming, proceed with selecting the right tool and color, and coloring the clothes. Continue with\\
other parts, like lips, hair, background, etc. , To merge all layers, go to Layer dock, right-click, and choose “Flatten Image” or “Merge Visible” on the options. , Adjust \\
the image tone, level, or brightness on the Image Menu to enhance the photo. , Press CTRL + S, choose JPEG as file type, add a file name, and click “Save” when you’re done!

\end{tabular} \\

\cdashline{1-3}

Beam \#1 & Summary & \begin{tabular}[c]{@{}l@{}}
\orange{You may also set the foreground color to get more colors on the Toolbar, below the Zoom Tool, which is the two square that indicates as foreground and background color.}
\end{tabular} \\
& Scores & Mean ROUGE: \textbf{13.01} (rank 7) || SummScore rank: \textbf{20}  \\

\cdashline{1-3}

Beam \#2 & Summary & \begin{tabular}[c]{@{}l@{}}
\teal{, On your scanner tool, set the PPI (Pixel per inch) to 350 so that it will create a high-quality image.;, Click on the program’s shortcut icon on your desktop or programs} \\
\teal{list to launch it., To change it, go to “Image” on the Menu bar then click “Mode” and choose “RGB.” To turn your scanned image fully grayscale, just in case you didn’t change}\\
\teal{it on your scanner setting and the image has traces of colors, change the Adjustment, under “Image”—on the Palette dock located at the bottom left of the screen—into “}
\end{tabular} \\
& Scores & Mean ROUGE: \textbf{23.18} (rank 1) || SummScore rank: \textbf{2}  \\

\cdashline{1-3}

Beam \#3 & Summary & \begin{tabular}[c]{@{}l@{}}
\teal{, On your scanner tool, set the PPI (Pixel per inch) to 350 so that it will create a high-quality image.;, Click on the program’s shortcut icon on your desktop or programs} \\
\teal{list to launch it., To change it, go to “Image” on the Menu bar then click “Mode” and choose “RGB.” To turn your scanned image fully grayscale, just in case you didn’t change}\\
\teal{it on your scanner setting and the image has traces of colors, change the Adjustment, under “Image”—on the Palette dock located at the bottom left of the screen, into “}
\end{tabular} \\
& Scores & Mean ROUGE: \textbf{23.18} (rank 1) || SummScore rank: \textbf{5}  \\

\cdashline{1-3}

Beam \#4 & Summary & \begin{tabular}[c]{@{}l@{}}
, On your scanner tool, set the PPI (Pixel per inch) to 350 so that it will create a high-quality image.;, Click on the program’s shortcut icon on your desktop or programs list \\
to launch it., To change it, go to “Image” on the Menu bar then click “Mode” and choose “RGB.” To turn your scanned image fully grayscale, just in case you didn’t change it\\
on your scanner setting and the image has traces of colors, change the Adjustment, under “Image”—on the menu bar, into “Black and White.”,On the
\end{tabular} \\
& Scores & Mean ROUGE: 21.05 (rank 5) || SummScore rank: 11  \\

\cdashline{1-3}

Beam \#5 & Summary & \begin{tabular}[c]{@{}l@{}}
, On your scanner tool, set the PPI (Pixel per inch) to 350 so that it will create a high-quality image.;, Click on the program’s shortcut icon on your desktop or programs list \\
to launch it., To change it, go to “Image” on the Menu bar then click “Mode” and choose “RGB.” To turn your scanned image fully grayscale, just in case you didn’t change it\\
on your scanner setting and the image has traces of colors, change the Adjustment, under “Image”— on the menu bar, into “Black and White.”,On the
\end{tabular} \\
& Scores & Mean ROUGE: 21.05 (rank 5) || SummScore rank: 13  \\

\cdashline{1-3}

Beam \#6 & Summary & \begin{tabular}[c]{@{}l@{}}
, On your scanner tool, set the PPI (Pixel per inch) to 350 so that it will create a high-quality image.;, Click on the program’s shortcut icon on your desktop or programs list \\
to launch it., To change it, go to “Image” on the Menu bar then click “Mode” and choose “RGB.” To turn your scanned image fully grayscale, just in case you didn’t change it\\
on your scanner setting and the image has traces of colors, change the Adjustment, under “Image”—on the Palette dock located at the bottom left of the screen—to “
\end{tabular} \\
& Scores & Mean ROUGE: 22.83 (rank 2) || SummScore rank: 4  \\

\cdashline{1-3}

Beam \#7 & Summary & \begin{tabular}[c]{@{}l@{}}
, On your scanner tool, set the PPI (Pixel per inch) to 350 so that it will create a high-quality image.;, Click on the program’s shortcut icon on your desktop or programs list \\
to launch it., To change it, go to “Image” on the Menu bar then click “Mode” and choose “RGB.” To turn your scanned image fully grayscale, just in case you didn’t change it\\
on your scanner setting and the image has traces of colors, change the Adjustment, under “Image”—on the menu bar—into “Black and White.”,On the
\end{tabular} \\
& Scores & Mean ROUGE: 21.05 (rank 5) || SummScore rank: 7  \\

\cdashline{1-3}

Beam \#8 & Summary & \begin{tabular}[c]{@{}l@{}}
\teal{On your scanner tool, set the PPI (Pixel per inch) to 350 so that it will create a high-quality image.;, Click on the program’s shortcut icon on your desktop or programs list} \\
\teal{to launch it., To change it, go to “Image” on the Menu bar then click “Mode” and choose “RGB.” To turn your scanned image fully grayscale, just in case you didn’t change it}\\
\teal{on your scanner setting and the image has traces of colors, change the Adjustment, under “Image”—on the Palette dock located at the bottom left of the screen— into “}
\end{tabular} \\
& Scores & Mean ROUGE: \textbf{23.18} (rank 1) || SummScore rank: \textbf{3}  \\

\cdashline{1-3}

... & & \\

\cdashline{1-3}

Beam \#11 & Summary & \begin{tabular}[c]{@{}l@{}}
\blue{, On your scanner tool, set the PPI (Pixel per inch) to 350 so that it will create a high-quality image.;, Click on the program’s shortcut icon on your desktop or programs} \\
\blue{list to launch it., To change it, go to “Image” on the Menu bar then click “Mode” and choose “RGB.” To turn your scanned image fully grayscale, just in case you didn’t change}\\
\blue{it on your scanner setting and the image has traces of colors, change the Adjustment, under “Image”—on the Palette dock located at the bottom left of the screen—and choose}
\end{tabular} \\
& Scores & Mean ROUGE: \textbf{22.71} (rank 3) || SummScore rank: \textbf{1} (\textbf{SummScore output}) \\

\cdashline{1-3}

... & & \\

\midrule

\multicolumn{2}{l}{Reference} & \begin{tabular}[c]{@{}l@{}}
Negotiators from Ireland's two biggest parties are to meet for a final effort to form a minority government.

\end{tabular}   \\
\bottomrule
\end{tabular}
}
\caption{\small SummScore re-ranking applied to the unsupervised PEGASUS with beam search on WikiHow.}
\label{tab:h_3_a}
\end{table*}

\begin{table*}
\centering
\resizebox{0.99\textwidth}{!}{
\begin{tabular}{lll}
\toprule
\multicolumn{3}{c}{\textbf{WikiHow: re-ranking from the PEGASUS trained on CNN/DM}} \\

\midrule

\multicolumn{2}{l}{Source} & \begin{tabular}[c]{@{}l@{}}
Gently stabilize it by holding it steady with one or both hands. Pet your cat and talk to it in a soothing voice to calm and reassure it.If your cat resists you or is prone\\
to scratching, then wrap your cat in the towel so that just its head is sticking out.  , Once your cat is calm, place your non-dominant hand on top of your cat’s head in front\\
of its ears. Wrap your fingers around the bottom of its cheekbones for leverage., This should cause your cat’s mouth to open involuntarily. Pick up the pill with your other \\
hand. Hold the pill between your index finger and thumb. Then place your middle or ring finger on the lower molars to keep its jaw open. Do not place your finger on top of \\
the canine tooth, i.e., the sharp fang, to keep its jaw open.If your cat will not open its mouth, then you will need to pry it open. Do this using the middle finger of the \\
same hand holding the pill. Place your middle finger over the small incisor teeth in the front of your cat’s mouth at the bottom. Then apply gentle pressure to push your \\
cat’s jaw open.  , Drop the pill as far back as you can on your cat’s tongue, i.e., the base of the tongue. Or, use your index finger and thumb to push the pill over the back\\
of your cat’s tongue. If you do it this way, you must do it quickly to prevent your fingers from getting bitten.Alternatively, you can use a pill syringe to place the pill at\\
the base of your cat’s tongue. This way you can avoid having to put your fingers in your cat’s mouth. Coat the pill with butter to make swallowing it easier.  , Once you have\\
placed the pill at the base of its tongue, use your hands to keep its mouth closed by applying gentle pressure. Also re-position your cat's head so that it is level instead \\
of tilted back; this will make it easier for your cat to swallow the pill. Gently rub your cat’s throat or blow in its nose to encourage it to swallow the pill.Try not to let\\
your cat go until the pil lis swallowed. Place a drop of water on its nose. If your cat licks the water off, then this means that it has swallowed the pill.  , Tricyclic \\
antidepressants should be given with food to decrease side effects such as vomiting. Therefore, feed your cat directly after giving it its medication. If your cat will not \\
eat, then entice it to eat by feeding it its favorite food like cheese, or by giving it a treat.Make sure to provide your cat with water as well. You can also give your cat\\
flavored water like juice from a tuna can to entice it to drink.

\end{tabular} \\

\cdashline{1-3}

Beam \#1 & Summary & \begin{tabular}[c]{@{}l@{}}
\orange{Pet your cat and talk to it in a soothing voice to calm and reassure it. If your cat resists you or is prone to scratching, then wrap your cat in a towel so that just its}\\
\orange{head is sticking out.}
\end{tabular} \\
& Scores & Mean ROUGE: \textbf{17.17} (rank 8) || SummScore rank: \textbf{12}  \\

\cdashline{1-3}

Beam \#2 & Summary & \begin{tabular}[c]{@{}l@{}}
Pet your cat and talk to it in a soothing voice to calm and reassure it. If your cat resists you or is prone to scratching, then wrap your cat in the towel so that just its \\
head is sticking out.
\end{tabular} \\
& Scores & Mean ROUGE: 18.78 (rank 6) || SummScore rank: 8  \\

\cdashline{1-3}

Beam \#3 & Summary & \begin{tabular}[c]{@{}l@{}}
Pet your cat and talk to it in a soothing voice to calm and reassure it. If your cat resists you or is prone to scratching, then wrap your cat in a towel so that just its \\
head is sticking out.
\end{tabular} \\
& Scores & Mean ROUGE: 17.17 (rank 8) || SummScore rank: 11  \\

\cdashline{1-3}

Beam \#4 & Summary & \begin{tabular}[c]{@{}l@{}}
Pet your cat and talk to it in a soothing voice to calm and reassure it.  gently stabilize it by holding it steady with one or both hands. Rub your cat’s throat or blow in \\
its nose to encourage it to swallow the pill.
\end{tabular} \\
& Scores & Mean ROUGE: 18.68 (rank 7) || SummScore rank: 7  \\

\cdashline{1-3}

Beam \#5 & Summary & \begin{tabular}[c]{@{}l@{}}
\blue{Pet your cat and talk to it in a soothing voice to calm and reassure it. Wrap your fingers around the bottom of its cheekbones for leverage. Alternatively, you can use}\\
\blue{a pill syringe to place the pill at the base of your cat’s tongue.}
\end{tabular} \\
& Scores & Mean ROUGE: \textbf{26.93} (rank 3) || SummScore rank: \textbf{1} (\textbf{SummScore output}) \\

\cdashline{1-3}

Beam \#6 & Summary & \begin{tabular}[c]{@{}l@{}}
Tricyclic antidepressants should be given with food to decrease side effects such as vomiting. If your cat resists you or is prone to scratching, then wrap your cat in a towel\\
so that just its head is sticking out.
\end{tabular} \\
& Scores & Mean ROUGE: 15.09 (rank 10) || SummScore rank: 17 \\

\cdashline{1-3}

Beam \#7 & Summary & \begin{tabular}[c]{@{}l@{}}
Pet your cat and talk to it in a soothing voice to calm and reassure it.  gently stabilize it by holding it steady with one or both hands.
\end{tabular} \\
& Scores & Mean ROUGE: 8.72 (rank 12) || SummScore rank: 19 \\

\cdashline{1-3}

Beam \#8 & Summary & \begin{tabular}[c]{@{}l@{}}
Gently stabilize it by holding it steady with one or both hands. If your cat resists you or is prone to scratching, then wrap your cat in a towel so that just its head is \\
sticking out.
\end{tabular} \\
& Scores & Mean ROUGE: 17.18 (rank 7) || SummScore rank: 14 \\

\cdashline{1-3}

... & & \\

\cdashline{1-3}

Beam \#14 & Summary & \begin{tabular}[c]{@{}l@{}}
\teal{If your cat resists you or is prone to scratching, then wrap your cat in a towel so that just its head is sticking out. Alternatively, you can use a pill syringe to place the}\\
\teal{pill at the base of your cat’s tongue.}
\end{tabular} \\
& Scores & Mean ROUGE: \textbf{30.74} (rank 1) || SummScore rank: \textbf{2}  \\

\cdashline{1-3}

... & & \\

\midrule

\multicolumn{2}{l}{Reference} & \begin{tabular}[c]{@{}l@{}}
Negotiators from Ireland's two biggest parties are to meet for a final effort to form a minority government.

\end{tabular}   \\
\bottomrule
\end{tabular}
}
\caption{\small SummScore re-ranking applied to the PEGASUS fine-tuned on CNN/DM with beam search on WikiHow.}
\label{tab:h_3_b}
\end{table*}

\begin{table*}
\centering
\resizebox{0.99\textwidth}{!}{
\begin{tabular}{lll}
\toprule
\multicolumn{3}{c}{\textbf{WikiHow: re-ranking from the self-trained PEGASUS}} \\

\midrule

\multicolumn{2}{l}{Source} & \begin{tabular}[c]{@{}l@{}}
Talking in private will help you feel less nervous and more comfortable about asking for help.It will also give your teacher more time to help you because they aren’t trying\\
to teach at the same time.   For example, you could say, “When you have time later today, could we talk about a problem I’m having?” If you’re afraid to approach your teacher,\\
put a note on their chair. The note could say, “Can we talk later about something personal? Thanks, Mark.” You could also send your teacher an email or message letting them\\
know you would like to ask their help with a personal issue.  , Sometimes teachers will try to offer advice on what you should do, when all you wanted was for them to listen.\\
If you let your teacher know how they can help, it will make it easier for them to give you the help you want and need.Think about what kind of help you want. Ask yourself, \\
“Do I want her to listen, to give me advice, or to do something about the problem?” Tell your teacher how they can help. For example, “Can you help me come up with ways to \\
make more friends?” If you don’t know how you want your teacher to help, you it’s okay for you to say that, too. Try saying, “I need your help with a problem, but I don’t \\
know what kind of help I need.”  , Also, being honest will help your teacher figure out the right way to help you.   The more truthful information your teacher has about what\\
is going on, the more they will be able to help. If you are afraid you will get in trouble, then say that. You could say, “I need your help with something but I’m scared I’ll\\
get in trouble.” If you are asking for help, but also trying not to get someone else in trouble, you could leave out names, but still be honest about what is going on. For \\
example, “My friend is thinking about cheating on a test and I need advice on how to stop them from doing something so stupid.”

\end{tabular} \\

\cdashline{1-3}

Beam \#1 & Summary & \begin{tabular}[c]{@{}l@{}}
\orange{You could say, “I need your help with something but I’m scared I’ll get in trouble.” If you are asking for help, but also trying not to get someone else in trouble, you could}\\
\orange{leave out names, but still be honest about what is going on. For example, “My friend is thinking about cheating on a test and I need advice on how to stop them from doing} \\
\orange{something so stupid.”}
\end{tabular} \\
& Scores & Mean ROUGE: \textbf{22.54} (rank 14) || SummScore rank: \textbf{14}  \\

\cdashline{1-3}

Beam \#2 & Summary & \begin{tabular}[c]{@{}l@{}}
Think about what kind of help you want. You could say, “I need your help with something but I’m scared I’ll get in trouble.” If you are asking for help, but also trying not \\
to get someone else in trouble, you could leave out names, but still be honest about what is going on.If you’re afraid to approach your teacher, put a note on their chair.
\end{tabular} \\
& Scores & Mean ROUGE: 30.92 (rank 7) || SummScore rank: 18  \\

\cdashline{1-3}

Beam \#3 & Summary & \begin{tabular}[c]{@{}l@{}}
For example, you could say, “When you have time later today, could we talk about a problem I’m having?” If you’re afraid to approach your teacher, put a note on their chair. \\
Try saying, “I need your help with a problem, but I don’t know what kind of help I need.”, Also, being honest will help your teacher figure out the right way to help you.
\end{tabular} \\
& Scores & Mean ROUGE: 29.41 (rank 9) || SummScore rank: 15  \\

\cdashline{1-3}

Beam \#4 & Summary & \begin{tabular}[c]{@{}l@{}}
Think about what kind of help you want. You could say, “I need your help with something but I’m scared I’ll get in trouble.” If you are asking for help, but also trying not to\\
get someone else in trouble, you could leave out names, but still be honest about what is going on.
\end{tabular} \\
& Scores & Mean ROUGE: 28.40 (rank 10) || SummScore rank: 20  \\

\cdashline{1-3}

Beam \#5 & Summary & \begin{tabular}[c]{@{}l@{}}
Think about what kind of help you want. You could say, “I need your help with something but I’m scared I’ll get in trouble.” If you are asking for help, but also trying not to\\
get someone else in trouble, you could leave out names, but still be honest about what is going on. For example, “My friend is thinking about cheating on a test and I need advice\\
on how to stop them from doing something so stupid.”
\end{tabular} \\
& Scores & Mean ROUGE: 27.36 (rank 12) || SummScore rank: 11  \\

\cdashline{1-3}

Beam \#6 & Summary & \begin{tabular}[c]{@{}l@{}}
Think about what kind of help you want. You could say, “I need your help with something but I’m scared I’ll get in trouble.” If you are asking for help, but also trying not to\\
get someone else in trouble, you could leave out names, but still be honest about what is going on.You could also send your teacher an email or message letting them know you\\
would like to ask their help with a personal issue.
\end{tabular} \\
& Scores & Mean ROUGE: 31.47 (rank 6) || SummScore rank: 9  \\

\cdashline{1-3}

Beam \#7 & Summary & \begin{tabular}[c]{@{}l@{}}
Think about what kind of help you want. You could say, “I need your help with something but I’m scared I’ll get in trouble.” If you are asking for help, but also trying not to\\
get someone else in trouble, you could leave out names, but still be honest about what is going on.If you are afraid to approach your teacher, put a note on their chair.
\end{tabular} \\
& Scores & Mean ROUGE: 30.92 (rank 7) || SummScore rank: 17  \\

\cdashline{1-3}

Beam \#8 & Summary & \begin{tabular}[c]{@{}l@{}}
You could say, “I need your help with something but I’m scared I’ll get in trouble.” If you are asking for help, but also trying not to get someone else in trouble, you could \\
leave out names, but still be honest about what is going on. You could also send your teacher an email or message letting them know you would like to ask their help with a \\
personal issue.
\end{tabular} \\
& Scores & Mean ROUGE: 28.06 (rank 11) || SummScore rank: 13  \\

\cdashline{1-3}

... & & \\

\cdashline{1-3}

Beam \#19 & Summary & \begin{tabular}[c]{@{}l@{}}
\teal{For example, you could say, “When you have time later today, could we talk about a problem I’m having?” If you’re afraid to approach your teacher, put a note on their chair.} \\
\teal{Try saying, “I need your help with a problem, but I don’t know what kind of help I need.”, Also, being honest will help your teacher figure out the right way to help you.} \\
\teal{The more truthful information your teacher has about what is going on, the more they will be able to help}
\end{tabular} \\
& Scores & Mean ROUGE: \textbf{35.08} (rank 1) || SummScore rank: \textbf{2}  \\

\cdashline{1-3}

Beam \#20 & Summary & \begin{tabular}[c]{@{}l@{}}
\blue{Think about what kind of help you want. You could say, “I need your help with something but I’m scared I’ll get in trouble.” If you are asking for help, but also trying not} \\
\blue{to get someone else in trouble, you could leave out names, but still be honest about what is going on.If you’re afraid to approach your teacher, put a note on their chair.} \\
\blue{It will also give your teacher more time to help you because they aren’t trying to teach at the same time}
\end{tabular} \\
& Scores & Mean ROUGE: \textbf{34.51} (rank 2) || SummScore rank: \textbf{1} (\textbf{SummScore output}) \\

\midrule

\multicolumn{2}{l}{Reference} & \begin{tabular}[c]{@{}l@{}}
Negotiators from Ireland's two biggest parties are to meet for a final effort to form a minority government.

\end{tabular}   \\
\bottomrule
\end{tabular}
}
\caption{\small Self-trained PEGASUS with beam search on WikiHow.}
\label{tab:h_3_c}
\end{table*}

\begin{table*}
\centering
\resizebox{0.95\textwidth}{!}{
\begin{tabular}{lll}
\toprule
\multicolumn{3}{c}{\textbf{SAMSum: re-ranking from the unsupervised PEGASUS}} \\

\midrule

\multicolumn{2}{l}{Source} & \begin{tabular}[c]{@{}l@{}}
Mike: dude, wendy has grown prettier \\
Dave:<laughing><laughing><thumb\_up> \\
Mike:<love><love> \\
Dave: i know right? \\
Mike: yeah, since she came from Houston, she looks like an angel \\
Dave: i'll have to hit on her soon \\
Mike: haha, stay off, i hear Jerry is her lover \\
Dave: really<sad> \\
Mike: yeah \\
Dave: since when \\
Mike: haha, i dont know, but you can push your luck \\
Dave: haha, i will \\
Mike:<blink> good luck \\
Dave: yeah, ill need it\\
\end{tabular} \\

\cdashline{1-3}

Beam \#1 & Summary & \begin{tabular}[c]{@{}l@{}}
\orange{Mike: dude, wendy has grown prettier Dave: Mike: Dave: i know right?}
\end{tabular} \\
& Scores & Mean ROUGE: \textbf{19.73} (rank 3) || SummScore rank: \textbf{19}  \\

\cdashline{1-3}

Beam \#2 & Summary & \begin{tabular}[c]{@{}l@{}}
Mike: guy, wendy has grown prettier Dave: Mike: Dave: i know right?
\end{tabular} \\
& Scores & Mean ROUGE: 19.73 (rank 3) || SummScore rank: 20  \\

\cdashline{1-3}

Beam \#3 & Summary & \begin{tabular}[c]{@{}l@{}}
Mike: Yeah, since she came from Houston, she looks like an angel Dave: i'll have to hit on her soon \\
Mike: haha, stay off, i hear Jerry is her lover Dave: really Mike: yeah Dave: since when Mike:haha, i dont know, but you can push
\end{tabular} \\
& Scores & Mean ROUGE: 22.31 (rank 2) || SummScore rank: 2  \\

\cdashline{1-3}

Beam \#4   & Summary          & \begin{tabular}[c]{@{}l@{}}
Mike: uh, since she came from Houston, she looks like an angel Dave: i'll have to hit on her soon \\
Mike: haha, stay off, i hear Jerry is her lover Dave: really Mike: yeah Dave: since when Mike:haha, i dont know, but you can push
\end{tabular} \\
& Scores & Mean ROUGE: 22.31 (rank 2) || SummScore rank: 11  \\

\cdashline{1-3}

Beam \#5   & Summary  & \begin{tabular}[c]{@{}l@{}}
Mike: sure, since she came from Houston, she looks like an angel Dave: i'll have to hit on her soon \\
Mike: haha, stay off, i hear Jerry is her lover Dave: really Mike: yeah Dave: since when Mike:haha, i dont know, but you can push
\end{tabular} \\
& Scores & Mean ROUGE: 22.31 (rank 2) || SummScore rank: 13  \\

\cdashline{1-3}

Beam \#6   & Summary          & \begin{tabular}[c]{@{}l@{}}
Mike:yeah, since she came from Houston, she looks like an angel Dave: i'll have to hit on her soon \\
Mike: haha, stay off, i hear Jerry is her lover Dave: really Mike: yeah Dave: since when Mike:haha, i dont know, but you can push
\end{tabular} \\
& Scores & Mean ROUGE: 22.31 (rank 2) || SummScore rank: 6 \\

\cdashline{1-3}

Beam \#7   & Summary          & \begin{tabular}[c]{@{}l@{}}
Mike: yes, since she came from Houston, she looks like an angel Dave: i'll have to hit on her soon \\
Mike: haha, stay off, i hear Jerry is her lover Dave: really Mike: yeah Dave: since when Mike:haha, i dont know, but you can push
\end{tabular} \\
& Scores & Mean ROUGE: 22.31 (rank 2) || SummScore rank: 10  \\

\cdashline{1-3}

Beam \#8   & Summary          & \begin{tabular}[c]{@{}l@{}}
\blue{Mike: Yeah, since she came from Houston, she looks like an angel Dave: i'll have to hit on her soon} \\
\blue{Mike: haha, stay off, i hear Jerry is her lover Dave: really Mike: yeah Dave: since when Mike: hahaha, i dont know, but you can push}
\end{tabular} \\
& Scores & Mean ROUGE: \textbf{22.31} (rank 2) || SummScore rank: \textbf{1 (SummScore output)} \\

\cdashline{1-3}

Beam \#9   & Summary          & \begin{tabular}[c]{@{}l@{}}
Mike: uh, yeah, since she came from Houston, she looks like an angel Dave: i'll have to hit on her soon \\
Mike: haha, stay off, i hear Jerry is her lover Dave: really Mike: yeah Dave: since when Mike is haha, i dont know, but you
\end{tabular} \\
& Scores & Mean ROUGE: 22.31 (rank 2) || SummScore rank: 18 \\

\cdashline{1-3}

Beam \#10  & Summary          & \begin{tabular}[c]{@{}l@{}}
Mike: obviously, since she came from Houston, she looks like an angel Dave: i'll have to hit on her soon \\
Mike: haha, stay off, i hear Jerry is her lover Dave: really Mike: yeah Dave: since when Mike:haha, i dont know, but you can push
\end{tabular} \\
& Scores & Mean ROUGE: 22.31 (rank 2) || SummScore rank: 15  \\

\cdashline{1-3}

Beam \#11  & Summary          & \begin{tabular}[c]{@{}l@{}}
Mike: uh, since she came from Houston, she looks like an angel Dave: i'll have to hit on her soon \\
Mike: haha, stay off, i hear Jerry is her lover Dave: really Mike: yeah Dave: since when Mike: hahaha, i dont know, but you can push
\end{tabular} \\
& Scores & Mean ROUGE: 22.31 (rank 2) || SummScore rank: 9  \\

\cdashline{1-3}

Beam \#12  & Summary          & \begin{tabular}[c]{@{}l@{}}
Mike: sure, since she came from Houston, she looks like an angel Dave: i'll have to hit on her soon \\
Mike: haha, stay off, i hear Jerry is her lover Dave: really Mike: yeah Dave: since when Mike: hahaha, i dont know, but you can push
\end{tabular} \\
& Scores & Mean ROUGE: 22.31 (rank 2) || SummScore rank: 12  \\

\cdashline{1-3}

Beam \#13  & Summary          & \begin{tabular}[c]{@{}l@{}}
\teal{Mike: uh, yeah, since she came from Houston, she looks like an angel Dave: i'll have to hit on her soon} \\
\teal{Mike: haha, stay off, i hear Jerry is her lover Dave: really Mike: yeah Dave: since when Mike:haha, i dont know, but you}
\end{tabular} \\
& Scores & Mean ROUGE: \textbf{22.64} (rank 1) || SummScore rank: \textbf{5}  \\

\cdashline{1-3}

Beam \#14  & Summary          & \begin{tabular}[c]{@{}l@{}}
Mike: yes, since she came from Houston, she looks like an angel Dave: i'll have to hit on her soon \\
Mike: haha, stay off, i hear Jerry is her lover Dave: really Mike: yeah Dave: since when Mike: hahaha, i dont know, but you can push
\end{tabular} \\
& Scores & Mean ROUGE: 22.31 (rank 2) || SummScore rank: 8 \\

\cdashline{1-3}

Beam \#15  & Summary          & \begin{tabular}[c]{@{}l@{}}
Mike: no, since she came from Houston, she looks like an angel Dave: i'll have to hit on her soon \\
Mike: haha, stay off, i hear Jerry is her lover Dave: really Mike: yeah Dave: since when Mike:haha, i dont know, but you can push
\end{tabular} \\
& Scores & Mean ROUGE: 22.31 (rank 2) || SummScore rank: 17  \\

\cdashline{1-3}

Beam \#16  & Summary          & \begin{tabular}[c]{@{}l@{}}
Mike:yeah, since she came from Houston, she looks like an angel Dave: i'll have to hit on her soon \\
Mike: haha, stay off, i hear Jerry is her lover Dave: really Mike: yeah Dave: since when Mike: hahaha, i dont know, but you can push
\end{tabular} \\
& Scores & Mean ROUGE: 22.31 (rank 2) || SummScore rank: 4  \\

\cdashline{1-3}

Beam \#17  & Summary          & \begin{tabular}[c]{@{}l@{}}
Mike: cool, since she came from Houston, she looks like an angel Dave: i'll have to hit on her soon \\
Mike: haha, stay off, i hear Jerry is her lover Dave: really Mike: yeah Dave: since when Mike:haha, i dont know, but you can push
\end{tabular} \\
& Scores & Mean ROUGE: 22.31 (rank 2) || SummScore rank: 14  \\

\cdashline{1-3}

Beam \#18  & Summary          & \begin{tabular}[c]{@{}l@{}}
Mike:Yeah, since she came from Houston, she looks like an angel Dave: i'll have to hit on her soon \\
Mike: haha, stay off, i hear Jerry is her lover Dave: really Mike: yeah Dave: since when Mike:haha, i dont know, but you can push
\end{tabular} \\
& Scores & Mean ROUGE: 22.31 (rank 2) || SummScore rank: 7 \\

\cdashline{1-3}

Beam \#19  & Summary          & \begin{tabular}[c]{@{}l@{}}
\teal{Mike: uh, yeah, since she came from Houston, she looks like an angel Dave: i'll have to hit on her soon} \\
\teal{Mike: haha, stay off, i hear Jerry is her lover Dave: really Mike: yeah Dave: since when Mike: hahaha, i dont know, but you}
\end{tabular} \\
& Scores & Mean ROUGE: \textbf{22.64} (rank 1) || SummScore rank: \textbf{3} \\

\cdashline{1-3}

Beam \#20  & Summary          & \begin{tabular}[c]{@{}l@{}}
\teal{Mike: uh, since she came from Houston, she looks like an angel Dave: i'll have to hit on her soon} \\
\teal{Mike: haha, stay off, i hear Jerry is her lover Dave: really Mike: yeah Dave: since when Mike:haha, i don't know, but you}
\end{tabular} \\
& Scores & Mean ROUGE: \textbf{22.64} (rank 1) || SummScore rank: \textbf{16}  \\

\midrule

\multicolumn{2}{l}{Reference} & \begin{tabular}[c]{@{}l@{}}
Mike and Dave notice Wendy got prettier. Dave wants to hit on her, but she's with Jerry. He'll try anyway.
\end{tabular}   \\
\bottomrule
\end{tabular}
}
\caption{\small SummScore re-ranking applied to the unsupervised PEGASUS with beam search on SAMSum.}
\label{tab:h_4_a}
\end{table*}

\begin{table*}
\centering
\resizebox{0.99\textwidth}{!}{
\begin{tabular}{lll}
\toprule
\multicolumn{3}{c}{\textbf{SAMSum: re-ranking from ChatGPT}} \\

\midrule

\multicolumn{2}{l}{Source} & \begin{tabular}[c]{@{}l@{}}
Adam: I have a juicy gossip for you guys ;) \\
Borys: About whom? \\
Adam: So apparently, Iga planned a weekend getaway with her boyfriend, but they had to cancel because he couldn't talk his group into changing the date of the presentation :D \\
Katia: And she is pissed? \\
Adam: What do you think? That is why they aren't sitting together since Monday ;) \\
Katia: Who would have expected such a gossip girl from you :D
\end{tabular} \\

\cdashline{1-3}

Candidate \#1 & Summary & \begin{tabular}[c]{@{}l@{}}
\orange{Adam shares a gossip about Iga's cancelled weekend getaway with her boyfriend due to his group's presentation, which has caused tension between them. Katia is surprised by Adam's}\\
\orange{gossiping behavior.}
\end{tabular} \\
& Scores & Mean ROUGE: \textbf{19.80} (rank 16) || SummScore rank: \textbf{7}  \\

\cdashline{1-3}

Candidate \#2 & Summary & \begin{tabular}[c]{@{}l@{}}
Adam shares juicy gossip about Iga's cancelled weekend getaway with her boyfriend due to a presentation, which caused her to be upset and avoid sitting with her boyfriend since Monday.\\
Katia jokes about Adam's unexpected role as a gossip girl.
\end{tabular} \\
& Scores & Mean ROUGE: 23.92 (rank 13) || SummScore rank: 5  \\

\cdashline{1-3}

Candidate \#3 & Summary & \begin{tabular}[c]{@{}l@{}}
Adam shares juicy gossip about Iga's cancelled weekend getaway with her boyfriend due to his group's presentation, causing tension between them, and surprising his friends with his\\
gossiping skills.
\end{tabular} \\
& Scores & Mean ROUGE: 20.38 (rank 15) || SummScore rank: 20 \\

\cdashline{1-3}

Candidate \#4 & Summary & \begin{tabular}[c]{@{}l@{}}
Adam shares gossip about Iga's cancelled weekend getaway with her boyfriend due to his group's presentation, causing tension between them and leading to them not sitting together.\\
Katia is surprised by Adam's gossiping tendencies.
\end{tabular} \\
& Scores & Mean ROUGE: 26.79 (rank 11) || SummScore rank: 11  \\

\cdashline{1-3}

Candidate \#5 & Summary & \begin{tabular}[c]{@{}l@{}}
Adam shares gossip about Iga's cancelled weekend getaway with her boyfriend, causing her to be angry and not sit with him since Monday, surprising his friends.
\end{tabular} \\
& Scores & Mean ROUGE: 29.91 (rank 7) || SummScore rank: 15  \\

\cdashline{1-3}

Candidate \#6 & Summary & \begin{tabular}[c]{@{}l@{}}
Adam shares juicy gossip about Iga's cancelled weekend getaway with her boyfriend due to his group's presentation date, which caused tension between them. Katia is surprised by \\
Adam's gossiping skills.
\end{tabular} \\
& Scores & Mean ROUGE: 19.80 (rank 16) || SummScore rank: 6  \\

\cdashline{1-3}

Candidate \#7 & Summary & \begin{tabular}[c]{@{}l@{}}
Adam shares gossip about Iga's cancelled weekend getaway with her boyfriend due to his group presentation, causing tension between them and resulting in them not sitting together\\
since Monday. Katia is surprised by Adam's gossiping.
\end{tabular} \\
& Scores & Mean ROUGE: 31.28 (rank 6) || SummScore rank: 10  \\

\cdashline{1-3}

Candidate \#8 & Summary & \begin{tabular}[c]{@{}l@{}}
Adam shares gossip about Iga's cancelled weekend getaway with her boyfriend due to his group's presentation, causing tension between them and resulting in them not sitting together\\
since Monday, surprising his friends.
\end{tabular} \\
& Scores & Mean ROUGE: 32.59 (rank 3) || SummScore rank: 19  \\

\cdashline{1-3}

Candidate \#9 & Summary & \begin{tabular}[c]{@{}l@{}}
Adam shares gossip with his friends about Iga's cancelled weekend getaway with her boyfriend, which has caused tension between them. His friends are surprised by his gossiping.
\end{tabular} \\
& Scores & Mean ROUGE: 18.28 (rank 18) || SummScore rank: 3  \\

\cdashline{1-3}

Candidate \#10 & Summary & \begin{tabular}[c]{@{}l@{}}
Adam shares gossip about Iga's canceled weekend getaway with her boyfriend due to his group's presentation date, causing tension between them and resulting in them not sitting \\
together since Monday. Katia teases Adam about his unexpected role as a gossip girl.
\end{tabular} \\
& Scores & Mean ROUGE: 31.42 (rank 5) || SummScore rank: 8  \\

\cdashline{1-3}

Candidate \#11 & Summary & \begin{tabular}[c]{@{}l@{}}
Adam shares juicy gossip that Iga's weekend getaway with her boyfriend was cancelled and she is angry about it, causing her to not sit with him since Monday. Katia is \\
surprised by Adam's gossiping.
\end{tabular} \\
& Scores & Mean ROUGE: 28.05 (rank 9) || SummScore rank: 2  \\

\cdashline{1-3}

Candidate \#12 & Summary & \begin{tabular}[c]{@{}l@{}}
Adam shares juicy gossip about Iga's cancelled weekend getaway with her boyfriend due to a presentation, causing tension between them and leading to them not sitting together. \\
Katia is surprised by Adam's gossiping tendencies.
\end{tabular} \\
& Scores & Mean ROUGE: 27.15 (rank 10) || SummScore rank: 12  \\

\cdashline{1-3}

Candidate \#13 & Summary & \begin{tabular}[c]{@{}l@{}}
Adam shares juicy gossip about Iga's cancelled weekend getaway with her boyfriend due to his work presentation, causing tension in their relationship and leading to them not sitting\\
together. Katia is surprised by Adam's gossiping tendencies.
\end{tabular} \\
& Scores & Mean ROUGE: 25.56 (rank 12) || SummScore rank: 15  \\

\cdashline{1-3}

Candidate \#14 & Summary & \begin{tabular}[c]{@{}l@{}}
Adam shares juicy gossip that Iga's weekend getaway with her boyfriend was cancelled due to a presentation date, leading to Iga being upset and not sitting with her boyfriend.
\end{tabular} \\
& Scores & Mean ROUGE: 31.49 (rank 4) || SummScore rank: 9 \\

\cdashline{1-3}

Candidate \#15 & Summary & \begin{tabular}[c]{@{}l@{}}
Adam shares juicy gossip with his friends about Iga's cancelled weekend getaway with her boyfriend due to work, causing tension between them. His friends are surprised by Adam's gossiping\\
tendencies.
\end{tabular} \\
& Scores & Mean ROUGE: 17.23 (rank 19) || SummScore rank: 13  \\

\cdashline{1-3}

Candidate \#16 & Summary & \begin{tabular}[c]{@{}l@{}}
Adam shares gossip about Iga's cancelled weekend getaway with her boyfriend, causing tension between them, and surprises Katia with his gossiping.
\end{tabular} \\
& Scores & Mean ROUGE: 19.00 (rank 17) || SummScore rank: 18  \\

\cdashline{1-3}

Candidate \#17 & Summary & \begin{tabular}[c]{@{}l@{}}
\teal{Adam shares juicy gossip about Iga's cancelled weekend getaway with her boyfriend due to his work presentation, causing tension between the couple and leading to them not sitting together}\\
\teal{since Monday.}
\end{tabular} \\
& Scores & Mean ROUGE: \textbf{33.53} (rank 1) || SummScore rank: \textbf{17}  \\

\cdashline{1-3}

Candidate \#18 & Summary & \begin{tabular}[c]{@{}l@{}}
\blue{Adam shares juicy gossip about Iga's cancelled weekend getaway with her boyfriend due to his group's presentation, which has caused tension between them and they are not sitting together}\\
\blue{since Monday. Katia is surprised by Adam's gossiping skills.}
\end{tabular} \\
& Scores & Mean ROUGE: \textbf{33.09} (rank 2) || SummScore rank: \textbf{1} (\textbf{SummScore output}) \\

\cdashline{1-3}

Candidate \#19 & Summary & \begin{tabular}[c]{@{}l@{}}
Adam shares juicy gossip that Iga's weekend getaway with her boyfriend got canceled because of his group's presentation and she is now not sitting with him. Katia is surprised by Adam's\\
gossiping skills.
\end{tabular} \\
& Scores & Mean ROUGE: 23.52 (rank 14) || SummScore rank: 4  \\

\cdashline{1-3}

Candidate \#20 & Summary & \begin{tabular}[c]{@{}l@{}}
Adam shares gossip about Iga's cancelled weekend getaway with her boyfriend due to his presentation, causing tension between them and resulting in them not sitting together since Monday, \\
surprising Katia who didn't expect such gossip from Adam.
\end{tabular} \\
& Scores & Mean ROUGE: 28.73 (rank 8) || SummScore rank: 16  \\

\midrule

\multicolumn{2}{l}{Reference} & \begin{tabular}[c]{@{}l@{}}
Iga and her boyfriend are not sitting together since Monday, because they had to cancel their weekend getaway. It made Iga very angry. Iga's boyfriend failed to convince a group to change \\
the date of the presentation.

\end{tabular}   \\
\bottomrule
\end{tabular}
}
\caption{\small SummScore re-ranking applied to ChatGPT with top-p sampling on SAMSum.}
\label{tab:h_4_b}
\end{table*}

\begin{table*}
\centering
\resizebox{0.99\textwidth}{!}{
\begin{tabular}{lll}
\toprule
\multicolumn{3}{c}{\textbf{SAMSum: re-ranking from the PEGASUS trained on XSum}} \\

\midrule

\multicolumn{2}{l}{Source} & \begin{tabular}[c]{@{}l@{}}
Mia: God, he keeps talking about the stupid wall \\
Jennifer: I think he's just senile \\
Karine: unfortunately it's much more \\
Karine: this populist strategy to scare people with a minority \\
Karine: and then present himself as the only savior \\
Peter: Karine is right, I think he's not as senile and stupid as we would like to believe Mia: so it's even worse - he's pure evil \\
Peter: possibly yes \\
Karine: I really think so, because it's not only about politics anymore \\
Karine: he is destroying the Earth \\
Karine: trying to shake the world's order \\
Jennifer: but he's also quite stupid Jennifer: and the dangerous thing is his self confidence \\
Jennifer: he's persistence  \\
Jennifer: and ignorance \\
Karine: but we have a chance to change it \\
Karine: maybe Trump will be over soon? \\
Mia: we will see \\
Mia: but we should do anything to win the next elections \\
Karine: everything!\\
\end{tabular} \\

\cdashline{1-3}

Beam \#1 & Summary & \begin{tabular}[c]{@{}l@{}}
\orange{In our series of letters from British schoolchildren, Mia, Jennifer and Karine give their views on US President Donald Trump.}
\end{tabular} \\
& Scores & Mean ROUGE: \textbf{23.16} (rank 5) || SummScore rank: \textbf{18}  \\

\cdashline{1-3}

Beam \#2 & Summary & \begin{tabular}[c]{@{}l@{}}
In our series of letters from British children, Mia, Jennifer and Karine give their views on US President Donald Trump.
\end{tabular} \\
& Scores & Mean ROUGE: 23.16 (rank 5) || SummScore rank: 17  \\

\cdashline{1-3}

Beam \#3 & Summary & \begin{tabular}[c]{@{}l@{}}
All images are copyrighted.
\end{tabular} \\
& Scores & Mean ROUGE: 00.00 (rank 7) || SummScore rank: 20  \\

\cdashline{1-3}

Beam \#4   & Summary          & \begin{tabular}[c]{@{}l@{}}
In our series of letters from British children, Mia, Jennifer and Karine tell us what they think about US President Donald Trump.
\end{tabular} \\
& Scores & Mean ROUGE: 21.65 (rank 6) || SummScore rank: 6  \\

\cdashline{1-3}

Beam \#5   & Summary  & \begin{tabular}[c]{@{}l@{}}
In our series of letters from British children, Mia, Jennifer, Karine and Peter tell us what they think about Donald Trump.
\end{tabular} \\
& Scores & Mean ROUGE: 36.19 (rank 2) || SummScore rank: 5  \\

\cdashline{1-3}

Beam \#6   & Summary          & \begin{tabular}[c]{@{}l@{}}
In our series of letters from British children, Mia, Jennifer, Karine and Peter tell us what they think about US President Donald Trump.
\end{tabular} \\
& Scores & Mean ROUGE: 33.89 (rank 3) || SummScore rank: 5 \\

\cdashline{1-3}

Beam \#7   & Summary          & \begin{tabular}[c]{@{}l@{}}
In our series of letters from British children, Mia, Jennifer, Karine and Peter tell us what they think of Donald Trump.
\end{tabular} \\
& Scores & Mean ROUGE: 36.19 (rank 2) || SummScore rank: 10  \\

\cdashline{1-3}

Beam \#8   & Summary          & \begin{tabular}[c]{@{}l@{}}
In our series of letters from British children, Mia, Jennifer, Karine and Peter tell us what they think of US President Donald Trump.
\end{tabular} \\
& Scores & Mean ROUGE: 33.89 (rank 3) || SummScore rank: \textbf{11} \\

\cdashline{1-3}

Beam \#9   & Summary          & \begin{tabular}[c]{@{}l@{}}
In our series of letters from British children, Mia, Jennifer, Karine and Peter give their views on US President Donald Trump.
\end{tabular} \\
& Scores & Mean ROUGE: 36.19 (rank 2) || SummScore rank: 16 \\

\cdashline{1-3}

Beam \#10  & Summary          & \begin{tabular}[c]{@{}l@{}}
In our series of letters from British children, Mia, Jennifer and Karine tell us what they think about Donald Trump.
\end{tabular} \\
& Scores & Mean ROUGE: 23.16 (rank 5) || SummScore rank: 3  \\

\cdashline{1-3}

Beam \#11  & Summary          & \begin{tabular}[c]{@{}l@{}}
In our series of letters from British schoolchildren, Mia, Jennifer and Karine give their views on Donald Trump.
\end{tabular} \\
& Scores & Mean ROUGE: 24.89 (rank 4) || SummScore rank: 19  \\

\cdashline{1-3}

Beam \#12  & Summary          & \begin{tabular}[c]{@{}l@{}}
In our series of letters from British children, Mia, Jennifer and Karine tell us what they think of Donald Trump.
\end{tabular} \\
& Scores & Mean ROUGE: 23.16 (rank 5) || SummScore rank: 9  \\

\cdashline{1-3}

Beam \#13  & Summary          & \begin{tabular}[c]{@{}l@{}}
In our series of letters from British schoolchildren, Mia, Jennifer and Karine tell us what they think of Donald Trump.
\end{tabular} \\
& Scores & Mean ROUGE: 23.16 (rank 5) || SummScore rank: 13 \\

\cdashline{1-3}

Beam \#14  & Summary          & \begin{tabular}[c]{@{}l@{}}
In our series of letters from British schoolchildren, Mia, Jennifer and Karine tell us what they think about Donald Trump.
\end{tabular} \\
& Scores & Mean ROUGE: 23.16 (rank 5) || SummScore rank: 8 \\

\cdashline{1-3}

Beam \#15  & Summary          & \begin{tabular}[c]{@{}l@{}}
In our series of letters from British children, Mia, Jennifer and Karine give their views on Donald Trump.
\end{tabular} \\
& Scores & Mean ROUGE: 24.89 (rank 4) || SummScore rank: 15  \\

\cdashline{1-3}

Beam \#16  & Summary          & \begin{tabular}[c]{@{}l@{}}
\teal{In our series of letters from British children, Mia, Jennifer, Karine and Peter give their views on Donald Trump.}
\end{tabular} \\
& Scores & Mean ROUGE: \textbf{38.83} (rank 1) || SummScore rank: \textbf{14}  \\

\cdashline{1-3}

Beam \#17  & Summary          & \begin{tabular}[c]{@{}l@{}}
In our series of letters from British children, Mia, Jennifer, Karine and Peter tell us what they think of Donald Trump's presidency.
\end{tabular} \\
& Scores & Mean ROUGE: 33.89 (rank 3) || SummScore rank: 7  \\

\cdashline{1-3}

Beam \#18  & Summary          & \begin{tabular}[c]{@{}l@{}}
In our series of letters from British children, Mia, Jennifer, Karine and Peter tell us what they think about Donald Trump's presidency.
\end{tabular} \\
& Scores & Mean ROUGE: 33.89 (rank 3) || SummScore rank: 2 \\

\cdashline{1-3}

Beam \#19  & Summary          & \begin{tabular}[c]{@{}l@{}}
n our series of letters from British children, Mia, Jennifer, Karine and Peter give their views on US President Donald Trump's presidency.
\end{tabular} \\
& Scores & Mean ROUGE: 33.89 (rank 3) || SummScore rank: 12 \\

\cdashline{1-3}

Beam \#20  & Summary          & \begin{tabular}[c]{@{}l@{}}
\blue{In our series of letters from British children, Mia, Jennifer, Karine and Peter tell us why they think Donald Trump is a fool.}
\end{tabular} \\
& Scores & Mean ROUGE: \textbf{33.89} (rank 3) || SummScore rank: \textbf{1 (SummScore output)}  \\

\midrule

\multicolumn{2}{l}{Reference} & \begin{tabular}[c]{@{}l@{}}
Mia, Jennifer, Karine and Peter do not like Trump.
\end{tabular}   \\
\bottomrule
\end{tabular}
}
\caption{\small SummScore re-ranking applied to the PEGASUS transferred from XSum with beam search on SAMSum.}
\label{tab:h_4_c}
\end{table*}

\begin{table*}
\centering
\resizebox{0.99\textwidth}{!}{
\begin{tabular}{lll}
\toprule
\multicolumn{3}{c}{\textbf{SAMSum: re-ranking from the self-trained PEGASUS}} \\

\midrule

\multicolumn{2}{l}{Source} & \begin{tabular}[c]{@{}l@{}}
Louis: Hey, hows your day? :D \\
Cheryl: Okaaay… I guess \\
Louis: Aha, someone’s feeling a bit down, am I right? \\
Cheryl: yea, sort of… \\
Louis: Go on, tell me what happened \\
Cheryl: I…just had an argument with my mom \\
Louis: Jesus what again \\
Cheryl: I forgot to close the window when I was leaving home! \\
Louis: And that’s it? \\
Cheryl: No, not only… Ya know, wouldn’t be that bad, but I got angry, started screaming and everything ;/ \\
Louis: not a good idea, babe \\
Cheryl: I knoooow  \\
Louis: Was it really bad?  \\
Cheryl: I suppose yea, she kicked me out xd \\
Louis: WHAT \\
Cheryl: I mean I don’t have to move right now, but she gave me time till the end of the year \\
Louis: I’m sorry… \\
Cheryl: Naah, don’t be, I believe it’s for good. I couldn’t stand her anyway xD \\
\end{tabular} \\

\cdashline{1-3}

Beam \#1 & Summary & \begin{tabular}[c]{@{}l@{}}
\orange{Cheryl: No, not only... Ya know, wouldn’t be that bad, but I got angry, started screaming and everything ;/} \\
\orange{Louis: not a good idea, babe Cheryl: I knoooow Louis: Was it really bad?}
\end{tabular} \\
& Scores & Mean ROUGE: \textbf{12.15} (rank 8) || SummScore rank: \textbf{8}  \\

\cdashline{1-3}

Beam \#2 & Summary & \begin{tabular}[c]{@{}l@{}}
Cheryl: no, not only... Ya know, wouldn’t be that bad, but I got angry, started screaming and everything ;/ \\
Louis: not a good idea, babe Cheryl: I knoooow Louis: Was it really bad?
\end{tabular} \\
& Scores & Mean ROUGE: 12.15 (rank 8) || SummScore rank: 10  \\

\cdashline{1-3}

Beam \#3 & Summary & \begin{tabular}[c]{@{}l@{}}
Cheryl: Ya know, wouldn’t be that bad, but I got angry, started screaming and everything ;/ \\
Louis: not a good idea, babe Cheryl: I knoooow Louis: Was it really bad?
\end{tabular} \\
& Scores & Mean ROUGE: 12.73 (rank 7) || SummScore rank: 18  \\

\cdashline{1-3}

Beam \#4   & Summary          & \begin{tabular}[c]{@{}l@{}}
Cheryl: No, not only... Ya know, wouldn’t be that bad, but I got angry, started screaming and everything ;/ \\
Louis: not a good idea, babe Cheryl: I knoooow Louis: Was it really bad? Cheryl:
\end{tabular} \\
& Scores & Mean ROUGE: 11.97 (rank 9) || SummScore rank: 9  \\

\cdashline{1-3}

Beam \#5   & Summary  & \begin{tabular}[c]{@{}l@{}}
 Cheryl: No, not only... Ya know, wouldn’t be that bad, but I got angry, started screaming and everything ;/ \\
 Louis: not a good idea, babe Cheryl: I knoooow Louis: Was it really bad? Cheryl
\end{tabular} \\
& Scores & Mean ROUGE: 11.97 (rank 9) || SummScore rank: 11  \\

\cdashline{1-3}

Beam \#6   & Summary          & \begin{tabular}[c]{@{}l@{}}
Cheryl: No, not only... Ya know, wouldn’t be that bad, but I got angry, started screaming and everything ;/ \\
Louis: not a good idea, babe Cheryl: i knoooow Louis: Was it really bad?
\end{tabular} \\
& Scores & Mean ROUGE: 12.15 (rank 8) || SummScore rank: 16 \\

\cdashline{1-3}

Beam \#7   & Summary          & \begin{tabular}[c]{@{}l@{}}
Cheryl: No, not only... Ya know, wouldn’t be that bad, but I got angry, started screaming and everything ;/ \\
Louis: not a good idea, babe Cheryl:I knoooow Louis: Was it really bad?
\end{tabular} \\
& Scores & Mean ROUGE: 12.15 (rank 8) || SummScore rank: 15  \\

\cdashline{1-3}

Beam \#8   & Summary          & \begin{tabular}[c]{@{}l@{}}
Cheryl: :D Cheryl: Okaaay... I guess Louis: Aha, someone’s feeling a bit down, am I right? Cheryl: yea, sort of... \\
Louis: Go on, tell me what happened Cheryl: I...just had an argument with my mom
\end{tabular} \\
& Scores & Mean ROUGE: 17.23 (rank 5) || SummScore rank: 5 \\

\cdashline{1-3}

Beam \#9   & Summary          & \begin{tabular}[c]{@{}l@{}}
Cheryl: No, not only... Ya know, wouldn’t be that bad, but I got angry, started screaming and everything ;/ \\
Louis: not a good idea, babe Cheryl: I knoooow ; Louis: Was it really bad?
\end{tabular} \\
& Scores & Mean ROUGE: 12.15 (rank 8) || SummScore rank: 12 \\

\cdashline{1-3}

Beam \#10  & Summary          & \begin{tabular}[c]{@{}l@{}}
\teal{Cheryl: yea, sort of... Louis: Go on, tell me what happened Cheryl: I...just had an argument with my mom} \\
\teal{Louis: Jesus what again Cheryl:I forgot to close the window when I was leaving home!}
\end{tabular} \\
& Scores & Mean ROUGE: \textbf{29.19} (rank 1) || SummScore rank: \textbf{17}  \\

\cdashline{1-3}

Beam \#11  & Summary          & \begin{tabular}[c]{@{}l@{}}
Cheryl: yea, sort of... Louis: Go on, tell me what happened Cheryl: I...just had an argument with my mom \\
Louis: Jesus what again Cheryl:I forgot to close the window when I was leaving home! Louis: And that’s it?
\end{tabular} \\
& Scores & Mean ROUGE: 29.00 (rank 2) || SummScore rank: 7  \\

\cdashline{1-3}

Beam \#12  & Summary          & \begin{tabular}[c]{@{}l@{}}
\blue{:D Cheryl: Okaaay... I guess Louis: Aha, someone’s feeling a bit down, am I right? Cheryl: yea, sort of...} \\
\blue{Louis: Go on, tell me what happened Cheryl: I...just had an argument with my mom}
\end{tabular} \\
& Scores & Mean ROUGE: \textbf{17.48} (rank 4) || SummScore rank: \textbf{1 (SummScore output)}  \\

\cdashline{1-3}

Beam \#13  & Summary          & \begin{tabular}[c]{@{}l@{}}
:D Cheryl: Okaaay... I guess Louis: Aha, someone’s feeling a bit down, am I right? Cheryl: yea, sort of... \\
Louis: Go on, tell me what happened Cheryl: I...just had an argument with my mom Louis
\end{tabular} \\
& Scores & Mean ROUGE: 17.23 (rank 5) || SummScore rank: 2 \\

\cdashline{1-3}

Beam \#14  & Summary          & \begin{tabular}[c]{@{}l@{}}
Cheryl: yea, sort of... Louis: Go on, tell me what happened Cheryl:I...just had an argument with my mom \\
Louis: Jesus what again Cheryl: I forgot to close the window when I was leaving home! Louis: And that’s it?
\end{tabular} \\
& Scores & Mean ROUGE: 29.00 (rank 2) || SummScore rank: 6 \\

\cdashline{1-3}

Beam \#15  & Summary          & \begin{tabular}[c]{@{}l@{}}
Cheryl: No, not only... Ya know, wouldn’t be that bad, but I got angry, started screaming and everything ;/ \\
Louis: not a good idea, babe Cheryl: I knoooow Louis: Was it really bad? Louis: Go on.
\end{tabular} \\
& Scores & Mean ROUGE: 11.62 (rank 10) || SummScore rank: 14  \\

\cdashline{1-3}

Beam \#16  & Summary          & \begin{tabular}[c]{@{}l@{}}
Cheryl: :D Cheryl: Okaaay... I guess Louis: Aha, someone’s feeling a bit down, am I right? \\
Louis: Go on, tell me what happened Cheryl: I...ju
st had an argument with my mom Louis: Jesus what again Cheryl:

\end{tabular} \\
& Scores & Mean ROUGE: 16.04 (rank 6) || SummScore rank: 4  \\

\cdashline{1-3}

Beam \#17  & Summary          & \begin{tabular}[c]{@{}l@{}}
:D Cheryl: Okaaay... I guess Louis: Aha, someone’s feeling a bit down, am I right? Cheryl: yea, sort of... \\
Louis: Go on, tell me what happened Cheryl: I...just had an argument with my mom Louis:
\end{tabular} \\
& Scores & Mean ROUGE: 17.23 (rank 5) || SummScore rank: 3  \\

\cdashline{1-3}

Beam \#18  & Summary          & \begin{tabular}[c]{@{}l@{}}
Cheryl: Naah, don’t be, I believe it’s for good. I couldn’t stand her anyway xD
\end{tabular} \\
& Scores & Mean ROUGE: 5.23 (rank 11) || SummScore rank: 19 \\

\cdashline{1-3}

Beam \#19  & Summary          & \begin{tabular}[c]{@{}l@{}}
Cheryl: Ok, I suppose yea, she kicked me out xd Louis: WHAT Cheryl: I mean I don’t have to move right now, but she gave me time till the end of the year \\
Louis: I’m sorry... Cheryl: Naah, don’
\end{tabular} \\
& Scores & Mean ROUGE: 28.10 (rank 3) || SummScore rank: 13 \\

\cdashline{1-3}

Beam \#20  & Summary          & \begin{tabular}[c]{@{}l@{}}
Cheryl: Hi Louis Louis: Hi Cheryl, how are you?
\end{tabular} \\
& Scores & Mean ROUGE: 3.17 (rank 12) || SummScore rank: 20  \\

\midrule

\multicolumn{2}{l}{Reference} & \begin{tabular}[c]{@{}l@{}}
Cheryl had an argument with her mom. She forgot to close the window, got angry and started a fight. Her mom gave her time till the end of the year to move out.
\end{tabular}   \\
\bottomrule
\end{tabular}
}
\caption{\small Self-trained PEGASUS with beam search on SAMSum.}
\label{tab:h_4_d}
\end{table*}

\end{document}